\documentclass{article}

\usepackage[eandd,final,nonanonymous]{neurips_2026}
\setcitestyle{numbers,square,comma,sort&compress}

\usepackage{graphicx}
\usepackage{booktabs}
\usepackage{amsmath,amssymb}
\usepackage{hyperref}
\usepackage{xcolor}
\usepackage{multirow}
\usepackage{threeparttable}
\usepackage{tikz}
\usetikzlibrary{arrows.meta}  %
\usepackage{placeins}  %

\usepackage{amsmath,amsfonts,bm}

\usepackage{amsthm}
\usepackage{mdframed}
\theoremstyle{definition}
\newmdtheoremenv[
  linewidth=0.75pt,
  linecolor=gray!55,
  backgroundcolor=gray!8,
  innertopmargin=6pt,
  innerbottommargin=6pt,
  skipabove=8pt,
  skipbelow=8pt,
]{principle}{Principle}

\def\1{\bm{1}}

\def\rvk{{\mathbf{k}}}

\def\rvm{{\mathbf{m}}}

\def\rvp{{\mathbf{p}}}

\def\rvv{{\mathbf{v}}}

\def\rvx{{\mathbf{x}}}

\def\rmI{{\mathbf{I}}}

\DeclareMathAlphabet{\mathsfit}{\encodingdefault}{\sfdefault}{m}{sl}
\SetMathAlphabet{\mathsfit}{bold}{\encodingdefault}{\sfdefault}{bx}{n}

\def\gN{{\mathcal{N}}}

\newcommand{\E}{\mathbb{E}}

 \usepackage{enumitem}
\usepackage{algorithm}
\usepackage{algpseudocode}
\usepackage{etoc}

\newcommand{\todo}[2]{}
\definecolor{linkblue}{RGB}{26,54,140}

\title{The DynaCell Evaluation Suite for Dynamic 3D Virtual Staining of Live Cells}

\author{
  \textbf{Alexandr A.~Kalinin\textsuperscript{1}\thanks{Equal contribution.}
  \quad
  Dihan Zheng\textsuperscript{2}\footnotemark[1] \quad
  Taylla Milena Theodoro\textsuperscript{1}\footnotemark[1]} \\
  \textbf{Ivan Ivanov\textsuperscript{1} \quad
  Eduardo Hirata-Miyasaki\textsuperscript{1} \quad
  See-Chi Lee\textsuperscript{1} \quad
  Aofei Liu\textsuperscript{1}} \\
  \textbf{Sricharan Reddy Varra\textsuperscript{1} \quad
  Talon Chandler\textsuperscript{1} \quad
  Soorya Pradeep\textsuperscript{1} \quad
  Chad Liu\textsuperscript{1}} \\
  \textbf{Manuel D.~Leonetti\textsuperscript{1} \quad
  Carolina Arias\textsuperscript{1} \quad
  Bo Huang\textsuperscript{2}\thanks{Corresponding authors:     \texttt{bo.huang@ucsf.edu},
    \texttt{shalin.mehta@biohub.org}.} \quad
  Shalin B.~Mehta\textsuperscript{1}\footnotemark[2]}
  \\[6pt]
  \textsuperscript{1}Biohub, San Francisco, 94158 \quad
  \textsuperscript{2}UCSF, San Francisco, 94158
  \\[4pt]
  \href{https://github.com/Biohub/VisCy/tree/main/applications/dynacell}{\raisebox{-0.25ex}{\includegraphics[height=2ex]{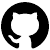}}\,\textcolor{linkblue}{\textbf{Code}}} \qquad
  \href{https://huggingface.co/spaces/biohub/dynacell}{\raisebox{-0.25ex}{\includegraphics[height=2ex]{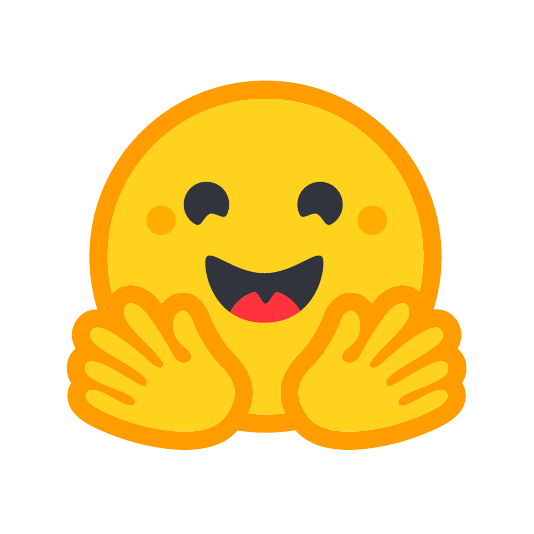}}\,\textcolor{linkblue}{\textbf{Demo}}} \qquad
  \href{https://biohub.github.io/dynacell/}{\raisebox{-0.25ex}{\includegraphics[height=2ex]{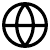}}\,\textcolor{linkblue}{\textbf{Website}}} \qquad
  \href{https://registry.opendata.aws/dynacell/}{\raisebox{-0.25ex}{\includegraphics[height=2ex]{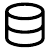}}\,\textcolor{linkblue}{\textbf{Dataset}}}
}

\begin{document}

\maketitle

\begin{abstract}

Understanding how cells respond to perturbations over time is central to cell biology and drug discovery. Live-cell imaging captures the dynamics of perturbation-induced phenotypes in individual cells. While fluorescence imaging can visualize specific molecular phenotypes, labeling and repeated illumination often affect cell health. Virtual staining offers an alternative by predicting these fluorescence images from images acquired without fluorescent labels. Although many virtual staining models have been developed, their performance on volumetric time-lapse data remains poorly characterized. We address this gap with an evaluation suite for 3D time-lapse virtual staining of live cells. The suite combines paired 3D time-lapse images acquired with and without fluorescent labels, a set of baseline models, and three metric tiers measuring pixel-level fidelity, organelle segmentation performance, and single-cell phenotype similarity. Using this evaluation suite, we study metric and model behavior across 3 cell types, 2 microscopes, 4 organelles, and 3 perturbation states. We find that regression baselines better predict the spatial localization of target organelles, whereas the generative baseline better captures the distribution of phenotypes across cells. These differences expose trade-offs between structural and phenotypic fidelity that are obscured by single-metric evaluation and suggest how virtual staining models can be developed for diverse downstream tasks such as localization, segmentation, tracking, and phenotypic profiling. We release the curated datasets, baseline checkpoints, and evaluation code as part of \mbox{DynaCell}, a broader framework for image-based profiling of cell-state dynamics.

\end{abstract}

\addtocontents{toc}{\protect\setcounter{tocdepth}{-2}}

\section{Introduction}
\label{sec:introduction}

Image-based profiling of cells under genetic and chemical perturbations has become central to scalable cell biology and drug discovery~\citep{chandrasekaran2021image,serrano2026progress}. Many perturbations induce changes in organelle organization and cellular state that unfold over time at different rates. However, capturing these dynamics with repeated live-cell measurements requires balancing the molecular specificity of fluorescence imaging against phototoxicity, photobleaching, and other effects on cell health. Virtual staining addresses this trade-off by predicting fluorescence-like images of cell structures from transmitted-light images acquired without fluorescent labels (``label-free'' imaging). The same cells can then be imaged repeatedly over time, without genetic modification or fluorescent dyes~\citep{christiansen2018silico,ounkomol2018label,elmalam2024silico}. Many image-to-image translation models have been developed for cell virtual staining~\citep{christiansen2018silico,ounkomol2018label,guo2020revealing,wang2021global,crosszamirski2022labelfree,crosszamirski2023classguided,francobarranco2024biapy,le2024insilicopainting,zhou2023repmode,zhou2024bagofexperts,wang2023brightfield,liu2025robust,kalinin2025spotlight} and applied to downstream tasks including nucleus and cell segmentation~\citep{ling2020analyzing,fishman2021practical}, label-free tracking~\citep{bragantini2025ultrack,theodoro2026dynatrack}, and perturbation profiling~\citep{crosszamirski2022labelfree,crosszamirski2023classguided}. However, their performance on volumetric time-lapse data remains poorly characterized, limiting both the practical use of virtual staining in biological studies and further method development.

\begin{figure}[!htb]
  \centering
  \includegraphics[width=\textwidth]{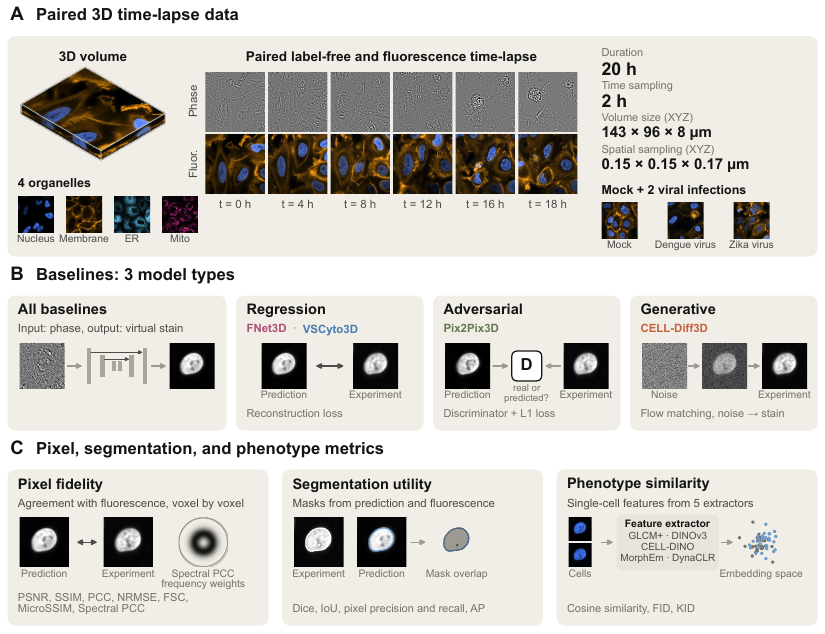}
  \caption{\textbf{DynaCell evaluation suite.} The suite consists of
   \textbf{(A)} high-resolution 3D time-lapse multichannel datasets, \textbf{(B)} regression, adversarial, and generative virtual-staining baselines trained and evaluated with a consistent protocol, and \textbf{(C)} three metric tiers that compare experimental fluorescence with virtual stains at the pixel, segmentation, and phenotype levels. Spectral PCC reduces sensitivity to noise in experimental fluorescence.}
  \label{fig:evaluation}
\end{figure}

Evaluation of virtual staining for live-cell profiling is constrained by limited paired live-cell data and inconsistent protocols. Public paired collections are predominantly static and 2D~\citep{chandrasekaran2023jump,lightmycells2024,trizna2023brightfield,wyrzykowska2025virvs}. Paired 3D data are less common, particularly time-lapse series capturing cell-state changes in response to perturbations~\citep{ounkomol2018label,viana2023integrated,guo2020revealing,ivanov2024mantis}. Most studies report only pixel-level reconstruction metrics~\citep{christiansen2018silico,ounkomol2018label,wang2021global,francobarranco2024biapy,le2024insilicopainting,zhou2023repmode,zhou2024bagofexperts,wyrzykowska2025virvs}, which measure resemblance to the target but not whether predictions can replace experimental fluorescence in organelle localization, segmentation, cell tracking, or phenotypic profiling. Segmentation~\citep{guo2020revealing,wang2023brightfield,liu2025robust,kalinin2025spotlight} and feature-level metrics~\citep{crosszamirski2022labelfree,crosszamirski2023classguided,kalinin2025spotlight} are less common and are rarely combined under shared splits and baselines.

To address these gaps, we introduce an evaluation suite for 3D time-lapse virtual staining of live cells, built around three linked components (Fig.~\ref{fig:evaluation}). This evaluation suite is part of DynaCell, a broader framework for image-based profiling of cell-state dynamics.

\textbf{Data.} To enable model comparison on dynamic live-cell data, we contribute a high-resolution 3D time-lapse A549 dataset acquired with paired label-free and light-sheet fluorescence imaging (Fig.~\ref{fig:evaluation}A). The dataset covers 20 hours of imaging of 4 organelle targets: nucleus, plasma membrane, endoplasmic reticulum (ER), and mitochondria. Infection with Zika virus (ZIKV) or Dengue virus (DENV) induces organelle remodeling over time~\citep{huang2024sun2,freppel2025dengue}. A semi-automated preprocessing and quality-control pipeline reduces the raw data to a curated, model-ready subset (App.~\ref{sec:data-preprocessing-details}). The curated dataset contains 262 field-of-view (FOV) time-lapse series across 12 organelle--condition combinations (four organelles under Mock, ZIKV, or DENV conditions). The test set contains 12 FOVs per organelle--condition combination; the remaining quality-controlled FOVs form the A549 training set. We also contribute reprocessed, curated subsets of other datasets for evaluation across cell types and microscopes: the Allen Institute WTC-11 human induced pluripotent stem cell (hiPSC) dataset~\citep{viana2023integrated} and the Biohub human embryonic kidney (HEK293T) dataset~\citep{ivanov2024mantis}.

\textbf{Baseline models.} To establish reference models, we train regression baselines ranging from a widely used 3D U-Net variant~\citep{ounkomol2018label} to modern convolutional models with masked-autoencoder pretraining~\citep{liu2025robust}. We also implement a 3D extension of the popular pix2pix model~\citep{isola2017image} and a 3D flow-matching model based on the 2D diffusion model CELL-Diff~\citep{zheng2025bridging} (Fig.~\ref{fig:evaluation}B).

\textbf{Metrics.} To evaluate virtual stains beyond image reconstruction alone, we use a three-tier panel spanning pixel-level fidelity, organelle-segmentation utility, and single-cell phenotype similarity (Fig.~\ref{fig:evaluation}C). We also introduce a frequency-aware variant of the Pearson correlation coefficient (PCC) to reduce sensitivity to noise in experimental fluorescence targets.

We release preprocessed data, trained model checkpoints, and reusable evaluation code for reproducible comparison. Together, these components help determine when virtual stains can support live-cell biological measurements, how failure modes can inform metric selection and model training, and how model choices affect downstream use.

\section{Background and Related Work}
\label{sec:related}

\textbf{Virtual staining methods.} Early virtual staining models~\citep{ounkomol2018label,christiansen2018silico} were U-Net~\citep{ronneberger2015unet} variants trained with pixel-wise mean absolute error (MAE) or mean squared error (MSE) regression losses. Subsequent work modernized the convolutional backbone~\citep{guo2020revealing,liu2025robust}, added self-attention modules~\citep{wang2021global}, predicted multiple organelles~\citep{zhou2023repmode}, or pretrained encoders on unannotated label-free images~\citep{francobarranco2024biapy,zhou2024bagofexperts,liu2025robust}. Training objectives have also incorporated structural similarity~\citep{wang2023brightfield,liu2025robust}, segmentation~\citep{wang2023brightfield,kalinin2025spotlight}, and adversarial losses~\citep{crosszamirski2022labelfree,wang2023brightfield,le2024insilicopainting,wyrzykowska2025virvs,tonks2025virtual}. Generative models that sample from a conditional distribution offer an alternative to regression. These models, including conditional diffusion~\citep{crosszamirski2023classguided,zheng2025bridging} and autoregressive~\citep{kambhampati2026vermeer} models, remain limited to 2D virtual staining.

\textbf{Virtual staining datasets and benchmarks.} Virtual staining models and their evaluation are currently limited by the available paired data, because 2D images contain no depth information needed for resolving 3D organelle structure~\citep{meyer2009automated,kalinin2018threedmorphometry,kalinin2021valproic}, static volumes contain no temporal information about cellular dynamics or perturbation responses, and small collections cover few cell states and conditions. Paired 2D static-snapshot datasets are available~\citep{chandrasekaran2023jump,lightmycells2024,trizna2023brightfield,wyrzykowska2025virvs}, but only Light My Cells~\citep{lightmycells2024}, Trizna et al.~\citep{trizna2023brightfield}, and VIRVS~\citep{wyrzykowska2025virvs} provide both baselines and scoring protocols, and none combine pixel, segmentation, and phenotype metrics. Paired 3D collections are rare and still lack a time axis~\citep{ounkomol2018label,guo2020revealing,viana2023integrated}. Public paired 3D time-lapse data are scarce~\citep{ivanov2024mantis} and lack the scale, organelle coverage, and perturbation panels needed for systematic evaluation (App.~Tab.~\ref{tab:dataset-comparison}).

\textbf{Metrics for evaluating virtual staining.} Pixel-level metrics such as peak signal-to-noise ratio (PSNR), structural similarity index measure (SSIM)~\citep{wang2004image}, normalized root mean square error (NRMSE), and PCC dominate virtual-staining evaluation~\citep{christiansen2018silico,ounkomol2018label,wang2021global,zhou2023repmode,wyrzykowska2025virvs,francobarranco2024biapy,le2024insilicopainting,zhou2024bagofexperts,lightmycells2024}. Natural-image quality metrics span only a small part of their nominal range, and their values vary with cell line, seeding density, and fluorescence channel~\citep{li2026doubting}. Microscopy-specific metrics such as Fourier shell correlation~\citep{harauz1986exact,koho2019fourier} and MicroSSIM~\citep{ashesh2024microssim} are used less frequently~\citep{kalinin2025spotlight,li2026doubting}, although they can be more sensitive to fine spatial structures of organelles. Segmentation metrics are computed on masks derived from predicted and reference fluorescence. They are reported less commonly~\citep{guo2020revealing,trizna2023brightfield,wang2023brightfield,liu2025robust,kalinin2025spotlight} and measure whether a virtual stain can support downstream tasks such as organelle localization or cell tracking. Feature-level metrics are also uncommon~\citep{crosszamirski2022labelfree,crosszamirski2023classguided,tonks2025virtual,tonks2026proper,kalinin2025spotlight}. They use feature extractors such as CellProfiler~\citep{stirling2021cellprofiler} or learned encoders to represent each cell, then compare predictions and experiments in the resulting feature spaces.

\textbf{Generalization.} Tonks et al.~\citep{tonks2025virtual} found that 2D virtual-nucleus and virtual-cytoplasm models trained on control conditions generalized to toxic conditions with the high compound concentrations and apoptosis-inducing compounds, whereas generalization to unseen cell types depended on the cell type. CellProfiler~\citep{stirling2021cellprofiler} features revealed failures missed by pixel- and instance-level scores. Cytoland~\citep{liu2025robust} showed that data augmentation and self-supervised pretraining improved the robustness of virtual staining across imaging parameters, cell states, and cell types.

The DynaCell evaluation suite addresses these gaps together. We reprocess a curated subset of the Allen Institute WTC-11 hiPSC dataset, the largest available paired 3D live-cell dataset, for paired training and in-distribution evaluation (Sec.~\ref{sec:dataset-training}). We combine the iPSC dataset with a new 3D time-lapse A549 dataset covering four organelles and three perturbation states (Sec.~\ref{sec:dataset-a549}), regression, adversarial, and generative baselines trained with a consistent protocol (Sec.~\ref{sec:baseline-models}), and a three-tier evaluation that combines pixel, segmentation, and feature-level metrics (Sec.~\ref{sec:evaluation-framework}).

\section{Evaluation Data}
\label{sec:datasets}

\begin{figure}[ht!]
  \centering
  \includegraphics[width=0.9\textwidth]{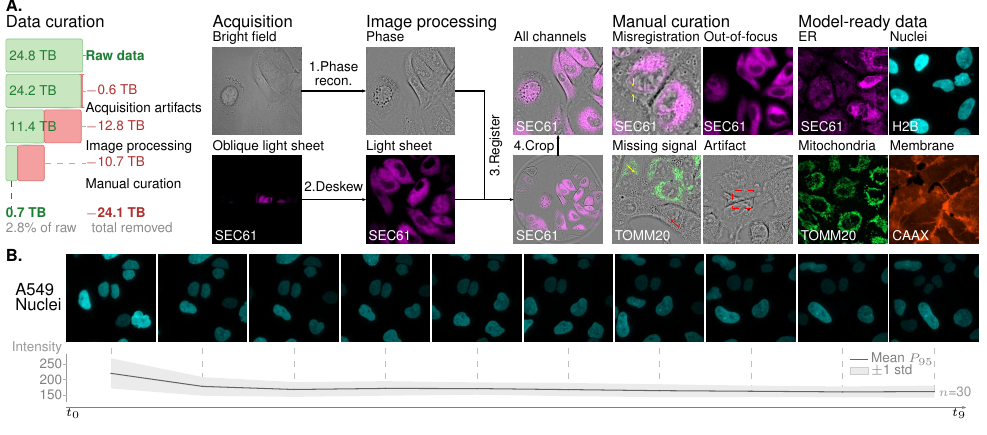}
  \caption{\textbf{A549 preprocessing and photobleaching.} \textbf{(A)} Preprocessing the raw 3D movies ($\sim$25\,TB) includes quantitative phase reconstruction in the label-free channel, fluorescence deskewing, 3D channel registration, cropping, manual curation, and subsampling. The resulting Zarr store is ${<}1$\,TB. \textbf{(B)} Nuclear fluorescence intensity decreases during imaging, motivating the frequency-aware Spectral PCC metric (Sec.~\ref{sec:eval-protocol}).}
  \label{fig:preprocessing}
\end{figure}

\subsection{A549 dataset (Biohub imaging infectomics dataset)}
\label{sec:dataset-a549}

The A549 dataset contains 3D time-lapse volumes of live A549 cells. Each transmitted-light (``label-free'') image is paired with a fluorescence image of the same cells at the same time point (Fig.~\ref{fig:evaluation}A). Volumes were acquired on the Mantis microscope~\citep{ivanov2024mantis}, which simultaneously captures label-free and light-sheet fluorescence channels at near-isotropic sampling with $0.174 \times 0.1494 \times 0.1494\,\mu$m (ZYX) voxel spacing. Low-phototoxicity light-sheet illumination enables continuous imaging over 20 hours. At each time point, each FOV includes a brightfield stack paired with a light-sheet fluorescence stack for one of the four markers: H2B (nucleus), a CAAX membrane-targeting motif (plasma membrane), SEC61B (endoplasmic reticulum), or TOMM20 (mitochondria)~\citep{cidrosas2026vswitch,czbiohub2025organellebox}. These four organelles are perturbed by Zika and Dengue infection, which has previously been shown to cause their remodeling~\citep{huang2024sun2,freppel2025dengue}. These organelles span nuclear morphology, cell boundaries, and intracellular architecture, revealing cell-state dynamics across scales.

\paragraph{Data preprocessing and curation.}
We reconstructed quantitative phase volumes from brightfield stacks~\citep{chandler2024waveorder}, deskewed the light-sheet fluorescence stacks, and registered the two modalities using a separate calibration field of view containing fluorescent beads~\citep{ivanov2024mantis}. We cropped the registered data to retain overlapping regions around the focal plane and excluded time points affected by acquisition or processing failures. We subsampled the A549 time-lapses onto a shared 2-hour grid to compare plates at matching times after infection. Lastly, we identified remaining problems by manual inspection, including out-of-focus images, missing fluorescence, and imaging artifacts. After processing and curation, a model-ready subset remained for each organelle (Fig.~\ref{fig:preprocessing}A; details in App.~\ref{sec:data-preprocessing-details}). Even with light-sheet imaging, fluorescent labels photobleach over the multihour acquisition, reducing fluorescence intensity over time (Fig.~\ref{fig:preprocessing}B). This signal variation motivates the frequency-aware Spectral PCC metric (Sec.~\ref{sec:eval-protocol}).

\paragraph{Splits and benchmarking axes.}
\label{sec:splits}
The dataset contains 118 training FOVs and 144 test FOVs, with balanced test coverage across organelles and conditions and approximately 17,000 test cells. Training and test FOVs come from distinct well positions. For each model, we held out 20\% of the training FOVs for validation with a fixed random seed and did not train on them. We never used the test set for training, checkpoint selection, or hyperparameter tuning. We also included an untreated, unpaired movie sampled every 10 min to evaluate temporal consistency (Sec.~\ref{sec:stability-perturbation-time}). Further data acquisition and split details are provided in App.~\ref{sec:a549-split-details}.

\subsection{iPSC dataset (Allen Institute WTC-11 hiPSC dataset)}
\label{sec:dataset-training}

The iPSC dataset was curated from the Allen Institute WTC-11 hiPSC dataset~\citep{viana2023integrated}, from which we selected the four benchmark organelles: endogenously tagged ER (SEC61B) and mitochondria (TOMM20), and chemically stained nucleus (Hoechst) and plasma membrane (CellMask). These cells were imaged live on a 3D confocal microscope. We reconstructed quantitative phase from brightfield stacks~\citep{chandler2024waveorder}, yielding paired phase--fluorescence data from a different cell type and microscope than A549. After quality control, the training set contains 500 FOVs each for nucleus, membrane, and mitochondria, and 423 for ER. The test set has 100 FOVs per organelle. Volume sizes (ZYX) are $44 \times 624 \times 924$ for training and $40 \times 512 \times 512$ for testing.

\subsection{HEK293T evaluation data}
\label{sec:dataset-hek}

To help separate cell-type from microscope variation, we reused 8 single-timepoint HEK293T FOVs from the Mantis dataset~\citep{ivanov2024mantis}: 4 with the cell membrane marker KRAS and 4 with the mitochondrial marker TOMM70A. These images were acquired on the same microscope as the A549 data and processed using phase reconstruction, fluorescence deskewing, channel registration, and cropping. The resulting $64 \times 960 \times 1184$ (ZYX) volumes contain approximately 120 cells per organelle. We used HEK293T only for evaluation, never for training (App.~\ref{sec:hek-evaluation-data}).

\section{Evaluation Metrics}
\label{sec:evaluation-framework}
\label{sec:eval-protocol}

We evaluate virtual stains across three metric tiers (Fig.~\ref{fig:evaluation}C). Each measures a different aspect of prediction quality, and together they reveal trade-offs that a single metric may obscure.

\textbf{Pixel fidelity.} Pixel-level metrics test whether the prediction agrees with experimental fluorescence at the spatial precision needed for direct visualization, colocalization, and intensity measurements. We report PSNR, SSIM, and NRMSE in both their conventional and scale-invariant (si-) forms, together with PCC, Fourier shell correlation (FSC)~\citep{koho2019fourier,kalinin2025cubic}, MicroSSIM~\citep{ashesh2024microssim}, and Spectral PCC. For each scale-invariant metric, we linearly rescale the prediction to best match the target intensities. These metrics remain sensitive to small spatial offsets and treat experimental fluorescence as the reference, even though fluorescence is itself a noisy measurement of the underlying biology.

\textit{Spectral PCC.} Live-cell imaging over extended periods causes photobleaching (Fig.~\ref{fig:preprocessing}B) and reduces the signal-to-noise ratio (SNR). High frequencies become dominated by noise, while low frequencies reflect autofluorescence and illumination shading rather than organelle structure. Standard PCC penalizes a prediction for not reproducing these non-biological artifacts in the target images. We propose Spectral PCC, a Fourier-domain correlation coefficient that excludes the lowest-frequency bins and weights the remaining frequencies by target signal above a noise floor estimated from the high-frequency tail~(App.~\ref{sec:spectral_pcc}).

\textbf{Segmentation utility.} Segmentation metrics test whether a predicted stain can substitute for experimental fluorescence in a downstream segmentation pipeline. For nucleus and membrane, we identify the focal plane and compute a maximum-intensity projection of the three Z slices centered on the focal plane. We then segment both predicted and target projections using the Cellpose v4 cpdino-ViT-L model~\citep{pachitariu2025cellpose} and compute instance Dice and average precision (AP). We report AP at an intersection-over-union (IoU) threshold of $0.50$ and mean AP over IoU thresholds $0.50{:}0.95$. For ER and mitochondria, we use the Allen Cell and Structure Segmenter~\citep{chen2018allen,viana2023integrated}. We also compute semantic Dice, IoU, precision, recall, and accuracy from binary masks for all four organelles. Without manual masks for A549, we could not validate segmenter accuracy, but on A549 nuclei, Cellpose v4 did not fail under infection and agreed with the Allen Cell and Structure Segmenter~(App.~\ref{sec:segmenter-ablation}).

\textbf{Phenotype similarity.} Phenotype metrics check whether a predicted stain preserves the distribution of single-cell morphology used in biological profiling, even when an individual prediction does not match its experimental counterpart voxel by voxel. For nucleus and membrane, we localize single cells by applying Cellpose v4 to target fluorescence. For ER and mitochondria, we apply Cellpose v4 to fluorescence predictions generated from phase images by Cytoland~\citep{liu2025robust}. We then apply the same masks to every model's prediction and its paired fluorescence target, avoiding model-dependent cell selection. We represent each cell with feature vectors (embeddings) from five feature extractors: GLCM+, DINOv3~\citep{simeoni2025dinov3}, DynaCLR~\citep{hirata2024dynaclr}, CELL-DINO~\citep{moutakanni2025celldino}, and MorphEm~\citep{agrawal2026chammi} (App.~Fig.~\ref{fig:feature_similarities}). GLCM+ combines gray-level co-occurrence matrix (GLCM) features with intensity-moment, gradient, and Laplacian descriptors, giving 22 handcrafted 3D features. Because the pretrained extractors we use operate on 2D inputs, we use maximum projections of five Z slices centered on the in-focus plane, whereas GLCM+ is computed on the full 3D volume. In each extractor's feature space, we compute the median cosine similarity between paired cells, together with the Fr\'echet Inception Distance (FID)~\citep{heusel2017gans} and Kernel Inception Distance (KID)~\citep{binkowski2018demystifying} between the predicted and experimental distributions~(App.~\ref{sec:feat_simi}). We abbreviate the GLCM+ median cosine similarity as ``GLCM+ cos.''

\section{Baseline Models}
\label{sec:baseline-models}

The main benchmark (Fig.~\ref{fig:evaluation}B) compares four models: the regression baselines FNet3D and VSCyto3D, adversarial Pix2Pix3D, and generative CELL-Diff3D. We also evaluated CELL-Diff3D's conditional-mean inference and a VSCyto3D reference initialized with random weights. We trained separate models for nucleus, membrane, ER, and mitochondria. App.\ Fig.~\ref{fig:baseline-architectures} shows the model architectures and training objectives.

\paragraph{Regression baselines.}
Regression baselines produce one spatially registered fluorescence estimate for each label-free input. FNet3D~\citep{ounkomol2018label} uses a 3D U-Net trained with MSE loss. VSCyto3D~\citep{liu2025robust} is a ConvNeXtV2 U-Net with an encoder pretrained with a fully convolutional masked autoencoder~\citep{woo2023convnext}. We trained VSCyto3D with a combined MAE/SSIM loss. We also evaluated an untrained VSCyto3D (``Rand-init'') to provide an empirical lower bound for interpreting the metrics~(Sec.~\ref{sec:metric-selection}). We trained VSCyto3D without encoder pretraining to isolate the contribution of pretraining~(App.~\ref{sec:vscyto3d-pretraining-ablation}).

\paragraph{Adversarial baseline.}
We introduce Pix2Pix3D, a new 3D adaptation of the popular pix2pix model~\citep{isola2017image} implemented for this benchmark. Pix2Pix3D extends the conditional adversarial objective~\citep{isola2017image} to volumetric data using a 3D U-Net generator and a multiscale 3D PatchGAN discriminator~\citep{wang2018highresolution}. Training combines the non-saturating adversarial objective~\citep{goodfellow2014generative} with an $L_1$ reconstruction loss and R1~\citep{mescheder2018training} and LeCam~\citep{tseng2021regularizing} regularization. We also maintained an exponential moving average of the generator weights (App.~Tab.~\ref{tab:benchmark_training}). Pix2Pix3D learns a deterministic mapping from input to target space~\citep{isola2017image}, receiving no noise input and producing one prediction per phase volume.

\paragraph{Generative baseline.}
The generative baseline allows a given phase volume to correspond to different organelle morphologies, especially for fine structures only partly visible in phase. We introduce CELL-Diff3D, a 3D flow-matching extension of the 2D diffusion-based CELL-Diff model~\citep{zheng2025bridging}. CELL-Diff3D samples plausible fluorescence volumes conditioned on phase, allowing us to evaluate population-level phenotypic realism. Independently generated 3D tiles can disagree at their boundaries and create seams. We therefore developed an iterative generation scheme based on flow-matching inpainting to maintain consistency across tiles during large-volume inference (App.~\ref{sec:iter_pred_fm}). We also evaluated CELL-Diff3D's single-pass conditional mean (App.~\ref{sec:mean_pred_fm}).

\paragraph{Training and inference protocol.}
For each organelle and model family, we trained on iPSC, A549, and the joint iPSC$+$A549 set, then evaluated every version on both cell types. We report batch size, learning rate, patch size, loss, and target normalization in App.~Tab.~\ref{tab:benchmark_training}. All model families use the same training pools, held-out test sets, and 80/20 training/validation ratio. All baselines use the shared augmentation pipeline~(App.~Tab.~\ref{tab:augmentation}), except FNet3D, which uses only random flips~\citep{ounkomol2018label}.

\section{Results}
\label{sec:results}

We used the DynaCell evaluation suite to study how virtual staining quality changes with the training dataset and model family. Pixel fidelity, segmentation, and phenotype similarity measure qualities relevant to different downstream tasks, so we interpret these metrics together. We report key findings below and further predictions across organelles and conditions in App.~\ref{sec:pred_overview_appendix}.

\subsection{Generalization gaps inform metric selection}
\label{sec:metric-selection}

\begin{figure}[t]
  \centering
  \includegraphics[width=\linewidth]{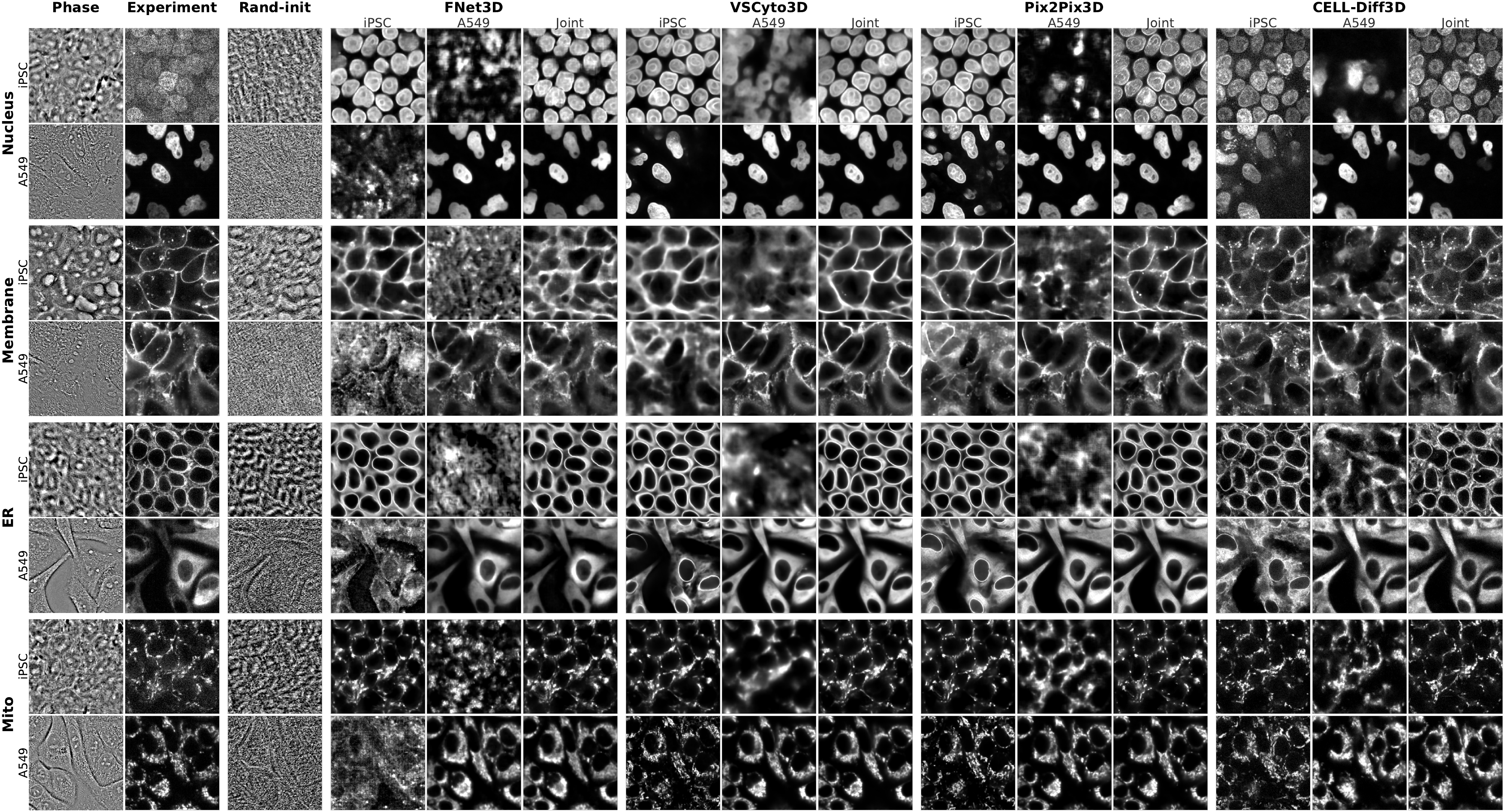}
  \caption{\textbf{Organelle prediction overview across models and training sets.} Single in-focus XY slices of virtual-staining predictions for nucleus, cell membrane, ER, and mitochondria on the iPSC test set (top) and A549 test set (bottom). Columns show the phase input, experimental stain, and predictions from randomly initialized VSCyto3D (Rand-init), FNet3D, VSCyto3D, Pix2Pix3D, and CELL-Diff3D. The four learned models are each trained on iPSC, A549, or both datasets.}
  \label{fig:prediction-overview}
\end{figure}

We trained each model on iPSC, A549, or both datasets jointly and evaluated the models on both cell types. Testing single-dataset models on the other dataset assesses generalization to out-of-distribution (OOD) data. Models generally reproduced organelle location and shape more accurately on the dataset used for training~(Fig.~\ref{fig:prediction-overview}). Some metrics reflected these generalization gaps more reliably than others. We started with a large library of metrics and selected those that enable robust analysis of OOD generalization gaps.

Predictions from a randomly initialized VSCyto3D model (``Rand-init'') contained little recognizable organelle structure and provided an empirical lower bound for interpreting the metrics. FNet3D's OOD predictions retained fluorescence-like texture but failed to reproduce the target organelle's location or shape, largely due to FNet3D training using small patches and basic augmentations~(App.~\ref{sec:fnet_patch_aug_ablation}).

We evaluated every metric in the pixel~(App.~Tab.~\ref{tab:mr-fnet3d-pixel}), segmentation~(App.~Tab.~\ref{tab:mr-fnet3d-seg}), and phenotype~(App.~Tab.~\ref{tab:mr-fnet3d-pheno}) tiers using Rand-init and FNet3D's OOD predictions~(App.~\ref{sec:metric_reliability_appendix}). Several metrics scored these poor predictions as favorably as in-distribution predictions. Some did so on the same test images and thus missed the failure, whereas others did so only across test sets, because their absolute values depend on the test images~\citep{li2026doubting}. Among pixel metrics, si-SSIM and Spectral PCC ranked the poor predictions correctly in both comparisons, unlike SSIM; we kept Spectral PCC over PCC because it is more robust to photobleaching~(App.~\ref{sec:spectral_pcc}). All segmentation metrics except pixel accuracy did so, and we kept Dice because it applies to all four organelles. Among the phenotype metrics computed with learned feature extractors, we chose to report CELL-DINO KID because it most consistently assigned less favorable scores to poor FNet3D OOD predictions and separated the models clearly, such that the model rankings remained consistent across embedding models under OOD evaluation. GLCM+ cos scored FNet3D OOD predictions below the in-distribution results for most organelles, more consistently than any other phenotype metric. We compare the feature extractors in detail in App.~\ref{sec:feature-space-agreement}.

The final selected five metric set included si-SSIM and Spectral PCC for pixel fidelity, Dice for segmentation, and CELL-DINO KID and GLCM+ cos for phenotype similarity. We used instance Dice for nucleus and membrane to assess individual objects, and semantic Dice for ER and mitochondria, whose masks represent organelle structures rather than separate cells.

\definecolor{tierpixel}{RGB}{46,139,131}
\definecolor{tierseg}{RGB}{201,162,39}
\definecolor{tierpheno}{RGB}{74,90,115}
\begin{table}[ht]
\centering
\small
\caption{\textbf{Prediction metrics across training and test datasets, per organelle.} Rows show five selected metrics for VSCyto3D, Pix2Pix3D, and CELL-Diff3D. Dice denotes instance Dice for nucleus and membrane and semantic Dice for ER and mitochondria. SPCC denotes Spectral PCC. CD KID denotes CELL-DINO KID. Columns are grouped by test dataset, then model and training dataset; Joint denotes training on the combined iPSC and A549 data. For each organelle, metric, and test dataset, bold compares all models across all training sets, marking the best mean and means within its SD, computed at the displayed precision. App.~\ref{sec:metric_reliability_appendix} reports the full FNet3D panel used to select these rows. SD values are omitted for readability. Colored squares denote the metric tier: \textcolor{tierpixel}{\rule{1.1ex}{1.1ex}}~pixel, \textcolor{tierseg}{\rule{1.1ex}{1.1ex}}~segmentation, \textcolor{tierpheno}{\rule{1.1ex}{1.1ex}}~phenotype. Arrows show the direction of metric improvement.}
\label{tab:mr-pix2pix}
\resizebox{\textwidth}{!}{%
\begin{tabular}{l cccccccccccccccccc}
\toprule
 & \multicolumn{9}{c}{\textbf{iPSC test}} & \multicolumn{9}{c}{\textbf{A549 test}} \\
\cmidrule(lr){2-10}\cmidrule(lr){11-19}
 & \multicolumn{3}{c}{\textbf{VSCyto3D}} & \multicolumn{3}{c}{\textbf{Pix2Pix3D}} & \multicolumn{3}{c}{\textbf{CELL-Diff3D}} & \multicolumn{3}{c}{\textbf{VSCyto3D}} & \multicolumn{3}{c}{\textbf{Pix2Pix3D}} & \multicolumn{3}{c}{\textbf{CELL-Diff3D}} \\
\cmidrule(lr){2-4}\cmidrule(lr){5-7}\cmidrule(lr){8-10}\cmidrule(lr){11-13}\cmidrule(lr){14-16}\cmidrule(lr){17-19}
\textbf{Metric} & \textbf{iPSC} & \textbf{A549} & \textbf{Joint} & \textbf{iPSC} & \textbf{A549} & \textbf{Joint} & \textbf{iPSC} & \textbf{A549} & \textbf{Joint} & \textbf{A549} & \textbf{iPSC} & \textbf{Joint} & \textbf{A549} & \textbf{iPSC} & \textbf{Joint} & \textbf{A549} & \textbf{iPSC} & \textbf{Joint} \\
\midrule
\multicolumn{19}{l}{\textbf{Nucleus}} \\
  \textcolor{tierpixel}{\rule{1.1ex}{1.1ex}}~si-SSIM ($\uparrow$) & $\mathbf{0.58}$ & $0.38$ & $\mathbf{0.58}$ & $\mathbf{0.58}$ & $0.38$ & $\mathbf{0.56}$ & $\mathbf{0.47}$ & $0.31$ & $\mathbf{0.46}$ & $\mathbf{0.89}$ & $0.68$ & $\mathbf{0.90}$ & $\mathbf{0.89}$ & $0.77$ & $\mathbf{0.89}$ & $0.83$ & $0.52$ & $0.85$ \\
  \textcolor{tierpixel}{\rule{1.1ex}{1.1ex}}~SPCC ($\uparrow$) & $\mathbf{0.90}$ & $0.56$ & $\mathbf{0.90}$ & $\mathbf{0.90}$ & $0.59$ & $\mathbf{0.89}$ & $0.81$ & $0.38$ & $0.81$ & $\mathbf{0.89}$ & $0.70$ & $\mathbf{0.90}$ & $\mathbf{0.91}$ & $\mathbf{0.86}$ & $\mathbf{0.91}$ & $\mathbf{0.87}$ & $0.72$ & $\mathbf{0.89}$ \\
  \textcolor{tierseg}{\rule{1.1ex}{1.1ex}}~Dice ($\uparrow$) & $\mathbf{0.89}$ & $0.63$ & $\mathbf{0.89}$ & $\mathbf{0.89}$ & $0.64$ & $\mathbf{0.89}$ & $\mathbf{0.84}$ & $0.39$ & $\mathbf{0.83}$ & $\mathbf{0.90}$ & $0.69$ & $\mathbf{0.91}$ & $\mathbf{0.92}$ & $0.73$ & $\mathbf{0.92}$ & $0.76$ & $0.67$ & $0.84$ \\
  \textcolor{tierpheno}{\rule{1.1ex}{1.1ex}}~CD KID ($\downarrow$) & $24.93$ & $30.62$ & $24.91$ & $23.99$ & $29.14$ & $16.40$ & $1.72$ & $12.28$ & $\mathbf{1.65}$ & $30.88$ & $19.80$ & $29.53$ & $26.04$ & $19.43$ & $20.10$ & $4.16$ & $9.95$ & $\mathbf{4.04}$ \\
  \textcolor{tierpheno}{\rule{1.1ex}{1.1ex}}~GLCM+ cos ($\uparrow$) & $-0.00$ & $-0.10$ & $0.00$ & $0.02$ & $-0.18$ & $0.02$ & $\mathbf{0.26}$ & $-0.18$ & $0.24$ & $0.36$ & $0.22$ & $0.35$ & $0.39$ & $0.35$ & $0.39$ & $0.52$ & $-0.06$ & $\mathbf{0.55}$ \\
\midrule
\multicolumn{19}{l}{\textbf{Membrane}} \\
  \textcolor{tierpixel}{\rule{1.1ex}{1.1ex}}~si-SSIM ($\uparrow$) & $\mathbf{0.58}$ & $0.31$ & $\mathbf{0.60}$ & $\mathbf{0.60}$ & $0.35$ & $\mathbf{0.59}$ & $0.47$ & $0.32$ & $\mathbf{0.48}$ & $\mathbf{0.64}$ & $0.43$ & $\mathbf{0.65}$ & $\mathbf{0.64}$ & $0.42$ & $\mathbf{0.64}$ & $\mathbf{0.57}$ & $0.41$ & $\mathbf{0.57}$ \\
  \textcolor{tierpixel}{\rule{1.1ex}{1.1ex}}~SPCC ($\uparrow$) & $\mathbf{0.89}$ & $0.26$ & $\mathbf{0.89}$ & $\mathbf{0.88}$ & $0.53$ & $\mathbf{0.88}$ & $\mathbf{0.82}$ & $0.49$ & $\mathbf{0.83}$ & $\mathbf{0.88}$ & $0.57$ & $\mathbf{0.89}$ & $\mathbf{0.89}$ & $0.57$ & $\mathbf{0.89}$ & $0.83$ & $0.63$ & $0.83$ \\
  \textcolor{tierseg}{\rule{1.1ex}{1.1ex}}~Dice ($\uparrow$) & $\mathbf{0.63}$ & $0.39$ & $\mathbf{0.71}$ & $\mathbf{0.69}$ & $0.26$ & $\mathbf{0.69}$ & $0.52$ & $0.26$ & $0.54$ & $\mathbf{0.71}$ & $0.36$ & $\mathbf{0.70}$ & $\mathbf{0.69}$ & $0.39$ & $\mathbf{0.69}$ & $\mathbf{0.63}$ & $0.17$ & $\mathbf{0.62}$ \\
  \textcolor{tierpheno}{\rule{1.1ex}{1.1ex}}~CD KID ($\downarrow$) & $43.62$ & $36.80$ & $36.76$ & $31.30$ & $34.94$ & $27.11$ & $4.22$ & $16.13$ & $\mathbf{3.58}$ & $24.96$ & $39.36$ & $25.83$ & $19.97$ & $19.86$ & $18.64$ & $\mathbf{1.85}$ & $8.82$ & $2.14$ \\
  \textcolor{tierpheno}{\rule{1.1ex}{1.1ex}}~GLCM+ cos ($\uparrow$) & $0.26$ & $0.08$ & $0.28$ & $0.28$ & $0.11$ & $0.30$ & $0.30$ & $0.13$ & $\mathbf{0.33}$ & $0.35$ & $0.10$ & $0.36$ & $0.38$ & $0.15$ & $0.39$ & $0.51$ & $0.11$ & $\mathbf{0.55}$ \\
\midrule
\multicolumn{19}{l}{\textbf{ER}} \\
  \textcolor{tierpixel}{\rule{1.1ex}{1.1ex}}~si-SSIM ($\uparrow$) & $\mathbf{0.36}$ & $0.13$ & $\mathbf{0.36}$ & $\mathbf{0.38}$ & $0.15$ & $\mathbf{0.37}$ & $0.23$ & $0.13$ & $0.23$ & $\mathbf{0.53}$ & $0.37$ & $\mathbf{0.53}$ & $\mathbf{0.53}$ & $0.39$ & $\mathbf{0.53}$ & $0.42$ & $0.33$ & $0.40$ \\
  \textcolor{tierpixel}{\rule{1.1ex}{1.1ex}}~SPCC ($\uparrow$) & $\mathbf{0.88}$ & $0.47$ & $\mathbf{0.88}$ & $\mathbf{0.90}$ & $0.53$ & $\mathbf{0.88}$ & $0.79$ & $0.51$ & $0.78$ & $\mathbf{0.82}$ & $0.68$ & $\mathbf{0.83}$ & $\mathbf{0.83}$ & $0.72$ & $\mathbf{0.83}$ & $0.71$ & $0.70$ & $0.73$ \\
  \textcolor{tierseg}{\rule{1.1ex}{1.1ex}}~Dice ($\uparrow$) & $\mathbf{0.26}$ & $0.01$ & $\mathbf{0.26}$ & $\mathbf{0.27}$ & $0.03$ & $\mathbf{0.27}$ & $\mathbf{0.23}$ & $0.10$ & $\mathbf{0.23}$ & $\mathbf{0.14}$ & $0.10$ & $\mathbf{0.15}$ & $\mathbf{0.12}$ & $0.10$ & $\mathbf{0.14}$ & $\mathbf{0.14}$ & $\mathbf{0.15}$ & $\mathbf{0.14}$ \\
  \textcolor{tierpheno}{\rule{1.1ex}{1.1ex}}~CD KID ($\downarrow$) & $36.64$ & $40.07$ & $37.04$ & $32.13$ & $38.81$ & $26.63$ & $3.16$ & $16.86$ & $\mathbf{2.38}$ & $28.33$ & $23.82$ & $28.61$ & $19.97$ & $19.07$ & $22.04$ & $\mathbf{4.26}$ & $5.80$ & $4.73$ \\
  \textcolor{tierpheno}{\rule{1.1ex}{1.1ex}}~GLCM+ cos ($\uparrow$) & $0.17$ & $0.01$ & $0.18$ & $0.18$ & $0.11$ & $0.19$ & $0.41$ & $0.15$ & $\mathbf{0.43}$ & $0.13$ & $0.03$ & $0.16$ & $0.17$ & $0.06$ & $0.19$ & $0.21$ & $0.06$ & $\mathbf{0.22}$ \\
\midrule
\multicolumn{19}{l}{\textbf{Mitochondria}} \\
  \textcolor{tierpixel}{\rule{1.1ex}{1.1ex}}~si-SSIM ($\uparrow$) & $\mathbf{0.54}$ & $0.32$ & $\mathbf{0.54}$ & $\mathbf{0.58}$ & $0.36$ & $\mathbf{0.57}$ & $0.43$ & $0.31$ & $0.42$ & $\mathbf{0.73}$ & $0.57$ & $\mathbf{0.74}$ & $\mathbf{0.76}$ & $0.60$ & $\mathbf{0.76}$ & $0.64$ & $0.52$ & $0.66$ \\
  \textcolor{tierpixel}{\rule{1.1ex}{1.1ex}}~SPCC ($\uparrow$) & $0.89$ & $0.53$ & $\mathbf{0.90}$ & $\mathbf{0.92}$ & $0.65$ & $\mathbf{0.92}$ & $0.85$ & $0.55$ & $0.84$ & $\mathbf{0.89}$ & $0.79$ & $\mathbf{0.90}$ & $\mathbf{0.92}$ & $0.82$ & $\mathbf{0.92}$ & $0.81$ & $0.75$ & $0.84$ \\
  \textcolor{tierseg}{\rule{1.1ex}{1.1ex}}~Dice ($\uparrow$) & $\mathbf{0.43}$ & $0.07$ & $\mathbf{0.43}$ & $\mathbf{0.47}$ & $0.19$ & $\mathbf{0.46}$ & $0.32$ & $0.12$ & $0.30$ & $\mathbf{0.37}$ & $0.22$ & $\mathbf{0.38}$ & $\mathbf{0.38}$ & $0.25$ & $\mathbf{0.39}$ & $0.25$ & $0.20$ & $0.27$ \\
  \textcolor{tierpheno}{\rule{1.1ex}{1.1ex}}~CD KID ($\downarrow$) & $30.45$ & $37.70$ & $31.25$ & $35.40$ & $32.57$ & $28.05$ & $\mathbf{3.13}$ & $21.84$ & $6.04$ & $30.24$ & $33.81$ & $29.51$ & $32.08$ & $33.59$ & $27.13$ & $6.72$ & $12.06$ & $\mathbf{5.68}$ \\
  \textcolor{tierpheno}{\rule{1.1ex}{1.1ex}}~GLCM+ cos ($\uparrow$) & $0.13$ & $0.06$ & $0.13$ & $0.12$ & $0.12$ & $0.14$ & $\mathbf{0.66}$ & $0.19$ & $0.56$ & $0.15$ & $0.11$ & $0.16$ & $0.20$ & $0.13$ & $0.20$ & $0.31$ & $0.04$ & $\mathbf{0.38}$ \\
\bottomrule
\end{tabular}%
}
\end{table}
 
None of the five selected metrics scored the poor FNet3D OOD predictions in Fig.~\ref{fig:prediction-overview} at least as favorably as the in-distribution model on the same test images. Spectral PCC, Dice, and si-SSIM also never did so across test sets, whereas GLCM+ cos did so on one of four organelles and CELL-DINO KID on two~(App.~\ref{sec:metric_reliability_appendix}). We therefore compared models within each test set and interpreted the metric panel alongside the prediction images.

\subsection{Joint training recovers cross-dataset performance}
\label{sec:cross-cell-type-generalization}

We next tested how models trained on one dataset generalized to the other cell type. Their prediction quality decreased OOD, although VSCyto3D, Pix2Pix3D, and CELL-Diff3D predictions usually degraded less than FNet3D predictions, and transfer from A549 to iPSC remained difficult for all four organelles. Jointly trained models largely closed this gap, approaching the in-distribution Spectral PCC and Dice scores on both test sets~(Tab.~\ref{tab:mr-pix2pix}; App.~Tab.~\ref{tab:training-matrix-three-tiers}).

We also tested membrane and mitochondria predictions on HEK293T images acquired with the same Mantis microscope as the A549 images, although the fluorescence markers differed between the two cell types. The decrease in OOD prediction quality was smaller for transfer from A549 to HEK293T than from A549/Mantis to iPSC/confocal, and jointly trained models performed best on HEK293T for both organelles~(App.~\ref{sec:hek-ablation-detail}).

\subsection{Regression and generative models are suitable for different downstream tasks}
\label{sec:regression-vs-generative}

We next compared regression (VSCyto3D), adversarial (Pix2Pix3D), and generative (CELL-Diff3D) models across the three evaluation tiers for every organelle, training set, and test set (Tab.~\ref{tab:mr-pix2pix}). Pix2Pix3D performed comparably to VSCyto3D, and one of the two had the best Spectral PCC and Dice in most settings, whereas CELL-Diff3D had the best CELL-DINO KID in every setting. This result held in the other learned feature spaces, where CELL-Diff3D had the lowest KID in 22--24 of 24 settings, although the extractors ranked the remaining models differently~(App.~\ref{sec:feature-space-agreement}).

To test whether this trade-off arises from the training objective rather than from architecture or capacity, we compared three models that share the UNetViT3D generator architecture and differ only in their training objective (App.~\ref{sec:matched-generator-ablation}). As in Tab.~\ref{tab:mr-pix2pix}, the regression and adversarial models performed better on pixel and segmentation metrics, whereas CELL-Diff3D better matched the population phenotype distribution. GLCM+ cos, although a paired measure, also favored CELL-Diff3D in all eight organelle and test-set groups of Tab.~\ref{tab:mr-pix2pix}, which we attribute to the smoother texture of regression and adversarial predictions falling outside the experimental GLCM+ distribution. We therefore recommend regression and adversarial models for organelle localization, segmentation, counting, and tracking, and generative models for assessing population morphology and uncertainty.

\subsection{Generative predictions best preserve perturbation responses}
\label{sec:perturbation-response-results}

\begin{figure}[ht]
  \centering
  \includegraphics[width=0.9\textwidth]{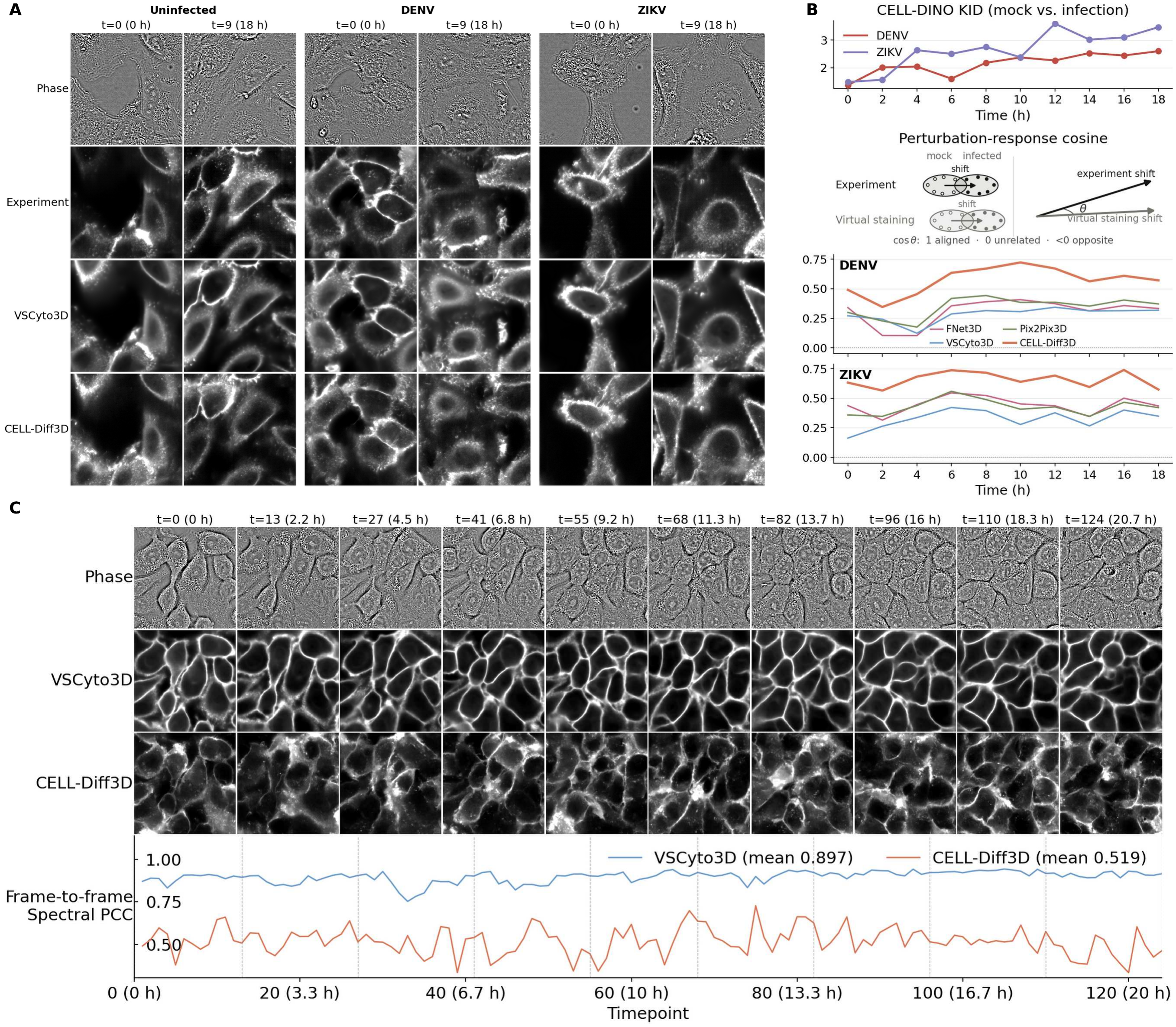}
\caption{\textbf{A549 predictions across perturbation and time.} Membrane predictions from A549-trained models. \textbf{(A)} One field per condition at the first and last paired timepoints. \textbf{(B)} Experimental separation between Mock and infected cells and the agreement between predicted and experimental mean-embedding shifts across the 2-hour grid. \textbf{(C)} Predictions on one unpaired field acquired every 10 min, with Spectral PCC between consecutive predicted volumes below.}
  \label{fig:a549-robustness}
\end{figure}

We compared membrane predictions at the first and last paired timepoints for mock-infected controls (Mock) and the DENV and ZIKV conditions (Fig.~\ref{fig:a549-robustness}A). The experimental phenotype distributions of Mock and infected cells separated more strongly over the A549 time course. To test whether virtual stains preserve this change, we measured the shift from Mock to infected cells in the mean single-cell embedding. We then compared the direction of this shift between predicted and experimental fluorescence. Generative CELL-Diff3D agreed most closely with the experimental response under both DENV and ZIKV~(Fig.~\ref{fig:a549-robustness}B). We describe the comparison and report the full results in App.~\ref{sec:temporal-perturbation-appendix}.

\subsection{Prediction quality and consistency vary over time}
\label{sec:stability-perturbation-time}

We evaluated two temporal properties of virtual staining predictions. The first, \emph{temporal robustness}, is the stability of prediction accuracy as the perturbation progresses, which we measured against paired fluorescence at each timepoint of the 2-hour A549 series under mock and infected conditions. Across experimental conditions and types of metrics, the ranking of models changed over time points (App.~Tabs.~\ref{tab:temporal-paired-robustness} and~\ref{tab:temporal-order-retention}).
 The second, \emph{temporal consistency}, is the agreement between successive predictions; it requires dense sampling but no fluorescence, so we measured it on the 10-minute unpaired label-free movie. CELL-Diff3D predictions were less consistent between adjacent frames than the regression and adversarial predictions across all four organelles (Fig.~\ref{fig:a549-robustness}C; App.~Figs.~\ref{fig:temporal-frame-to-frame} and~\ref{fig:movie-selected-organelles}). Because the movie has no paired fluorescence target, these comparisons measure self-consistency rather than temporal accuracy. We report the complete analyses in App.~\ref{sec:temporal-perturbation-appendix}.

\section{Discussion}
\label{sec:discussion}

Our results show that virtual staining quality depends on the training data, the downstream task, and when predictions are evaluated. OOD failures reveal weaknesses in commonly used metrics, motivating evaluation across pixel fidelity, segmentation, and phenotype similarity. Some metrics miss these failures on the same images; others shift in value between test sets, so scores should be compared within a test set, consistent with an independent analysis of Cell Painting data~\citep{li2026doubting}. Joint training recovers performance on the two represented datasets. Regression and adversarial models more accurately reproduce individual organelles, whereas generative predictions better preserve population phenotype distributions and perturbation responses. Better population agreement, however, does not imply more consistent predictions over time. Together, these findings show why a single reconstruction score or timepoint is insufficient to judge virtual stains for live-cell profiling. The DynaCell evaluation suite provides a shared basis for choosing models according to the biological measurement they need to support. Beyond model selection, the suite can test whether a new metric separates known failure modes, such as OOD collapse and target photobleaching, from genuine prediction errors. A virtual stain can be checked for whether it preserves infection responses before it replaces fluorescence in a perturbation screen. Models trained on one cell type and microscope can be tested on another. A densely sampled paired release would extend these uses to temporal consistency and time-aware models.

\section{Limitations and Future Work}
\label{sec:limitations}

Our results support comparisons of paired 3D reconstruction, segmentation, phenotype distributions, and performance over time for the cell types, microscopes, organelles, and perturbations studied here. Important extensions include varying cell type and microscope independently, adding manual segmentation labels and a broader range of perturbations, and collecting densely sampled paired movies. The pretrained phenotype extractors use 2D maximum projections because most available deep feature extractors are 2D. Only GLCM+ is computed in 3D. We do not evaluate time-aware models: frames on the 2-hour grid do not form trajectories they could exploit, and the one densely sampled movie has no paired fluorescence. We therefore leave temporal stability and time-aware training objectives to future work. Dense paired movies are limited by the photobleaching and phototoxicity of correlative light-sheet imaging, and collecting them is the next step. Training and evaluating the full 3D benchmark is computationally costly. A 2D variant, which trains each model family on single in-focus planes and scores it at one focus plane per field of view, is about 27$\times$ cheaper to train and reproduces the coarse model tiers but often not the best-ranked family (App.~\ref{sec:twod-benchmark}). A lite test split with half the test volumes reverses none of the model orderings that the full benchmark resolves (App.~\ref{sec:lite-split}). App.~\ref{sec:limitations-detail} discusses these limitations and planned baseline extensions in detail.

\section{Data and code availability}
The dataset, Croissant metadata, and full license terms are described in App.~\ref{sec:documentation-license}. The DynaCell evaluation suite is available under BSD-3-Clause at \url{https://github.com/Biohub/VisCy/tree/main/applications/dynacell}.

\section*{Acknowledgments}
We thank Jason Swedlow, Cathy Stolitzka, Maadhav Kothari, Kevin Liu, and Emanuele Bezzi for enabling data sharing through AWS Open Data. We thank Alan Lowe and Markus Marks for suggesting the randomly initialized models as a baseline. This work was supported by the Biohub Imaging Residency Program, Biohub San Francisco and its donors Priscilla Chan and Mark Zuckerberg, and NIH R01GM131641 (B.H.).

\bibliographystyle{unsrtnat}

\newpage
\appendix
\addtocontents{toc}{\protect\setcounter{tocdepth}{2}}

\section*{Appendix}

\etocsettocstyle{\subsection*{Table of Contents}}{}
\etocsetnexttocdepth{subsection}
\tableofcontents
\newpage

\section{Dataset details and release}
\label{sec:documentation-license}

 OME-Zarr stores are hosted through AWS Open Data at \url{s3://dynacell/v1} and described at \url{https://registry.opendata.aws/dynacell/}. Trained checkpoints for all baselines are at \url{s3://dynacell/v1/models/}, with a manifest at \url{https://dynacell.s3.us-west-2.amazonaws.com/v1/models/README.md}. A demo subsample for inspecting the data without downloading the full release is available at \url{https://dynacell.s3.us-west-2.amazonaws.com/v1/demo/dynacell_a549_demo.zip} (App.~\ref{sec:reviewer-sample}).
 
\subsection{A549 acquisition intervals and split composition}
\label{sec:a549-split-details}

The original A549 time-lapse series were acquired at intervals of 10, 30, or 120 min. We subsample all plates onto a common 2-hour grid spanning approximately 5--23 hours post-infection (hpi), yielding 10 timepoints for most FOVs. The training set contains 118 FOVs: 30 each for nucleus, membrane, and mitochondria, and 28 for ER. The test set contains 12 FOVs per organelle--condition combination across four organelles and three conditions, giving 144 FOVs. Each released test volume has dimensions $48\times640\times960$ (ZYX), and the test set contains approximately 17,000 cells across the four organelles. Test FOVs come from well positions absent from the training set. One additional untreated FOV is retained at its native 10-min interval for unpaired temporal-consistency evaluation (Sec.~\ref{sec:stability-perturbation-time}).

\subsection{Comparison with related resources}
Tab.~\ref{tab:dataset-comparison} compares the A549 and iPSC datasets and HEK293T evaluation data used by the DynaCell evaluation suite with other paired label-free and fluorescence resources.

\begin{table*}[h]
\centering
\scriptsize
\caption{\textbf{Paired label-free/fluorescence resources on the axes used by the DynaCell evaluation suite.} The three organelle columns are nested counts: available targets, targets imaged over time, and time-lapse targets imaged under perturbation with a matched control. Tiers are pixel (px), segmentation (seg), and phenotype (phe) metrics applied to the predicted stain. A dash is a measured absence; n/r is not reported. Pair counts use one label-free--target pair as the common unit. The A549 and iPSC datasets and HEK293T evaluation data contain 6,998 paired 3D volumes, of which 4,822 are scored by the benchmark protocol. $\dagger$ Mantis counts come from the deposited stores; $\ddagger$ JUMP reports profiling rather than a virtual-staining score; $\S$ and $\P$ denote non-like-for-like phenotype and viability scores, respectively.}
\label{tab:dataset-comparison}
\resizebox{\textwidth}{!}{%
\begin{tabular}{lcccccccccc}
\toprule
\textbf{Dataset} & \textbf{3D} & \textbf{Live} & \textbf{Perturbation} & \textbf{Organelles} & \textbf{\ldots{}as time-lapse} & \textbf{\ldots{}under perturbation} & \textbf{Cell types} & \textbf{Protocol} & \textbf{Tiers} & \textbf{Paired examples} \\
\midrule
Light My Cells (2026) & no & yes & no & 4 & 0 & 0 & n/r & yes & px, phe \S & 4,602 (2D) \\
Trizna et al. (2023) & no & yes & drug & 0 & 0 & 0 & 1 & yes & seg \P & 504 (2D) \\
VIRVS (2025) & no & mixed & 5 viruses & 0 & 0 & 0 & 3 & yes & px, seg & 3,820 (2D) \\
JUMP Cell Painting (2023) & no & no & chem + gene & 8 & 0 & 0 & 1--2 & no & --- $\ddagger$ & \textasciitilde{}1.9 M (2D) \\
Ounkomol et al. (2018) & yes & yes & no & 12 & 0 & 0 & 4 & no & px & \textasciitilde{}420 (3D) \\
Guo et al. QLIPP (2020) & yes & no & no & 3 & 0 & 0 & tissue & no & px & 320 (3D) \\
Mantis --- A549 (2024) & yes & yes & no & 2 & 2 & 0 & 1 & no & --- & 960 (3D) $\dagger$ \\
Mantis --- HEK OpenCell (2024) & yes & yes & OC43 & 15 & 6 & \textbf{0} $\dagger$ & 1 & no & seg & 1,792 (3D) $\dagger$ \\
Viana et al. WTC-11 (2023) & yes & yes & no & 25 & 0 & 0 & 1 & no & --- & 54,300 (3D) \\
\textbf{A549 + iPSC + HEK293T eval.} & \textbf{yes} & \textbf{yes} & \textbf{ZIKV / DENV} & \textbf{4} & \textbf{4} & \textbf{4} & \textbf{3} & \textbf{yes} & \textbf{px, seg, phe} & \textbf{6,998 (3D)} \\
\bottomrule
\end{tabular}%
}
\end{table*}
 
\subsection{HEK293T evaluation data: acquisition details}
\label{sec:hek-evaluation-data}

HEK293T serves as a held-out evaluation set acquired on the Mantis microscope. We evaluate four $64\times960\times1184$ ZYX membrane volumes and four mitochondria volumes (approximately 120 cells per organelle), selected from 8 of 80 positions in S-BIAD1287. HEK293T and A549 share the microscope but use different markers (KRAS versus CAAX; TOMM70A versus TOMM20), so the measured gap includes both cell-type and marker effects.

\subsection{Broader impact and responsible use}
\label{sec:broader-impact}

Reducing phototoxicity and spectral crowding in live-cell imaging remains a challenge in cell biology and drug discovery: each additional fluorescent probe increases the light dose and narrows the spectral window available to other probes.
Virtual staining can alleviate these constraints by predicting organelle-specific fluorescence from a single label-free acquisition, allowing continuous observation without additional genetic modification or fluorescent dyes.
The DynaCell evaluation suite assesses this capability on living, perturbed cells across complementary measures of image quality and biological utility.

Responsible use requires recognizing that virtually stained images are model predictions, not direct measurements.
The virtual stains evaluated here have been validated only on the cell lines, perturbation conditions, and microscope platforms represented in this benchmark; their accuracy in other contexts has not been established.
Results from the DynaCell evaluation suite must not be used for clinical or diagnostic decision-making. Its perturbation-response scores compare research images and are not clinical assays.
Cell-line provenance (A549: ATCC CCL-185, an immortalized human lung adenocarcinoma line; WTC-11: Coriell GM25256, a hiPSC line with established consent for research use) is documented in~App.~\ref{sec:datasheet}.

\subsection{License, hosting, and long-term maintenance}
\label{sec:license-hosting}

\paragraph{Dataset licenses.}
The A549 and iPSC datasets have different licenses.
The A549 dataset (acquired on Mantis at Biohub SF) is released under the Creative Commons Attribution 4.0 International License (CC-BY-4.0).
The iPSC dataset is a curated, preprocessed derivative of the Allen Institute WTC-11 hiPSC dataset~\citep{aics2018hipsc,viana2023integrated}. It is redistributed under the upstream Allen Institute Terms of Use (\url{https://www.allencell.org/terms-of-use.html}), which permit use, redistribution, and creation of derivative works for research or other noncommercial purposes. Commercial redistribution requires written permission from the Allen Institute.
Users of the iPSC dataset must comply with the Allen Institute Terms of Use and cite the original dataset under the Allen Institute Citation Policy (\url{https://www.allencell.org/citation-policy.html}); the A549 dataset carries no such restriction.
Benchmark code, preprocessing pipelines, and evaluation scripts are released under the BSD-3-Clause License at \url{https://github.com/Biohub/VisCy/tree/main/applications/dynacell}.

\paragraph{Author responsibility statement.}
The authors confirm the licenses declared above and accept full responsibility for any violation of rights.
Where the redistributed iPSC dataset is concerned, the authors have only the rights granted to them by the Allen Institute Terms of Use and pass through those rights, unmodified, to downstream users.

\paragraph{Hosting and persistent identifiers.}
OME-Zarr datasets are hosted through the AWS Open Data Sponsorship Program at \texttt{s3://dynacell/v1/} and are publicly accessible without registration. A human-readable manifest with direct HTTPS URLs, per-file sizes, a version log, and license documentation is available at \href{https://dynacell.s3.us-west-2.amazonaws.com/v1/README.md}{https://dynacell.s3.us-west-2.amazonaws.com/v1/README.md}.
Benchmark code is versioned on GitHub at \url{https://github.com/Biohub/VisCy/tree/main/applications/dynacell} and is regularly maintained. It will be released through PyPI upon publication.

\paragraph{Long-term maintenance.}
Biohub SF commits to maintaining access through AWS Open Data and the GitHub code repository over the long term, with no plans for removal.
The named technical contact is Shalin B.\ Mehta (\texttt{shalin.mehta@biohub.org}); errors and inconsistencies can be reported via GitHub Issues at \url{https://github.com/Biohub/VisCy}.

\paragraph{Datasheet and machine-readable metadata.}
A datasheet following Gebru et al. appears in App.~\ref{sec:datasheet}; Croissant metadata with responsible-AI fields accompanies the release.

\subsection{Demo subsample}
\label{sec:reviewer-sample}
A demo subsample for inspecting the data and reproducing the per-tier metrics without downloading the full release is available at \url{https://dynacell.s3.us-west-2.amazonaws.com/v1/demo/dynacell_a549_demo.zip} (745\,MiB). It holds one A549 field of view under mock and ZIKV infection at two timepoints as OME-Zarr, with the paired membrane fluorescence, the joint-trained FNet3D, VSCyto3D, and CELL-Diff3D predictions, and their segmentations, together with single-cell embeddings, demo scripts, and a README. Time-lapse movies that illustrate the temporal kinetics in Fig.~\ref{fig:a549-robustness}C and other demonstrations are available in the \href{https://dynacell.s3.us-west-2.amazonaws.com/v1/supplementary/}{data-release S3 bucket}. The data and evaluation tutorial appear in the README at \href{https://dynacell.s3.us-west-2.amazonaws.com/v1/README.md}{https://dynacell.s3.us-west-2.amazonaws.com/v1/README.md}.

\FloatBarrier

\section{Metrics and validation}

\subsection{Spectral PCC: motivation and computation}
\label{sec:spectral_pcc}

\paragraph{Motivation.} Even with the Mantis light-sheet microscope, the endogenously tagged A549 cells photobleach over the multihour movie. The result is a frequency-dependent SNR: low-frequency bands carry organelle structure with high SNR, while high-frequency bands are dominated by Poisson photon noise and acquisition artifacts that a virtual-staining model cannot predict. Standard PCC weights every Fourier component equally, so a prediction that suppresses noise is penalized for not matching the target noise. As bleaching pushes more bands below the noise floor, this penalty can produce a declining PCC curve that partly reflects target SNR rather than prediction quality.

\paragraph{Approach.}
We propose Spectral PCC, a weighted Pearson correlation in the Fourier domain. The weights estimate which target frequencies carry signal. Given a prediction $\hat{\rvx}$ and target $\rvx$ on a grid with voxel spacing $\Delta$, we compute the score as follows:
\begin{enumerate}[noitemsep]
  \item Subtract the mean from $\hat{\rvx}$ and $\rvx$, then apply a Tukey window to suppress edge ringing.
  \item Compute the FFTs $F_{\hat{\rvx}}(\rvk)$ and $F_{\rvx}(\rvk)$.
  \item Form the radial power spectrum of the target by azimuthally averaging $|F_{\rvx}(\rvk)|^2$ into bins indexed by $|\rvk|$, giving $P(k)$.
  \item Estimate the noise floor $N$ as the mean power in the highest-frequency $20\%$ of bins, exploiting the white-noise assumption that pure Poisson noise contributes uniform power across frequency.
  \item Define per-bin weights $w(k) = \max(0,\,P(k) - N)$ and normalize so that $\max_k w(k) = 1$.
  \item Suppress the lowest few bins via a hard zero (DC, slow background drift, autofluorescence, illumination shading) or a soft cosine taper.
  \item For each Fourier voxel, use the weight $w(|\rvk|)$ of its radial bin and compute
  \begin{equation}
    \mathrm{SpectralPCC}(\hat{\rvx}, \rvx) =
    \frac{\sum_{\rvk} w(|\rvk|)\,\Re\bigl\{F_{\hat{\rvx}}(\rvk)\,F_{\rvx}^{*}(\rvk)\bigr\}}
         {\sqrt{\sum_{\rvk} w(|\rvk|)\,|F_{\hat{\rvx}}(\rvk)|^2 \,\cdot\, \sum_{\rvk} w(|\rvk|)\,|F_{\rvx}(\rvk)|^2}}.
    \label{eq:spectral_pcc}
  \end{equation}
\end{enumerate}
The default high-frequency-tail estimator is a simple scalar choice. It assumes that the target signal spectrum has decayed below noise by the highest $20\%$ of bins. This assumption appears to hold for these diffraction-limited data but should be revisited for super-resolution or strongly band-limited imaging.

\paragraph{Properties.}
The metric inherits four properties from this construction. First, $\mathrm{SpectralPCC}\in[-1,1]$, with $1$ for identical images and $-1$ for perfect anti-correlation. Second, when the noise floor is zero and the target spectrum is broadband, all weights collapse to a constant and the metric reduces to standard PCC by Parseval's theorem. Third, when high-frequency bands are noise-dominated, their weights vanish, so the score is invariant to whether the prediction reproduces that noise. Fourth, the noise floor is estimated for each image rather than fixed, so the metric tracks the changing reliability of the target across a time series instead of conflating it with prediction quality.

\paragraph{Hyperparameters.}
We use a radial bin width of one Fourier index, a tail fraction of $0.2$ for noise-floor estimation, a Tukey apodization window, and a low-frequency exclusion of the first three bins to prevent DC and slow background drift from inflating the score. These match the configuration shipped with the released evaluation pipeline.

\begin{figure}[htbp]
  \centering
  \includegraphics[width=0.92\linewidth]{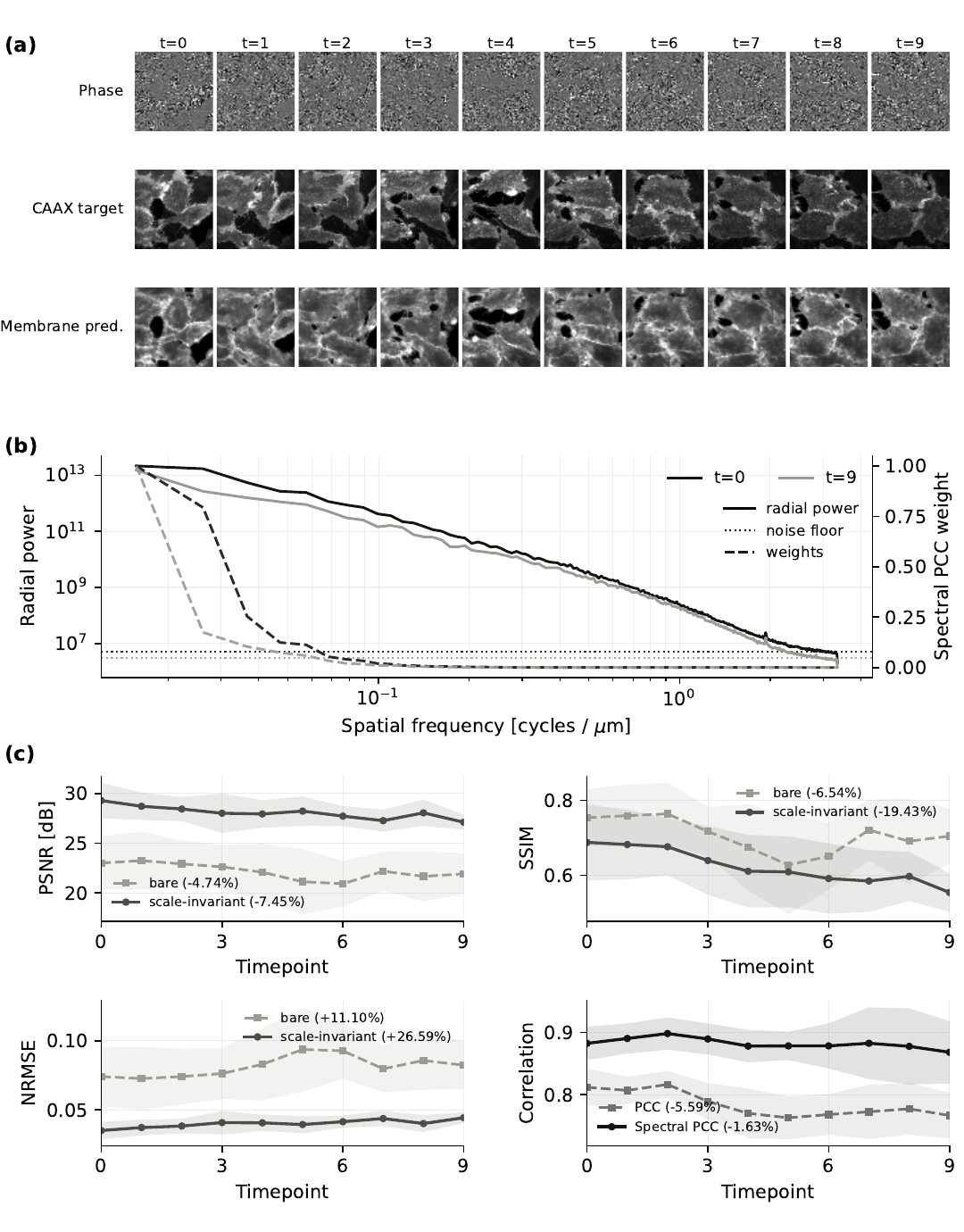}
  \caption{\textbf{Pixel metrics under photobleaching on the A549 Mock test set.}
  \textbf{(a)} Phase input, CAAX fluorescence target, and FNet3D membrane prediction over one ten-frame FOV, shown on fixed row-wise intensity scales.
  \textbf{(b)} Radial power spectra of the $t{=}0$ and $t{=}9$ frames from the same FOV (solid); per-image noise-floor estimates as horizontal dotted lines (mean of the highest 20\% of radial bins); and per-frequency Spectral PCC weights as dashed curves on the right axis. Bleaching lowers signal power below about $1$~cycle/$\mu$m while leaving the high-frequency noise floor nearly unchanged. The weights are nonzero only where the spectrum exceeds the noise floor and decrease slightly as $t$ increases.
  \textbf{(c)} Per-timepoint conventional and scale-invariant PSNR, SSIM, and NRMSE, together with PCC and Spectral PCC, for jointly trained FNet3D membrane predictions (mean $\pm$ one SD across 12 FOVs). Parenthetical values give the first-to-last relative change; Spectral PCC changes least among the correlation and scale-invariant structural scores.}
  \label{fig:spectral-pcc-bleaching}
\end{figure}

\paragraph{Validation with synthetic and A549 data.}
We validated the metric on a synthetic phantom of fluorescent beads convolved with a physically accurate widefield optical transfer function and corrupted with Poisson noise under exponential bleaching. In this simulation, the underlying signal is fixed and only the noise floor changes. Standard PCC decreases as the bleaching SNR falls, while Spectral PCC remains close to unity, correctly indicating that the prediction has not changed. The same pattern occurs empirically on the A549 Mantis test set (Fig.~\ref{fig:spectral-pcc-bleaching}): a membrane time course visibly photobleaches over ten timepoints (mean in-cell target intensity falls by $17.6\%$), lowering the radial power spectrum while the noise floor stays nearly constant. Across 12 Mock-condition FOVs, Spectral PCC for membrane predictions from jointly trained FNet3D decreases by only $1.63\%$ from $t{=}0$ to $t{=}9$. Over the same interval, PCC decreases by $5.59\%$ and si-SSIM by $19.43\%$, while si-NRMSE increases by $26.59\%$. These trajectories support reporting intensity-sensitive and scale-invariant scores alongside Spectral PCC.

This robustness reduces sensitivity to high-frequency degradation. In a controlled corruption ladder, we normalize each score change by the median gap between the best and worst in-distribution models. A lateral Gaussian blur with $\sigma=4$ voxels changes Spectral PCC by only $0.01$ of this between-model spread, compared with $3.12$ for si-SSIM. We therefore use si-SSIM and FSC, rather than Spectral PCC alone, when interpreting resolution.

\FloatBarrier
\subsection{Nuclear-segmenter comparison}
\label{sec:segmenter-ablation}

\begin{figure}[htbp]
  \centering
  \includegraphics[width=0.92\linewidth]{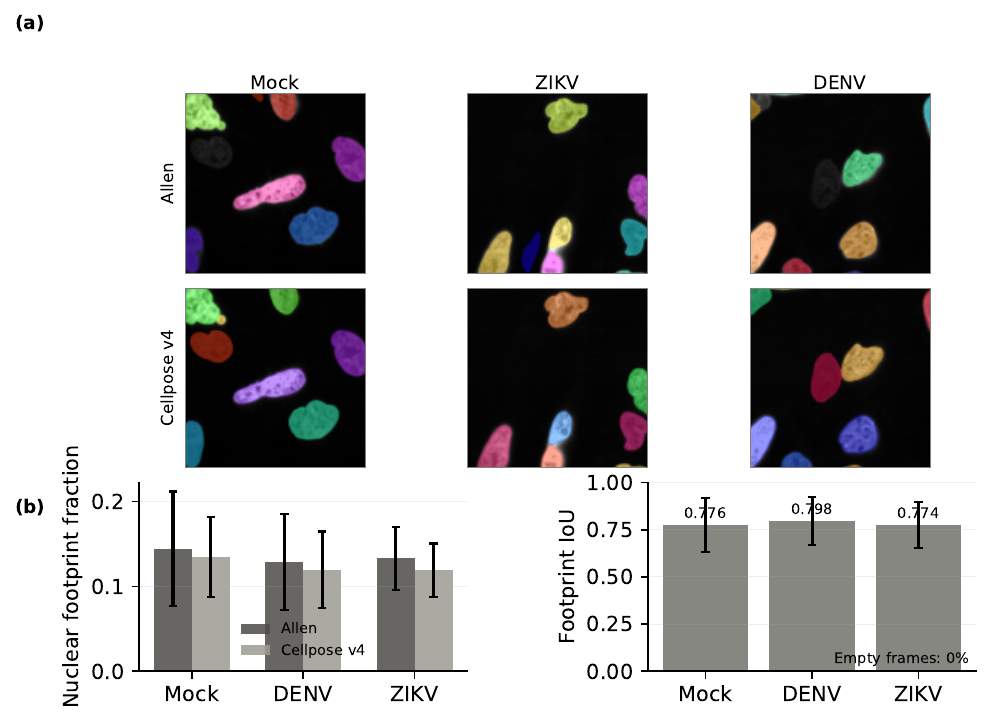}
  \caption{\textbf{Allen--Cellpose nuclear-segmenter comparison on A549.} \textbf{(a)} Representative target-fluorescence masks for Mock, ZIKV, and DENV at the last timepoint. \textbf{(b)} Nuclear footprint fraction and Allen--Cellpose footprint IoU across the full paired A549 test series (mean $\pm$ 1 SD over FOVs). Both methods shift comparably under infection, and Cellpose yields no empty frames.}
  \label{fig:segmenter-comparison}
\end{figure}

The Allen Cell and Structure Segmenter and Cellpose v4 cpdino-ViT-L produce similar target-nucleus footprints across all three A549 conditions (Fig.~\ref{fig:segmenter-comparison}). Their mean footprint IoU is $0.776$, $0.798$, and $0.774$ for Mock, DENV, and ZIKV, respectively, with no empty Cellpose frames. Relative to Mock, the segmented footprint decreases by approximately $9\%$ for Allen and $11\%$ for Cellpose under infection. These matched changes do not support an infection-specific failure of either backend; the choice of Cellpose reflects visual inspection, not validation against manual annotations.

\FloatBarrier
\subsection{Computation of feature similarities}
\label{sec:feat_simi}

\begin{figure}[htbp]
  \centering
  \includegraphics[width=\textwidth]{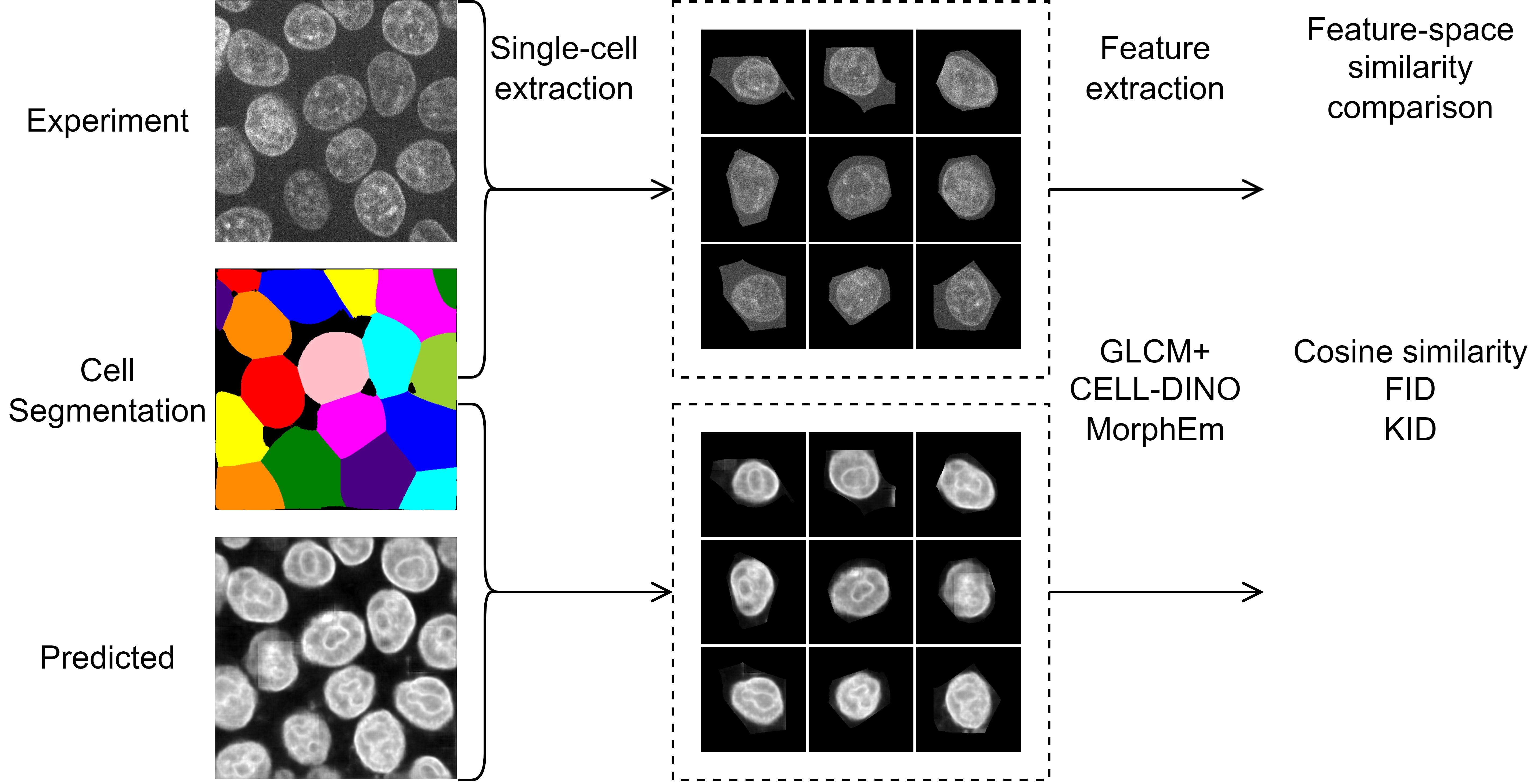}
  \caption{\textbf{Computation of single-cell feature metrics.}
  Individual cells are segmented from VSCyto3D nucleus and membrane predictions using Cellpose; the same masks are applied to every predicted and experimental fluorescence pair. GLCM+ computes 22 handcrafted 3D features from masked volumes. DINOv3, DynaCLR, CELL-DINO, and MorphEm encode 2D cell crops from a maximum projection of the target-defined five-plane focus slab. Feature agreement is quantified by paired cosine similarity and distributional FID and KID in each space.}
  \label{fig:feature_similarities}
\end{figure}

We compute feature-based metrics at the single-cell level to assess whether virtual stains preserve phenotypic information beyond pixel-level agreement. The pipeline consists of three steps: cell-region extraction, feature extraction, and feature-space comparison (Fig.~\ref{fig:feature_similarities}).

\paragraph{Cell-region extraction.}
We identify individual cell regions with a membrane-based segmentation workflow. Because membrane segmentation is often more robust on virtually stained membrane images than on experimental fluorescence~\citep{liu2025robust}, we predict the nucleus and membrane channels from the label-free input using pretrained VSCyto3D from Cytoland~\citep{liu2025robust}. The 3D membrane prediction is projected along the axial dimension, and Cellpose v4 with the cpdino-ViT-L model~\citep{pachitariu2025cellpose} produces 2D instance masks. We broadcast these masks along the axial dimension to define 3D cell regions.

\paragraph{Feature extraction.}
For each segmented cell, we extract features from the prediction and corresponding experimental fluorescence using five extractors. GLCM+ takes the full 3D volume and the broadcast 3D cell mask, producing 22 handcrafted features: seven GLCM/Haralick texture descriptors and 15 intensity-distribution and local-gradient/Laplacian descriptors. DINOv3, DynaCLR, CELL-DINO, and MorphEm operate on 2D images. For these models, we select a target-defined focus plane, maximum-project the five-plane slab centered on it, crop each cell with its 2D instance mask, and encode the crop. Before GLCM+ features are compared, near-constant and redundant features are removed once per organelle, using experimental cells only, pooled across the iPSC and A549 test sets. A feature is removed if one value is more than 20 times as frequent as any other or fewer than 1\% of its values are distinct, and from every pair of features with absolute Pearson correlation above 0.9, the one more correlated with the remaining features is removed. This keeps 9, 10, 12 and 12 of the 22 features for nucleus, membrane, ER and mitochondria, and every model is scored on the same subset. Each retained feature is then standardized with the mean and standard deviation of the experimental cells of the same test set, applied identically to predicted and experimental cells, so an offset or change of scale in a prediction remains visible in GLCM+ cos, FID and KID. Standardized values are clipped at $\pm 20$; this affects 2 of 23{,}970 experimental cells and leaves the model ranking unchanged for any larger bound. GLCM+ therefore penalizes predictions that do not reproduce the texture of the experimental images, including texture a model smooths away.

\paragraph{Paired feature similarity.}

Paired feature similarity measures cell-level agreement between a predicted image and its matched experimental image. For each image pair, we use the same cell segmentation masks to extract features from corresponding cells in the predicted and experimental fluorescence images.

Let \(\mathbf{f}^{\mathrm{pred}}_{b,i}\) and \(\mathbf{f}^{\mathrm{gt}}_{b,i}\) denote the feature vectors extracted from the predicted and experimental images for cell \(i\) in image pair \(b\). We compute the cosine similarity for each matched cell,
\begin{equation}
    s_{b,i}
    =
    \frac{
    \left\langle
    \mathbf{f}^{\mathrm{pred}}_{b,i},
    \mathbf{f}^{\mathrm{gt}}_{b,i}
    \right\rangle
    }{
    \left\|
    \mathbf{f}^{\mathrm{pred}}_{b,i}
    \right\|_2
    \left\|
    \mathbf{f}^{\mathrm{gt}}_{b,i}
    \right\|_2
    } .
\end{equation}
We summarize each image pair by the median cosine similarity across all matched cells in that pair,
\begin{equation}
    S_{\mathrm{cos}}^{(b)}
    =
    \operatorname{median}_{i}
    \left( s_{b,i} \right).
\end{equation}
The final paired feature-similarity score is the mean of these image-level scores across all matched image pairs,
\begin{equation}
    S_{\mathrm{cos}}
    =
    \frac{1}{B}
    \sum_{b=1}^{B}
    S_{\mathrm{cos}}^{(b)},
\end{equation}
where \(B\) is the number of image pairs in the test set. This metric evaluates whether each predicted image preserves the single-cell phenotypic representations of its paired experimental fluorescence image.

\paragraph{Distribution-level feature metrics.}

Distribution-level feature metrics compare the predicted and experimental feature distributions over the entire test dataset. Unlike paired feature similarity, these metrics do not compute a separate score for each image pair. Instead, we pool all single-cell features from the predicted images and all single-cell features from the experimental images across the full test set, then compute one distribution-level score for the dataset.

We compare the pooled predicted and experimental feature distributions using Fr\'echet Inception Distance (FID) and Kernel Inception Distance (KID). These metrics evaluate whether the virtual stains reproduce the population-level phenotype distribution in the experimental fluorescence images.

Let
\[
    \mathcal{F}_{p}
    =
    \left\{
    \mathbf{f}^{\mathrm{pred}}_{b,i}
    \right\}_{b=1,i=1}^{B,N_b},
    \qquad
    \mathcal{F}_{g}
    =
    \left\{
    \mathbf{f}^{\mathrm{gt}}_{b,i}
    \right\}_{b=1,i=1}^{B,N_b}
\]
denote the pooled predicted and experimental single-cell features from all image pairs in the test set, where \(N_b\) is the number of segmented cells in image pair \(b\). Let their empirical means be \(\boldsymbol{\mu}_{p}\) and \(\boldsymbol{\mu}_{g}\), and their covariance matrices be \(\boldsymbol{\Sigma}_{p}\) and \(\boldsymbol{\Sigma}_{g}\). FID is defined as
\begin{equation}
    \mathrm{FID}
    =
    \left\|
    \boldsymbol{\mu}_{p}
    -
    \boldsymbol{\mu}_{g}
    \right\|_2^2
    +
    \operatorname{Tr}
    \left(
    \boldsymbol{\Sigma}_{p}
    +
    \boldsymbol{\Sigma}_{g}
    -
    2
    \left(
    \boldsymbol{\Sigma}_{p}
    \boldsymbol{\Sigma}_{g}
    \right)^{1/2}
    \right).
\end{equation}
Lower FID indicates closer agreement between the predicted and experimental feature distributions under a Gaussian approximation.

KID estimates the squared maximum mean discrepancy (MMD) between the pooled predicted and experimental features using a third-degree polynomial kernel. We repeatedly sample equal-size subsets without replacement and average the unbiased MMD estimator. For predicted features \(\{\mathbf{x}_i\}_{i=1}^{s}\) and experimental features \(\{\mathbf{y}_j\}_{j=1}^{s}\) in one subset,
\begin{equation}
    \widehat{\mathrm{MMD}}_u^2
    =
    \frac{1}{s(s-1)}\sum_{i\ne i'} k(\mathbf{x}_i,\mathbf{x}_{i'})
    +
    \frac{1}{s(s-1)}\sum_{j\ne j'} k(\mathbf{y}_j,\mathbf{y}_{j'})
    -
    \frac{2}{s^2}\sum_{i=1}^{s}\sum_{j=1}^{s} k(\mathbf{x}_i,\mathbf{y}_j).
\end{equation}
KID is the mean of \(\widehat{\mathrm{MMD}}_u^2\) across the sampled subsets.
We use the polynomial kernel
\begin{equation}
    k(\mathbf{x},\mathbf{y})
    =
    \left(
    \frac{1}{d}
    \mathbf{x}^{\top}\mathbf{y}
    +
    1
    \right)^3,
\end{equation}
where \(d\) is the feature dimension. Lower KID indicates closer agreement between the predicted and experimental single-cell feature distributions. Because the estimator is unbiased, finite-sample values can be slightly negative; values near zero indicate no detectable difference.

\paragraph{Limit of polynomial-kernel KID.}
Polynomial-kernel MMD is the squared distance between distribution embeddings in the kernel feature space. A degree-three polynomial kernel compares moments only through order three, so low KID need not imply that a prediction covers the full experimental distribution. A predictor concentrated near common experimental features can therefore score well while missing rare or diverse phenotypes. We separate fidelity from diversity using Precision and Recall in App.~\ref{sec:precision_recall}. More generally, KID, FID, and Precision and Recall compare pooled distributions and cannot test whether each cell's prediction is correct; information gain scores per-cell predictive distributions directly~\citep{tonks2026proper}.

\FloatBarrier
\subsection{Agreement among phenotype feature extractors}
\label{sec:feature-space-agreement}

We compared KID across five feature extractors using predictions from the four main models in 24 matched settings: four organelles, three training sets (iPSC, A549, or both), and two test sets (iPSC and A549). Because KID has a different numerical scale in each feature space, we compare model rankings rather than raw values across extractors (Tab.~\ref{tab:kid-rank-consistency}).

\begin{table}[ht]
\centering
\definecolor{failsame}{RGB}{191,77,67}
\small
\caption{\textbf{Model ranking sensitivity to feature extractor.} Each extractor ranks the four main models by KID in 24 matched settings: four organelles, iPSC/A549/joint training, and iPSC/A549 testing. Mean rank uses 1 for the lowest KID and average ranks for ties. Wins count settings where sampled CELL-Diff3D has strictly lower KID than all three other models. A ranking is separated when every adjacent KID gap exceeds twice the combined resampling SD $\sqrt{s_a^2+s_b^2}$; this is a descriptive threshold, not a significance test. Mean Kendall $\tau_b$ averages ranking agreement with each of the four other extractors over all 24 settings (96 comparisons per extractor; 1 indicates identical rankings). \emph{OOD worse} counts strictly higher KID for OOD-trained models than for same-dataset-trained models on the indicated test set (four models $\times$ four organelles). The iPSC row compares A549 versus iPSC training; the A549 row compares iPSC versus A549 training. Joint training is excluded from these two rows.}
\label{tab:kid-rank-consistency}
\resizebox{\ifdim\width>\textwidth \textwidth\else\width\fi}{!}{%
\begin{tabular}{l c c c c c}
\toprule
 & \multicolumn{5}{c}{\textbf{Feature extractor}} \\
\cmidrule(lr){2-6}
\textbf{Statistic} & \textbf{GLCM+} & \textbf{DINOv3} & \textbf{DynaCLR} & \textbf{CELL-DINO} & \textbf{MorphEm} \\
\midrule
  \textbf{FNet3D mean rank} & $2.46$ & $3.33$ & $2.42$ & $3.33$ & $3.46$ \\
  \textbf{VSCyto3D mean rank} & $3.25$ & $3.21$ & $3.83$ & $3.38$ & $3.33$ \\
  \textbf{Pix2Pix3D mean rank} & $2.75$ & $2.38$ & $2.67$ & $2.29$ & $2.04$ \\
  \textbf{CELL-Diff3D mean rank} & $1.54$ & $1.08$ & $1.08$ & $1.00$ & $1.17$ \\
  \textbf{CELL-Diff3D wins} & $18/24$ & $22/24$ & $23/24$ & $24/24$ & $22/24$ \\
  \textbf{Separated rankings} & $12/24$ & $17/24$ & $17/24$ & $17/24$ & $19/24$ \\
  \textbf{Mean Kendall $\tau_b$} & $+0.39$ & $+0.55$ & $+0.49$ & $+0.53$ & $+0.55$ \\
  \textbf{OOD worse: iPSC test} & $12/16$ & $14/16$ & $8/16$ & $14/16$ & $14/16$ \\
  \textbf{OOD worse: A549 test} & $9/16$ & $10/16$ & $6/16$ & $11/16$ & $13/16$ \\
\bottomrule
\end{tabular}%
}
\end{table}
 
CELL-Diff3D predictions have the lowest KID among the four models in 22--24 of the 24 comparisons, depending on the learned feature extractor. With GLCM+ features, they have the lowest KID in 18 comparisons. The learned feature extractors differ mainly in how they rank the other three models. GLCM+ separates the full model ranking relative to KID resampling variation in 12 settings, compared with 17--19 for the learned extractors. Thus, CELL-Diff3D's advantage in population-level phenotype similarity holds across the learned feature spaces and in most settings with handcrafted features.

The extractors also differ in their sensitivity to OOD predictions. With the test set fixed, DynaCLR assigns higher KID to OOD predictions than to same-dataset predictions in only 14 of 32 comparisons, versus 21--27 for the other extractors. These results support using CELL-DINO KID for the main comparison, with GLCM+ cos providing a separate measure of paired-cell similarity.

\subsection{Metric reliability}
\label{sec:metric_reliability_appendix}

We assessed metric reliability using two sets of predictions whose poor quality is clear in Fig.~\ref{fig:prediction-overview}. Rand-init predictions contain little recognizable organelle structure, while FNet3D's OOD predictions have little spatial agreement with the experimental target. Both should therefore score worse than the in-distribution predictions. We do not assume an ordering between Rand-init and FNet3D's OOD predictions or between the two OOD directions.

Tabs.~\ref{tab:mr-fnet3d-pixel}--\ref{tab:mr-fnet3d-pheno} report all metrics defined in Sec.~\ref{sec:evaluation-framework}, grouped into pixel, segmentation, and phenotype tiers. Colored entries show FNet3D OOD predictions that receive a score at least as favorable as that of an in-distribution FNet3D prediction despite their poor visual quality. We distinguish two kinds of failure. A same-test-set failure (red) scores an OOD prediction at least as favorably as the in-distribution model on the same images (A549$\rightarrow$A549 for iPSC$\rightarrow$A549; iPSC$\rightarrow$iPSC for A549$\rightarrow$iPSC), so the metric misses the failure. A cross-test-set failure (purple) ranks the OOD prediction correctly on its own test set but scores it at least as favorably as the in-distribution prediction on the other test set, so the metric's absolute values depend on the test images~\citep{li2026doubting}. Rand-init predictions give an empirical floor for each test set: Rand-init nucleus MicroSSIM is $0.52$ on iPSC but $0.77$ on A549. These results motivated the five metrics retained in Tab.~\ref{tab:mr-pix2pix}.

\paragraph{Pixel tier.} Many pixel metrics assign unexpectedly favorable scores to the poor OOD predictions (Tab.~\ref{tab:mr-fnet3d-pixel}). PSNR, SSIM, NRMSE, and MicroSSIM do so for three of the four organelles, si-PSNR for two, and si-NRMSE for one. On the same test images, only PSNR, SSIM, and NRMSE do so, each for two of the four organelles. MicroSSIM, si-PSNR, and si-NRMSE rank the OOD predictions below the in-distribution model on the same images and fail only across test sets. In contrast, si-SSIM, PCC, Spectral PCC, and both FSC axes consistently score the OOD predictions below the in-distribution results in both comparisons. We therefore retain si-SSIM rather than SSIM. The si- prefix indicates that the prediction is fitted to the target with a least-squares affine transformation and that the data range is taken from the target. The unprefixed form rescales the prediction and target independently to $[0,1]$. The results for si-PSNR and si-NRMSE show that this affine fit alone does not make a metric comparable across test sets.

FSC reports the resolution in $\mu$m at which the shell correlation falls below its threshold, separately for the lateral and axial directions; lower values are better. If the correlation never crosses the threshold, FSC returns no value. All fields of view resolve in the in-distribution cases, compared with 68 of 120 for the FNet3D OOD nucleus prediction and 4 of 120 for the A549 Rand-init membrane prediction. FSC may therefore underestimate the degradation because its averages exclude unresolved fields. We use FSC and si-SSIM, rather than Spectral PCC alone, when assessing resolution loss (App.~\ref{sec:spectral_pcc}).

\definecolor{tierpixel}{RGB}{46,139,131}
\definecolor{tierseg}{RGB}{201,162,39}
\definecolor{tierpheno}{RGB}{74,90,115}
\definecolor{failsame}{RGB}{191,77,67}
\definecolor{crossfail}{RGB}{139,74,156}
\begin{table}[ht]
\centering
\small
\caption{\textbf{Metric behavior across in-distribution and OOD predictions, per organelle. Pixel-fidelity tier.} \textcolor{failsame}{Red} marks an OOD prediction scored at least as favorably as the in-distribution model on the same test images; \textcolor{crossfail}{Purple} marks one scored below the in-distribution model on the same test images but at least as favorably as an in-distribution prediction on another test set. Colored squares denote the metric tier: \textcolor{tierpixel}{\rule{1.1ex}{1.1ex}}~pixel.}
\label{tab:mr-fnet3d-pixel}
\renewcommand{\arraystretch}{0.85}
\resizebox{\textwidth}{!}{%
\begin{tabular}{l cc cccc}
\toprule
 & \multicolumn{2}{c}{\textbf{Rand-init}} & \multicolumn{4}{c}{\textbf{FNet3D}} \\
\cmidrule(lr){2-3}\cmidrule(lr){4-7}
\textbf{Metric} & \textbf{iPSC} & \textbf{A549} & \textbf{\shortstack{iPSC$\rightarrow$\\iPSC}} & \textbf{\shortstack{A549$\rightarrow$\\iPSC}} & \textbf{\shortstack{A549$\rightarrow$\\A549}} & \textbf{\shortstack{iPSC$\rightarrow$\\A549}} \\
\midrule
\multicolumn{7}{l}{\textbf{Nucleus}} \\
  \textcolor{tierpixel}{\rule{1.1ex}{1.1ex}}~PSNR ($\uparrow$) & $11.7 \pm 3.7$ & $7.3 \pm 0.4$ & $17.5 \pm 4.3$ & $15.5 \pm 3.3$ & $22.5 \pm 2.0$ & $\textcolor{crossfail}{21.1 \pm 1.3}$ \\
  \textcolor{tierpixel}{\rule{1.1ex}{1.1ex}}~si-PSNR ($\uparrow$) & $28.8 \pm 4.3$ & $22.6 \pm 1.7$ & $31.2 \pm 4.0$ & $\textcolor{crossfail}{28.9 \pm 4.2}$ & $28.8 \pm 2.1$ & $22.6 \pm 1.7$ \\
  \textcolor{tierpixel}{\rule{1.1ex}{1.1ex}}~SSIM ($\uparrow$) & $0.33 \pm 0.07$ & $0.15 \pm 0.03$ & $0.40 \pm 0.15$ & $0.31 \pm 0.14$ & $0.60 \pm 0.11$ & $\textcolor{failsame}{0.70 \pm 0.07}$ \\
  \textcolor{tierpixel}{\rule{1.1ex}{1.1ex}}~si-SSIM ($\uparrow$) & $0.26 \pm 0.19$ & $0.21 \pm 0.12$ & $0.56 \pm 0.16$ & $0.26 \pm 0.19$ & $0.88 \pm 0.02$ & $0.21 \pm 0.12$ \\
  \textcolor{tierpixel}{\rule{1.1ex}{1.1ex}}~PCC ($\uparrow$) & $-0.02 \pm 0.01$ & $-0.01 \pm 0.01$ & $0.62 \pm 0.11$ & $0.05 \pm 0.09$ & $0.87 \pm 0.05$ & $0.05 \pm 0.06$ \\
  \textcolor{tierpixel}{\rule{1.1ex}{1.1ex}}~NRMSE ($\downarrow$) & $0.28 \pm 0.10$ & $0.43 \pm 0.02$ & $0.15 \pm 0.07$ & $0.18 \pm 0.07$ & $0.08 \pm 0.02$ & $\textcolor{crossfail}{0.09 \pm 0.01}$ \\
  \textcolor{tierpixel}{\rule{1.1ex}{1.1ex}}~si-NRMSE ($\downarrow$) & $0.04 \pm 0.02$ & $0.08 \pm 0.01$ & $0.03 \pm 0.01$ & $0.04 \pm 0.02$ & $0.04 \pm 0.01$ & $0.08 \pm 0.01$ \\
  \textcolor{tierpixel}{\rule{1.1ex}{1.1ex}}~MicroSSIM ($\uparrow$) & $0.52 \pm 0.08$ & $0.77 \pm 0.04$ & $0.67 \pm 0.09$ & $0.47 \pm 0.08$ & $0.86 \pm 0.03$ & $\textcolor{crossfail}{0.78 \pm 0.05}$ \\
  \textcolor{tierpixel}{\rule{1.1ex}{1.1ex}}~Spectral PCC ($\uparrow$) & $-0.23 \pm 0.14$ & $0.16 \pm 0.12$ & $0.89 \pm 0.06$ & $0.11 \pm 0.19$ & $0.91 \pm 0.05$ & $0.05 \pm 0.18$ \\
  \textcolor{tierpixel}{\rule{1.1ex}{1.1ex}}~FSC (XY, $\mu$m) ($\downarrow$) & $8.05 \pm 4.95$ & $29.72 \pm 8.35$ & $0.77 \pm 0.20$ & $6.34 \pm 3.71$ & $0.84 \pm 0.18$ & $19.57 \pm 16.48$ \\
  \textcolor{tierpixel}{\rule{1.1ex}{1.1ex}}~FSC (Z, $\mu$m) ($\downarrow$) & $6.02 \pm 1.30$ & -- & $2.06 \pm 0.37$ & $5.19 \pm 1.76$ & $1.51 \pm 0.19$ & $6.41 \pm 1.54$ \\
\midrule
\multicolumn{7}{l}{\textbf{Membrane}} \\
  \textcolor{tierpixel}{\rule{1.1ex}{1.1ex}}~PSNR ($\uparrow$) & $7.7 \pm 0.7$ & $7.8 \pm 0.5$ & $19.0 \pm 4.0$ & $14.2 \pm 2.0$ & $24.3 \pm 1.9$ & $13.1 \pm 2.4$ \\
  \textcolor{tierpixel}{\rule{1.1ex}{1.1ex}}~si-PSNR ($\uparrow$) & $28.3 \pm 3.6$ & $23.9 \pm 1.3$ & $31.0 \pm 3.3$ & $\textcolor{crossfail}{28.6 \pm 3.6}$ & $28.0 \pm 1.4$ & $24.4 \pm 1.3$ \\
  \textcolor{tierpixel}{\rule{1.1ex}{1.1ex}}~SSIM ($\uparrow$) & $0.16 \pm 0.05$ & $0.15 \pm 0.03$ & $0.55 \pm 0.14$ & $0.29 \pm 0.10$ & $0.78 \pm 0.07$ & $0.31 \pm 0.08$ \\
  \textcolor{tierpixel}{\rule{1.1ex}{1.1ex}}~si-SSIM ($\uparrow$) & $0.28 \pm 0.16$ & $0.17 \pm 0.05$ & $0.58 \pm 0.12$ & $0.30 \pm 0.16$ & $0.63 \pm 0.09$ & $0.35 \pm 0.06$ \\
  \textcolor{tierpixel}{\rule{1.1ex}{1.1ex}}~PCC ($\uparrow$) & $0.04 \pm 0.03$ & $-0.03 \pm 0.02$ & $0.66 \pm 0.11$ & $0.18 \pm 0.15$ & $0.78 \pm 0.05$ & $0.32 \pm 0.07$ \\
  \textcolor{tierpixel}{\rule{1.1ex}{1.1ex}}~NRMSE ($\downarrow$) & $0.41 \pm 0.03$ & $0.41 \pm 0.02$ & $0.12 \pm 0.05$ & $0.20 \pm 0.04$ & $0.06 \pm 0.01$ & $0.23 \pm 0.06$ \\
  \textcolor{tierpixel}{\rule{1.1ex}{1.1ex}}~si-NRMSE ($\downarrow$) & $0.04 \pm 0.01$ & $0.06 \pm 0.01$ & $0.03 \pm 0.01$ & $\textcolor{crossfail}{0.04 \pm 0.01}$ & $0.04 \pm 0.01$ & $0.06 \pm 0.01$ \\
  \textcolor{tierpixel}{\rule{1.1ex}{1.1ex}}~MicroSSIM ($\uparrow$) & $0.46 \pm 0.16$ & $0.54 \pm 0.06$ & $0.59 \pm 0.17$ & $0.49 \pm 0.16$ & $0.80 \pm 0.03$ & $\textcolor{crossfail}{0.64 \pm 0.05}$ \\
  \textcolor{tierpixel}{\rule{1.1ex}{1.1ex}}~Spectral PCC ($\uparrow$) & $0.32 \pm 0.23$ & $-0.33 \pm 0.16$ & $0.89 \pm 0.11$ & $0.40 \pm 0.34$ & $0.88 \pm 0.05$ & $0.59 \pm 0.11$ \\
  \textcolor{tierpixel}{\rule{1.1ex}{1.1ex}}~FSC (XY, $\mu$m) ($\downarrow$) & $6.21 \pm 4.37$ & $6.08 \pm 0.57$ & $1.39 \pm 0.49$ & $5.66 \pm 3.20$ & $0.64 \pm 0.12$ & $2.14 \pm 3.74$ \\
  \textcolor{tierpixel}{\rule{1.1ex}{1.1ex}}~FSC (Z, $\mu$m) ($\downarrow$) & $5.04 \pm 2.18$ & $8.12 \pm 0.02$ & $2.73 \pm 0.42$ & $5.19 \pm 1.94$ & $1.23 \pm 0.37$ & $5.35 \pm 1.48$ \\
\midrule
\multicolumn{7}{l}{\textbf{ER}} \\
  \textcolor{tierpixel}{\rule{1.1ex}{1.1ex}}~PSNR ($\uparrow$) & $8.7 \pm 0.9$ & $9.4 \pm 1.9$ & $17.9 \pm 1.4$ & $\textcolor{failsame}{18.0 \pm 2.2}$ & $19.3 \pm 2.7$ & $\textcolor{failsame}{19.6 \pm 1.8}$ \\
  \textcolor{tierpixel}{\rule{1.1ex}{1.1ex}}~si-PSNR ($\uparrow$) & $22.3 \pm 1.0$ & $22.1 \pm 1.6$ & $25.3 \pm 1.2$ & $22.5 \pm 1.1$ & $25.5 \pm 1.6$ & $22.3 \pm 1.5$ \\
  \textcolor{tierpixel}{\rule{1.1ex}{1.1ex}}~SSIM ($\uparrow$) & $0.20 \pm 0.02$ & $0.29 \pm 0.08$ & $0.49 \pm 0.05$ & $0.37 \pm 0.07$ & $0.63 \pm 0.09$ & $\textcolor{crossfail}{0.58 \pm 0.08}$ \\
  \textcolor{tierpixel}{\rule{1.1ex}{1.1ex}}~si-SSIM ($\uparrow$) & $0.07 \pm 0.02$ & $0.10 \pm 0.04$ & $0.35 \pm 0.07$ & $0.09 \pm 0.07$ & $0.52 \pm 0.08$ & $0.08 \pm 0.04$ \\
  \textcolor{tierpixel}{\rule{1.1ex}{1.1ex}}~PCC ($\uparrow$) & $-0.01 \pm 0.03$ & $0.01 \pm 0.03$ & $0.70 \pm 0.07$ & $0.09 \pm 0.18$ & $0.73 \pm 0.05$ & $0.16 \pm 0.11$ \\
  \textcolor{tierpixel}{\rule{1.1ex}{1.1ex}}~NRMSE ($\downarrow$) & $0.37 \pm 0.04$ & $0.35 \pm 0.07$ & $0.13 \pm 0.02$ & $0.13 \pm 0.03$ & $0.11 \pm 0.03$ & $\textcolor{failsame}{0.11 \pm 0.02}$ \\
  \textcolor{tierpixel}{\rule{1.1ex}{1.1ex}}~si-NRMSE ($\downarrow$) & $0.08 \pm 0.01$ & $0.08 \pm 0.01$ & $0.05 \pm 0.01$ & $0.08 \pm 0.01$ & $0.05 \pm 0.01$ & $0.08 \pm 0.01$ \\
  \textcolor{tierpixel}{\rule{1.1ex}{1.1ex}}~MicroSSIM ($\uparrow$) & $0.24 \pm 0.07$ & $0.55 \pm 0.05$ & $0.54 \pm 0.09$ & $0.29 \pm 0.07$ & $0.75 \pm 0.03$ & $\textcolor{crossfail}{0.61 \pm 0.05}$ \\
  \textcolor{tierpixel}{\rule{1.1ex}{1.1ex}}~Spectral PCC ($\uparrow$) & $0.06 \pm 0.19$ & $0.13 \pm 0.14$ & $0.88 \pm 0.07$ & $0.05 \pm 0.31$ & $0.80 \pm 0.06$ & $0.32 \pm 0.21$ \\
  \textcolor{tierpixel}{\rule{1.1ex}{1.1ex}}~FSC (XY, $\mu$m) ($\downarrow$) & $7.39 \pm 4.42$ & $20.63 \pm 5.18$ & $0.64 \pm 0.07$ & $6.89 \pm 3.74$ & $0.94 \pm 0.54$ & $5.03 \pm 1.89$ \\
  \textcolor{tierpixel}{\rule{1.1ex}{1.1ex}}~FSC (Z, $\mu$m) ($\downarrow$) & $5.44 \pm 1.74$ & $4.30 \pm 0.82$ & $1.78 \pm 0.24$ & $6.05 \pm 1.34$ & $2.39 \pm 0.99$ & $6.53 \pm 1.36$ \\
\midrule
\multicolumn{7}{l}{\textbf{Mitochondria}} \\
  \textcolor{tierpixel}{\rule{1.1ex}{1.1ex}}~PSNR ($\uparrow$) & $9.2 \pm 0.8$ & $8.4 \pm 1.4$ & $21.7 \pm 1.5$ & $20.8 \pm 1.9$ & $21.7 \pm 3.9$ & $\textcolor{failsame}{22.2 \pm 2.8}$ \\
  \textcolor{tierpixel}{\rule{1.1ex}{1.1ex}}~si-PSNR ($\uparrow$) & $25.7 \pm 1.5$ & $25.6 \pm 1.5$ & $28.6 \pm 1.5$ & $25.8 \pm 1.5$ & $29.8 \pm 1.8$ & $25.7 \pm 1.5$ \\
  \textcolor{tierpixel}{\rule{1.1ex}{1.1ex}}~SSIM ($\uparrow$) & $0.26 \pm 0.04$ & $0.23 \pm 0.09$ & $0.50 \pm 0.07$ & $\textcolor{failsame}{0.56 \pm 0.07}$ & $0.66 \pm 0.14$ & $\textcolor{crossfail}{0.65 \pm 0.14}$ \\
  \textcolor{tierpixel}{\rule{1.1ex}{1.1ex}}~si-SSIM ($\uparrow$) & $0.21 \pm 0.06$ & $0.27 \pm 0.11$ & $0.55 \pm 0.06$ & $0.26 \pm 0.07$ & $0.75 \pm 0.07$ & $0.22 \pm 0.09$ \\
  \textcolor{tierpixel}{\rule{1.1ex}{1.1ex}}~PCC ($\uparrow$) & $0.01 \pm 0.03$ & $0.00 \pm 0.04$ & $0.70 \pm 0.03$ & $0.17 \pm 0.07$ & $0.78 \pm 0.05$ & $0.16 \pm 0.04$ \\
  \textcolor{tierpixel}{\rule{1.1ex}{1.1ex}}~NRMSE ($\downarrow$) & $0.35 \pm 0.03$ & $0.38 \pm 0.05$ & $0.08 \pm 0.02$ & $0.09 \pm 0.02$ & $0.09 \pm 0.05$ & $\textcolor{failsame}{0.08 \pm 0.03}$ \\
  \textcolor{tierpixel}{\rule{1.1ex}{1.1ex}}~si-NRMSE ($\downarrow$) & $0.05 \pm 0.01$ & $0.05 \pm 0.01$ & $0.04 \pm 0.01$ & $0.05 \pm 0.01$ & $0.03 \pm 0.01$ & $0.05 \pm 0.01$ \\
  \textcolor{tierpixel}{\rule{1.1ex}{1.1ex}}~MicroSSIM ($\uparrow$) & $0.45 \pm 0.04$ & $0.67 \pm 0.07$ & $0.72 \pm 0.04$ & $0.40 \pm 0.11$ & $0.82 \pm 0.04$ & $0.69 \pm 0.07$ \\
  \textcolor{tierpixel}{\rule{1.1ex}{1.1ex}}~Spectral PCC ($\uparrow$) & $-0.04 \pm 0.18$ & $0.13 \pm 0.09$ & $0.91 \pm 0.03$ & $0.33 \pm 0.17$ & $0.92 \pm 0.03$ & $0.22 \pm 0.10$ \\
  \textcolor{tierpixel}{\rule{1.1ex}{1.1ex}}~FSC (XY, $\mu$m) ($\downarrow$) & $5.67 \pm 2.61$ & $16.37 \pm 13.55$ & $0.63 \pm 0.01$ & $4.24 \pm 2.68$ & $1.10 \pm 0.71$ & $5.24 \pm 1.62$ \\
  \textcolor{tierpixel}{\rule{1.1ex}{1.1ex}}~FSC (Z, $\mu$m) ($\downarrow$) & $6.78 \pm 1.43$ & $7.12 \pm 1.58$ & $1.60 \pm 0.17$ & $5.06 \pm 1.06$ & $2.14 \pm 0.85$ & $6.02 \pm 1.89$ \\
\bottomrule
\end{tabular}%
}
\end{table}
 
\paragraph{Segmentation tier.} The segmentation metrics agree more closely with the images (Tab.~\ref{tab:mr-fnet3d-seg}). Semantic Dice, IoU, pixel precision and recall, instance Dice, AP@$0.50$, and mAP assign lower scores to the FNet3D OOD predictions than to the in-distribution predictions wherever these metrics are defined. Pixel accuracy is the exception for ER and mitochondria (for mitochondria also on the same test images) because background voxels dominate these sparse targets and keep accuracy near 1 even for poor predictions. In this full-metric table, ``Dice'' denotes semantic Dice, and instance Dice is listed separately. Precision, recall, and accuracy marked ``px'' are computed from binary-mask voxels. Instance Dice, AP@$0.50$, and mAP are defined only for nucleus and membrane. Instance Dice averages each object's best Dice overlap in both directions; missed and spurious objects contribute zero. Pairs with no objects in either mask have undefined instance Dice and are excluded from its averages. For subsequent model comparisons, we retain instance Dice for nucleus and membrane and semantic Dice for ER and mitochondria. Semantic Dice is modest even in distribution for ER ($0.09$--$0.24$) and mitochondria ($0.38$--$0.46$), so these values distinguish relative performance but should not be treated as absolute measures of prediction quality.

\definecolor{tierpixel}{RGB}{46,139,131}
\definecolor{tierseg}{RGB}{201,162,39}
\definecolor{tierpheno}{RGB}{74,90,115}
\definecolor{failsame}{RGB}{191,77,67}
\definecolor{crossfail}{RGB}{139,74,156}
\begin{table}[ht]
\centering
\small
\caption{\textbf{Metric behavior across in-distribution and OOD predictions, per organelle. Segmentation-utility tier.} \textcolor{failsame}{Red} marks an OOD prediction scored at least as favorably as the in-distribution model on the same test images; \textcolor{crossfail}{Purple} marks one scored below the in-distribution model on the same test images but at least as favorably as an in-distribution prediction on another test set. Colored squares denote the metric tier: \textcolor{tierseg}{\rule{1.1ex}{1.1ex}}~segmentation.}
\label{tab:mr-fnet3d-seg}
\renewcommand{\arraystretch}{0.85}
\resizebox{\textwidth}{!}{%
\begin{tabular}{l cc cccc}
\toprule
 & \multicolumn{2}{c}{\textbf{Rand-init}} & \multicolumn{4}{c}{\textbf{FNet3D}} \\
\cmidrule(lr){2-3}\cmidrule(lr){4-7}
\textbf{Metric} & \textbf{iPSC} & \textbf{A549} & \textbf{\shortstack{iPSC$\rightarrow$\\iPSC}} & \textbf{\shortstack{A549$\rightarrow$\\iPSC}} & \textbf{\shortstack{A549$\rightarrow$\\A549}} & \textbf{\shortstack{iPSC$\rightarrow$\\A549}} \\
\midrule
\multicolumn{7}{l}{\textbf{Nucleus}} \\
  \textcolor{tierseg}{\rule{1.1ex}{1.1ex}}~Dice ($\uparrow$) & $0.02 \pm 0.09$ & $0.00 \pm 0.01$ & $0.95 \pm 0.04$ & $0.15 \pm 0.17$ & $0.96 \pm 0.02$ & $0.06 \pm 0.07$ \\
  \textcolor{tierseg}{\rule{1.1ex}{1.1ex}}~IoU ($\uparrow$) & $0.01 \pm 0.06$ & $0.00$ & $0.90 \pm 0.06$ & $0.09 \pm 0.11$ & $0.93 \pm 0.03$ & $0.03 \pm 0.04$ \\
  \textcolor{tierseg}{\rule{1.1ex}{1.1ex}}~Precision (px) ($\uparrow$) & $0.06 \pm 0.21$ & $0.00 \pm 0.02$ & $0.95 \pm 0.05$ & $0.52 \pm 0.36$ & $0.96 \pm 0.03$ & $0.31 \pm 0.32$ \\
  \textcolor{tierseg}{\rule{1.1ex}{1.1ex}}~Recall (px) ($\uparrow$) & $0.01 \pm 0.07$ & $0.00$ & $0.95 \pm 0.03$ & $0.10 \pm 0.13$ & $0.97 \pm 0.01$ & $0.03 \pm 0.05$ \\
  \textcolor{tierseg}{\rule{1.1ex}{1.1ex}}~Accuracy (px) ($\uparrow$) & $0.49 \pm 0.09$ & $0.87 \pm 0.05$ & $0.95 \pm 0.03$ & $0.52 \pm 0.08$ & $0.99 \pm 0.01$ & $0.86 \pm 0.05$ \\
  \textcolor{tierseg}{\rule{1.1ex}{1.1ex}}~Instance Dice ($\uparrow$) & $0.01 \pm 0.07$ & $0.00 \pm 0.01$ & $0.89 \pm 0.08$ & $0.14 \pm 0.14$ & $0.91 \pm 0.05$ & $0.05 \pm 0.06$ \\
  \textcolor{tierseg}{\rule{1.1ex}{1.1ex}}~AP@0.50 ($\uparrow$) & $0.01 \pm 0.05$ & $0.00$ & $0.87 \pm 0.14$ & $0.05 \pm 0.09$ & $0.88 \pm 0.11$ & $0.01 \pm 0.03$ \\
  \textcolor{tierseg}{\rule{1.1ex}{1.1ex}}~mAP ($\uparrow$) & $0.00 \pm 0.03$ & $0.00$ & $0.71 \pm 0.15$ & $0.02 \pm 0.04$ & $0.76 \pm 0.12$ & $0.01 \pm 0.01$ \\
\midrule
\multicolumn{7}{l}{\textbf{Membrane}} \\
  \textcolor{tierseg}{\rule{1.1ex}{1.1ex}}~Dice ($\uparrow$) & $0.11 \pm 0.13$ & $0.00 \pm 0.01$ & $0.87 \pm 0.14$ & $0.06 \pm 0.14$ & $0.78 \pm 0.13$ & $0.35 \pm 0.20$ \\
  \textcolor{tierseg}{\rule{1.1ex}{1.1ex}}~IoU ($\uparrow$) & $0.06 \pm 0.08$ & $0.00 \pm 0.01$ & $0.79 \pm 0.16$ & $0.04 \pm 0.10$ & $0.65 \pm 0.15$ & $0.23 \pm 0.15$ \\
  \textcolor{tierseg}{\rule{1.1ex}{1.1ex}}~Precision (px) ($\uparrow$) & $0.58 \pm 0.46$ & $0.02 \pm 0.15$ & $0.94 \pm 0.14$ & $0.31 \pm 0.46$ & $0.77 \pm 0.16$ & $0.74 \pm 0.27$ \\
  \textcolor{tierseg}{\rule{1.1ex}{1.1ex}}~Recall (px) ($\uparrow$) & $0.06 \pm 0.08$ & $0.00 \pm 0.01$ & $0.84 \pm 0.11$ & $0.04 \pm 0.10$ & $0.81 \pm 0.14$ & $0.25 \pm 0.18$ \\
  \textcolor{tierseg}{\rule{1.1ex}{1.1ex}}~Accuracy (px) ($\uparrow$) & $0.60 \pm 0.09$ & $0.62 \pm 0.15$ & $0.91 \pm 0.07$ & $0.59 \pm 0.10$ & $0.84 \pm 0.07$ & $0.69 \pm 0.13$ \\
  \textcolor{tierseg}{\rule{1.1ex}{1.1ex}}~Instance Dice ($\uparrow$) & $0.09 \pm 0.10$ & $0.00 \pm 0.02$ & $0.67 \pm 0.12$ & $0.04 \pm 0.09$ & $0.70 \pm 0.13$ & $0.31 \pm 0.16$ \\
  \textcolor{tierseg}{\rule{1.1ex}{1.1ex}}~AP@0.50 ($\uparrow$) & $0.01 \pm 0.02$ & $0.00 \pm 0.01$ & $0.50 \pm 0.19$ & $0.01 \pm 0.03$ & $0.60 \pm 0.19$ & $0.14 \pm 0.14$ \\
  \textcolor{tierseg}{\rule{1.1ex}{1.1ex}}~mAP ($\uparrow$) & $0.00 \pm 0.01$ & $0.00$ & $0.19 \pm 0.09$ & $0.00 \pm 0.01$ & $0.37 \pm 0.13$ & $0.05 \pm 0.06$ \\
\midrule
\multicolumn{7}{l}{\textbf{ER}} \\
  \textcolor{tierseg}{\rule{1.1ex}{1.1ex}}~Dice ($\uparrow$) & $0.07 \pm 0.03$ & $0.08 \pm 0.03$ & $0.24 \pm 0.05$ & $0.05 \pm 0.03$ & $0.09 \pm 0.03$ & $0.07 \pm 0.01$ \\
  \textcolor{tierseg}{\rule{1.1ex}{1.1ex}}~IoU ($\uparrow$) & $0.04 \pm 0.01$ & $0.04 \pm 0.01$ & $0.14 \pm 0.03$ & $0.02 \pm 0.01$ & $0.05 \pm 0.02$ & $0.03 \pm 0.01$ \\
  \textcolor{tierseg}{\rule{1.1ex}{1.1ex}}~Precision (px) ($\uparrow$) & $0.06 \pm 0.03$ & $0.05 \pm 0.02$ & $0.55 \pm 0.05$ & $0.10 \pm 0.03$ & $0.32 \pm 0.10$ & $0.10 \pm 0.03$ \\
  \textcolor{tierseg}{\rule{1.1ex}{1.1ex}}~Recall (px) ($\uparrow$) & $0.10 \pm 0.03$ & $0.19 \pm 0.03$ & $0.16 \pm 0.04$ & $0.03 \pm 0.02$ & $0.06 \pm 0.03$ & $0.05 \pm 0.01$ \\
  \textcolor{tierseg}{\rule{1.1ex}{1.1ex}}~Accuracy (px) ($\uparrow$) & $0.80 \pm 0.02$ & $0.77 \pm 0.02$ & $0.92 \pm 0.02$ & $0.90 \pm 0.02$ & $0.94 \pm 0.02$ & $\textcolor{crossfail}{0.92 \pm 0.02}$ \\
\midrule
\multicolumn{7}{l}{\textbf{Mitochondria}} \\
  \textcolor{tierseg}{\rule{1.1ex}{1.1ex}}~Dice ($\uparrow$) & $0.09 \pm 0.03$ & $0.07 \pm 0.02$ & $0.46 \pm 0.04$ & $0.11 \pm 0.03$ & $0.38 \pm 0.07$ & $0.02 \pm 0.01$ \\
  \textcolor{tierseg}{\rule{1.1ex}{1.1ex}}~IoU ($\uparrow$) & $0.05 \pm 0.02$ & $0.03 \pm 0.01$ & $0.30 \pm 0.03$ & $0.06 \pm 0.02$ & $0.24 \pm 0.05$ & $0.01 \pm 0.01$ \\
  \textcolor{tierseg}{\rule{1.1ex}{1.1ex}}~Precision (px) ($\uparrow$) & $0.06 \pm 0.02$ & $0.04 \pm 0.01$ & $0.51 \pm 0.06$ & $0.15 \pm 0.05$ & $0.42 \pm 0.07$ & $0.08 \pm 0.02$ \\
  \textcolor{tierseg}{\rule{1.1ex}{1.1ex}}~Recall (px) ($\uparrow$) & $0.20 \pm 0.04$ & $0.23 \pm 0.04$ & $0.42 \pm 0.06$ & $0.11 \pm 0.05$ & $0.36 \pm 0.09$ & $0.01 \pm 0.01$ \\
  \textcolor{tierseg}{\rule{1.1ex}{1.1ex}}~Accuracy (px) ($\uparrow$) & $0.75 \pm 0.03$ & $0.75 \pm 0.03$ & $0.94 \pm 0.01$ & $0.89 \pm 0.03$ & $0.96 \pm 0.01$ & $\textcolor{failsame}{0.96 \pm 0.01}$ \\
\bottomrule
\end{tabular}%
}
\end{table}
 
\paragraph{Phenotype tier.} Most phenotype metrics do not consistently reflect the poor OOD predictions (Tab.~\ref{tab:mr-fnet3d-pheno}). Fourteen of the fifteen extractor--metric combinations assign an unexpectedly favorable score on at least two organelles. GLCM+ cos is the only exception, doing so only for nucleus. CELL-DINO KID and MorphEm KID each give favorable scores to poor predictions on two organelles, compared with three for DINOv3 KID and four for GLCM+ KID and DynaCLR KID. On the same test images, all three DynaCLR metrics fail on all four organelles, GLCM+ KID on three, GLCM+ FID on two, and each DINOv3 metric on mitochondria. The CELL-DINO and MorphEm metrics and GLCM+ cos fail only across test sets, although CELL-DINO KID separates the A549 ER predictions by only $0.4$, close to its resampling SD of $0.2$--$0.3$. We retain CELL-DINO KID because it separates the models more consistently in the comparison of feature extractors (App.~\ref{sec:feature-space-agreement}) and report GLCM+ cos as a complementary paired-cell measure. FID and KID values are comparable across model predictions within a feature space but not across feature extractors, whose numerical scales differ substantially. We therefore interpret phenotype metrics together with the pixel and segmentation results and the prediction images.

\definecolor{tierpixel}{RGB}{46,139,131}
\definecolor{tierseg}{RGB}{201,162,39}
\definecolor{tierpheno}{RGB}{74,90,115}
\definecolor{failsame}{RGB}{191,77,67}
\definecolor{crossfail}{RGB}{139,74,156}
\begin{table}[ht]
\centering
\small
\caption{\textbf{Metric behavior across in-distribution and OOD predictions, per organelle. Phenotype-similarity tier.} \textcolor{failsame}{Red} marks an OOD prediction scored at least as favorably as the in-distribution model on the same test images; \textcolor{crossfail}{Purple} marks one scored below the in-distribution model on the same test images but at least as favorably as an in-distribution prediction on another test set. Colored squares denote the metric tier: \textcolor{tierpheno}{\rule{1.1ex}{1.1ex}}~phenotype.}
\label{tab:mr-fnet3d-pheno}
\renewcommand{\arraystretch}{0.85}
\resizebox{\textwidth}{!}{%
\begin{tabular}{l cc cccc}
\toprule
 & \multicolumn{2}{c}{\textbf{Rand-init}} & \multicolumn{4}{c}{\textbf{FNet3D}} \\
\cmidrule(lr){2-3}\cmidrule(lr){4-7}
\textbf{Metric} & \textbf{iPSC} & \textbf{A549} & \textbf{\shortstack{iPSC$\rightarrow$\\iPSC}} & \textbf{\shortstack{A549$\rightarrow$\\iPSC}} & \textbf{\shortstack{A549$\rightarrow$\\A549}} & \textbf{\shortstack{iPSC$\rightarrow$\\A549}} \\
\midrule
\multicolumn{7}{l}{\textbf{Nucleus}} \\
  \textcolor{tierpheno}{\rule{1.1ex}{1.1ex}}~GLCM+ cos ($\uparrow$) & $-0.06$ & $-0.10$ & $0.00$ & $-0.21$ & $0.39$ & $\textcolor{crossfail}{0.14}$ \\
  \textcolor{tierpheno}{\rule{1.1ex}{1.1ex}}~GLCM+ FID ($\downarrow$) & $34.00$ & $281.08$ & $175.64$ & $338.58$ & $8.14$ & $\textcolor{crossfail}{8.29}$ \\
  \textcolor{tierpheno}{\rule{1.1ex}{1.1ex}}~GLCM+ KID ($\downarrow$) & $64.28 \pm 0.31$ & $31120.73 \pm 41.63$ & $10114.81 \pm 450.35$ & $73201.77 \pm 1969.79$ & $7.57 \pm 0.47$ & $\textcolor{failsame}{2.60 \pm 0.03}$ \\
  \textcolor{tierpheno}{\rule{1.1ex}{1.1ex}}~DINOv3 Median cos ($\uparrow$) & $0.54$ & $0.42$ & $0.73$ & $0.59$ & $0.78$ & $0.62$ \\
  \textcolor{tierpheno}{\rule{1.1ex}{1.1ex}}~DINOv3 FID ($\downarrow$) & $825$ & $1367$ & $359$ & $724$ & $376$ & $788$ \\
  \textcolor{tierpheno}{\rule{1.1ex}{1.1ex}}~DINOv3 KID ($\downarrow$) & $4.25 \pm 0.07$ & $8.26 \pm 0.04$ & $1.79 \pm 0.04$ & $3.80 \pm 0.07$ & $2.49 \pm 0.03$ & $5.15 \pm 0.04$ \\
  \textcolor{tierpheno}{\rule{1.1ex}{1.1ex}}~DynaCLR Median cos ($\uparrow$) & $0.78$ & $0.49$ & $0.16$ & $\textcolor{failsame}{0.66}$ & $0.59$ & $\textcolor{failsame}{0.67}$ \\
  \textcolor{tierpheno}{\rule{1.1ex}{1.1ex}}~DynaCLR FID ($\downarrow$) & $273$ & $717$ & $1051$ & $\textcolor{failsame}{404}$ & $466$ & $\textcolor{failsame}{330}$ \\
  \textcolor{tierpheno}{\rule{1.1ex}{1.1ex}}~DynaCLR KID ($\downarrow$) & $2.07 \pm 0.03$ & $4.13 \pm 0.01$ & $6.62 \pm 0.08$ & $\textcolor{failsame}{2.95 \pm 0.03}$ & $2.74 \pm 0.02$ & $\textcolor{failsame}{1.60 \pm 0.01}$ \\
  \textcolor{tierpheno}{\rule{1.1ex}{1.1ex}}~CELL-DINO Median cos ($\uparrow$) & $0.07$ & $0.09$ & $0.13$ & $0.07$ & $0.30$ & $\textcolor{crossfail}{0.19}$ \\
  \textcolor{tierpheno}{\rule{1.1ex}{1.1ex}}~CELL-DINO FID ($\downarrow$) & $4754$ & $5029$ & $4015$ & $4598$ & $3133$ & $\textcolor{crossfail}{3695}$ \\
  \textcolor{tierpheno}{\rule{1.1ex}{1.1ex}}~CELL-DINO KID ($\downarrow$) & $43.3 \pm 0.3$ & $54.1 \pm 0.1$ & $24.6 \pm 0.2$ & $38.7 \pm 0.2$ & $21.5 \pm 0.1$ & $30.4 \pm 0.1$ \\
  \textcolor{tierpheno}{\rule{1.1ex}{1.1ex}}~MorphEm Median cos ($\uparrow$) & $0.19$ & $0.30$ & $0.27$ & $0.22$ & $0.71$ & $\textcolor{crossfail}{0.38}$ \\
  \textcolor{tierpheno}{\rule{1.1ex}{1.1ex}}~MorphEm FID ($\downarrow$) & $4749$ & $4169$ & $4080$ & $4376$ & $1447$ & $\textcolor{crossfail}{2993}$ \\
  \textcolor{tierpheno}{\rule{1.1ex}{1.1ex}}~MorphEm KID ($\downarrow$) & $793 \pm 6$ & $738 \pm 2$ & $510 \pm 4$ & $623 \pm 4$ & $184 \pm 1$ & $518 \pm 1$ \\
\midrule
\multicolumn{7}{l}{\textbf{Membrane}} \\
  \textcolor{tierpheno}{\rule{1.1ex}{1.1ex}}~GLCM+ cos ($\uparrow$) & $-0.04$ & $-0.06$ & $0.20$ & $0.03$ & $0.32$ & $0.09$ \\
  \textcolor{tierpheno}{\rule{1.1ex}{1.1ex}}~GLCM+ FID ($\downarrow$) & $57.89$ & $825.76$ & $23.10$ & $\textcolor{failsame}{21.57}$ & $9.27$ & $29.91$ \\
  \textcolor{tierpheno}{\rule{1.1ex}{1.1ex}}~GLCM+ KID ($\downarrow$) & $228.27 \pm 1.57$ & $568204.87 \pm 403.35$ & $25.61 \pm 0.49$ & $\textcolor{failsame}{20.14 \pm 0.28}$ & $4.27 \pm 0.04$ & $38.46 \pm 0.11$ \\
  \textcolor{tierpheno}{\rule{1.1ex}{1.1ex}}~DINOv3 Median cos ($\uparrow$) & $0.55$ & $0.51$ & $0.66$ & $0.56$ & $0.75$ & $\textcolor{crossfail}{0.69}$ \\
  \textcolor{tierpheno}{\rule{1.1ex}{1.1ex}}~DINOv3 FID ($\downarrow$) & $766$ & $962$ & $504$ & $827$ & $361$ & $\textcolor{crossfail}{399}$ \\
  \textcolor{tierpheno}{\rule{1.1ex}{1.1ex}}~DINOv3 KID ($\downarrow$) & $3.36 \pm 0.04$ & $4.63 \pm 0.01$ & $2.51 \pm 0.05$ & $4.22 \pm 0.07$ & $1.59 \pm 0.01$ & $\textcolor{crossfail}{1.64 \pm 0.01}$ \\
  \textcolor{tierpheno}{\rule{1.1ex}{1.1ex}}~DynaCLR Median cos ($\uparrow$) & $0.74$ & $0.49$ & $0.45$ & $\textcolor{failsame}{0.56}$ & $0.62$ & $\textcolor{failsame}{0.76}$ \\
  \textcolor{tierpheno}{\rule{1.1ex}{1.1ex}}~DynaCLR FID ($\downarrow$) & $333$ & $660$ & $628$ & $\textcolor{failsame}{471}$ & $455$ & $\textcolor{failsame}{179}$ \\
  \textcolor{tierpheno}{\rule{1.1ex}{1.1ex}}~DynaCLR KID ($\downarrow$) & $2.07 \pm 0.03$ & $4.30 \pm 0.01$ & $4.35 \pm 0.06$ & $\textcolor{failsame}{3.29 \pm 0.06}$ & $3.00 \pm 0.01$ & $\textcolor{failsame}{1.03 \pm 0.01}$ \\
  \textcolor{tierpheno}{\rule{1.1ex}{1.1ex}}~CELL-DINO Median cos ($\uparrow$) & $0.13$ & $0.10$ & $0.21$ & $0.17$ & $0.41$ & $\textcolor{crossfail}{0.27}$ \\
  \textcolor{tierpheno}{\rule{1.1ex}{1.1ex}}~CELL-DINO FID ($\downarrow$) & $4325$ & $5060$ & $3714$ & $4047$ & $2594$ & $\textcolor{crossfail}{3388}$ \\
  \textcolor{tierpheno}{\rule{1.1ex}{1.1ex}}~CELL-DINO KID ($\downarrow$) & $40.7 \pm 0.3$ & $55.4 \pm 0.1$ & $36.4 \pm 0.2$ & $41.7 \pm 0.3$ & $25.1 \pm 0.1$ & $\textcolor{crossfail}{36.2 \pm 0.1}$ \\
  \textcolor{tierpheno}{\rule{1.1ex}{1.1ex}}~MorphEm Median cos ($\uparrow$) & $0.40$ & $0.31$ & $0.62$ & $0.58$ & $0.82$ & $\textcolor{crossfail}{0.70}$ \\
  \textcolor{tierpheno}{\rule{1.1ex}{1.1ex}}~MorphEm FID ($\downarrow$) & $3405$ & $4272$ & $1903$ & $2019$ & $782$ & $\textcolor{crossfail}{1046}$ \\
  \textcolor{tierpheno}{\rule{1.1ex}{1.1ex}}~MorphEm KID ($\downarrow$) & $718 \pm 5$ & $914 \pm 1$ & $447 \pm 5$ & $466 \pm 4$ & $205 \pm 1$ & $\textcolor{crossfail}{242 \pm 1}$ \\
\midrule
\multicolumn{7}{l}{\textbf{ER}} \\
  \textcolor{tierpheno}{\rule{1.1ex}{1.1ex}}~GLCM+ cos ($\uparrow$) & $-0.05$ & $-0.04$ & $0.17$ & $0.02$ & $0.22$ & $-0.03$ \\
  \textcolor{tierpheno}{\rule{1.1ex}{1.1ex}}~GLCM+ FID ($\downarrow$) & $91.50$ & $70.84$ & $136.36$ & $235.58$ & $27.18$ & $\textcolor{failsame}{16.01}$ \\
  \textcolor{tierpheno}{\rule{1.1ex}{1.1ex}}~GLCM+ KID ($\downarrow$) & $506.29 \pm 2.79$ & $215.69 \pm 0.89$ & $1905.38 \pm 42.11$ & $9202.61 \pm 211.27$ & $37.01 \pm 4.36$ & $\textcolor{failsame}{6.25 \pm 0.21}$ \\
  \textcolor{tierpheno}{\rule{1.1ex}{1.1ex}}~DINOv3 Median cos ($\uparrow$) & $0.56$ & $0.50$ & $0.72$ & $0.57$ & $0.71$ & $0.67$ \\
  \textcolor{tierpheno}{\rule{1.1ex}{1.1ex}}~DINOv3 FID ($\downarrow$) & $869$ & $1006$ & $523$ & $875$ & $475$ & $\textcolor{crossfail}{506}$ \\
  \textcolor{tierpheno}{\rule{1.1ex}{1.1ex}}~DINOv3 KID ($\downarrow$) & $5.50 \pm 0.10$ & $5.20 \pm 0.10$ & $3.69 \pm 0.06$ & $5.72 \pm 0.09$ & $2.58 \pm 0.07$ & $\textcolor{crossfail}{2.65 \pm 0.08}$ \\
  \textcolor{tierpheno}{\rule{1.1ex}{1.1ex}}~DynaCLR Median cos ($\uparrow$) & $0.84$ & $0.66$ & $0.45$ & $\textcolor{crossfail}{0.41}$ & $0.37$ & $\textcolor{failsame}{0.84}$ \\
  \textcolor{tierpheno}{\rule{1.1ex}{1.1ex}}~DynaCLR FID ($\downarrow$) & $178$ & $448$ & $789$ & $\textcolor{failsame}{747}$ & $838$ & $\textcolor{failsame}{136}$ \\
  \textcolor{tierpheno}{\rule{1.1ex}{1.1ex}}~DynaCLR KID ($\downarrow$) & $1.30 \pm 0.02$ & $3.37 \pm 0.03$ & $5.19 \pm 0.05$ & $\textcolor{failsame}{4.97 \pm 0.06}$ & $6.08 \pm 0.05$ & $\textcolor{failsame}{0.89 \pm 0.01}$ \\
  \textcolor{tierpheno}{\rule{1.1ex}{1.1ex}}~CELL-DINO Median cos ($\uparrow$) & $0.13$ & $0.08$ & $0.16$ & $0.09$ & $0.19$ & $0.13$ \\
  \textcolor{tierpheno}{\rule{1.1ex}{1.1ex}}~CELL-DINO FID ($\downarrow$) & $4348$ & $5046$ & $4218$ & $4654$ & $3820$ & $\textcolor{crossfail}{4005}$ \\
  \textcolor{tierpheno}{\rule{1.1ex}{1.1ex}}~CELL-DINO KID ($\downarrow$) & $42.5 \pm 0.3$ & $50.3 \pm 0.4$ & $34.5 \pm 0.2$ & $42.6 \pm 0.2$ & $30.3 \pm 0.2$ & $\textcolor{crossfail}{30.7 \pm 0.3}$ \\
  \textcolor{tierpheno}{\rule{1.1ex}{1.1ex}}~MorphEm Median cos ($\uparrow$) & $0.45$ & $0.30$ & $0.59$ & $\textcolor{crossfail}{0.51}$ & $0.48$ & $0.44$ \\
  \textcolor{tierpheno}{\rule{1.1ex}{1.1ex}}~MorphEm FID ($\downarrow$) & $3085$ & $4154$ & $2327$ & $2601$ & $2447$ & $2658$ \\
  \textcolor{tierpheno}{\rule{1.1ex}{1.1ex}}~MorphEm KID ($\downarrow$) & $715 \pm 6$ & $693 \pm 6$ & $563 \pm 4$ & $638 \pm 5$ & $378 \pm 5$ & $\textcolor{crossfail}{495 \pm 5}$ \\
\midrule
\multicolumn{7}{l}{\textbf{Mitochondria}} \\
  \textcolor{tierpheno}{\rule{1.1ex}{1.1ex}}~GLCM+ cos ($\uparrow$) & $-0.02$ & $0.01$ & $0.16$ & $0.03$ & $0.27$ & $0.00$ \\
  \textcolor{tierpheno}{\rule{1.1ex}{1.1ex}}~GLCM+ FID ($\downarrow$) & $199.31$ & $122.74$ & $182.37$ & $233.11$ & $35.39$ & $\textcolor{crossfail}{56.12}$ \\
  \textcolor{tierpheno}{\rule{1.1ex}{1.1ex}}~GLCM+ KID ($\downarrow$) & $4834.83 \pm 21.26$ & $1157.15 \pm 9.03$ & $4933.10 \pm 175.35$ & $9675.48 \pm 381.95$ & $75.57 \pm 7.52$ & $\textcolor{crossfail}{150.84 \pm 2.25}$ \\
  \textcolor{tierpheno}{\rule{1.1ex}{1.1ex}}~DINOv3 Median cos ($\uparrow$) & $0.54$ & $0.42$ & $0.62$ & $\textcolor{failsame}{0.66}$ & $0.67$ & $0.57$ \\
  \textcolor{tierpheno}{\rule{1.1ex}{1.1ex}}~DINOv3 FID ($\downarrow$) & $874$ & $1250$ & $751$ & $\textcolor{failsame}{535}$ & $645$ & $\textcolor{crossfail}{728}$ \\
  \textcolor{tierpheno}{\rule{1.1ex}{1.1ex}}~DINOv3 KID ($\downarrow$) & $4.77 \pm 0.06$ & $6.34 \pm 0.06$ & $4.70 \pm 0.10$ & $\textcolor{failsame}{3.12 \pm 0.05}$ & $3.99 \pm 0.07$ & $\textcolor{failsame}{3.64 \pm 0.06}$ \\
  \textcolor{tierpheno}{\rule{1.1ex}{1.1ex}}~DynaCLR Median cos ($\uparrow$) & $0.76$ & $0.62$ & $0.51$ & $\textcolor{failsame}{0.64}$ & $0.51$ & $\textcolor{failsame}{0.76}$ \\
  \textcolor{tierpheno}{\rule{1.1ex}{1.1ex}}~DynaCLR FID ($\downarrow$) & $278$ & $494$ & $620$ & $\textcolor{failsame}{437}$ & $612$ & $\textcolor{failsame}{258}$ \\
  \textcolor{tierpheno}{\rule{1.1ex}{1.1ex}}~DynaCLR KID ($\downarrow$) & $2.14 \pm 0.03$ & $3.86 \pm 0.03$ & $4.40 \pm 0.05$ & $\textcolor{failsame}{3.14 \pm 0.04}$ & $4.39 \pm 0.05$ & $\textcolor{failsame}{2.04 \pm 0.03}$ \\
  \textcolor{tierpheno}{\rule{1.1ex}{1.1ex}}~CELL-DINO Median cos ($\uparrow$) & $0.09$ & $0.11$ & $0.14$ & $0.10$ & $0.22$ & $\textcolor{crossfail}{0.16}$ \\
  \textcolor{tierpheno}{\rule{1.1ex}{1.1ex}}~CELL-DINO FID ($\downarrow$) & $4694$ & $4906$ & $4128$ & $4512$ & $3403$ & $4642$ \\
  \textcolor{tierpheno}{\rule{1.1ex}{1.1ex}}~CELL-DINO KID ($\downarrow$) & $45.6 \pm 0.3$ & $52.5 \pm 0.4$ & $30.2 \pm 0.2$ & $41.2 \pm 0.3$ & $22.6 \pm 0.2$ & $55.9 \pm 0.3$ \\
  \textcolor{tierpheno}{\rule{1.1ex}{1.1ex}}~MorphEm Median cos ($\uparrow$) & $0.29$ & $0.35$ & $0.60$ & $0.51$ & $0.54$ & $0.30$ \\
  \textcolor{tierpheno}{\rule{1.1ex}{1.1ex}}~MorphEm FID ($\downarrow$) & $4095$ & $3839$ & $2103$ & $2345$ & $2336$ & $4150$ \\
  \textcolor{tierpheno}{\rule{1.1ex}{1.1ex}}~MorphEm KID ($\downarrow$) & $802 \pm 6$ & $778 \pm 6$ & $446 \pm 5$ & $465 \pm 4$ & $428 \pm 5$ & $904 \pm 6$ \\
\bottomrule
\end{tabular}%
}
\end{table}
 
\FloatBarrier

\section{Models, training, and ablations}

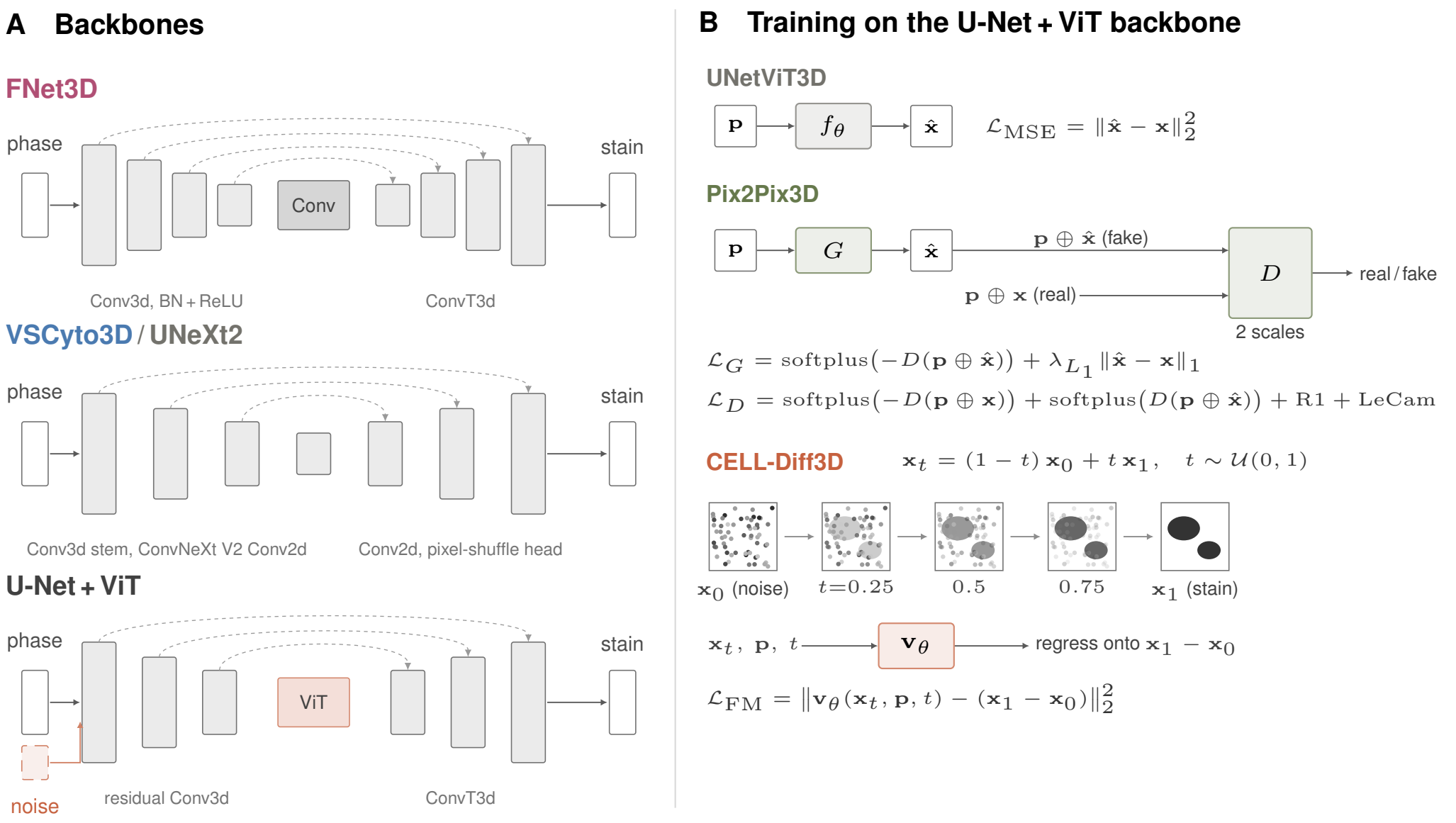
\begin{figure}[!ht]
  \centering
  \definecolor{mFNet}{HTML}{B04F74}
  \definecolor{mVSC}{HTML}{4A7AB0}
  \definecolor{mP2P}{HTML}{66794C}
  \definecolor{mCD}{HTML}{C6613F}
  \definecolor{mAbl}{HTML}{73726C}
  \begin{tikzpicture}[
      x=1pt, y=1pt,
      font=\sffamily,
      blk/.style={draw=black!55, line width=0.3pt, fill=black!8,
                  rounded corners=0.6pt, inner sep=0pt, minimum width=9pt},
      bneck/.style={draw=black!55, line width=0.3pt, fill=black!16,
                    rounded corners=0.6pt, inner sep=0pt,
                    minimum width=19pt, minimum height=13pt,
                    font=\sffamily\fontsize{5}{5}\selectfont, text=black!75},
      vit/.style={bneck, fill=mCD!20, draw=mCD!75},
      io/.style={draw=black!55, line width=0.3pt, fill=white,
                 rounded corners=0.6pt, inner sep=0pt,
                 minimum width=7pt, minimum height=17pt},
      flow/.style={-{Latex[length=1.8pt,width=1.8pt]}, line width=0.35pt, black!65},
      skip/.style={dashed, dash pattern=on 1pt off 1pt, line width=0.25pt, black!40},
      skiparr/.style={skip, -{Latex[length=2pt,width=1.8pt]}},
      tag/.style={font=\sffamily\fontsize{5.4}{6}\selectfont, anchor=south, text=black!70},
      blab/.style={font=\sffamily\fontsize{4.6}{5}\selectfont, anchor=north, text=black!55},
      mname/.style={font=\sffamily\bfseries\fontsize{7}{8}\selectfont, anchor=west,
                    inner sep=0pt},
      ptitle/.style={font=\sffamily\bfseries\fontsize{7.5}{8}\selectfont, anchor=west,
                     inner sep=0pt},
      stitle/.style={font=\sffamily\bfseries\fontsize{6.2}{7}\selectfont, anchor=west,
                     inner sep=0pt},
      vol/.style={draw=black!55, line width=0.3pt, fill=white, rounded corners=0.6pt,
                  minimum width=11pt, minimum height=11pt, inner sep=0pt,
                  font=\fontsize{5.2}{5.2}\selectfont},
      net/.style={draw=#1!75, fill=#1!12, line width=0.4pt, rounded corners=1pt,
                  minimum width=20pt, minimum height=12pt, inner sep=0pt,
                  font=\fontsize{5.8}{5.8}\selectfont},
      eq/.style={font=\sffamily\fontsize{5.2}{6}\selectfont, anchor=west, inner sep=0pt,
                 text=black!80},
      mlab/.style={font=\sffamily\fontsize{5}{5.5}\selectfont, inner sep=0.5pt, text=black!75},
    ]

    \def\xin{8}  \def\xout{164}
    \newcommand{\backbone}[6]{%
      \node[io] at (\xin,#1) {};
      \foreach \x/\h in {#2} \node[blk, minimum height=\h pt] at (\x,#1) {};
      \foreach \x/\h in {#3} \node[blk, minimum height=\h pt] at (\x,#1) {};
      \node[io] at (\xout,#1) {};
      \draw[flow] (\xin+4,#1) -- (20,#1);
      \draw[flow] (144,#1) -- (\xout-4,#1);
      \foreach \xa/\xb/\h in {#4}
        \draw[skiparr] (\xa,#1+\h/2) .. controls (\xa,#1+\h/2+9) and (\xb,#1+\h/2+9)
                               .. (\xb,#1+\h/2);
      \node[tag] at (\xin,#1+10) {phase};
      \node[tag] at (\xout,#1+10) {stain};
      \node[blab] at (43,#1-20) {#5};
      \node[blab] at (121,#1-20) {#6};
    }
    \def\rowA{0}  \def\rowB{-66}  \def\rowC{-132}

    \node[ptitle] at (0,48) {A\quad Backbones};

    \node[mname, text=mFNet] at (0,\rowA+31) {FNet3D};
    \backbone{\rowA}{25/32, 37/24, 49/17, 61/11}{103/11, 115/17, 127/24, 139/32}
      {25/139/32, 37/127/24, 49/115/17, 61/103/11}{Conv3d, BN\,+\,ReLU}{ConvT3d}
    \node[bneck] at (82,\rowA) {Conv};

    \node[mname] at (0,\rowB+31)
      {\textcolor{mVSC}{VSCyto3D}\textcolor{mAbl}{\,/\,UNeXt2}};
    \backbone{\rowB}{25/32, 44/24, 63/17, 82/11}{101/17, 120/24, 139/32}
      {25/139/32, 44/120/24, 63/101/17}
      {Conv3d stem, ConvNeXt V2 Conv2d}{Conv2d, pixel-shuffle head}

    \node[mname, text=black!75] at (0,\rowC+31) {U-Net\,+\,ViT};
    \backbone{\rowC}{25/32, 41/24, 57/17}{107/17, 123/24, 139/32}
      {25/139/32, 41/123/24, 57/107/17}{residual Conv3d}{ConvT3d}
    \node[vit] at (82,\rowC) {ViT};
    \node[io, draw=mCD!80, fill=mCD!10, densely dashed, minimum height=9pt]
      at (\xin,\rowC-16) {};
    \node[tag, text=mCD, anchor=north] at (\xin,\rowC-22) {noise};
    \draw[flow, draw=mCD!80] (\xin+4,\rowC-16) -| (20,\rowC-5);

    \draw[black!15, line width=0.4pt] (178,52) -- (178,-160);

    \def\xb{186}
    \node[ptitle] at (\xb-2,48) {B\quad Training on the U-Net\,+\,ViT backbone};

    \node[stitle, text=mAbl] at (\xb,34) {UNetViT3D};
    \node[vol] (u-p)  at (\xb+8,21)  {$\rvp$};
    \node[net=mAbl] (u-f) at (\xb+34,21) {$f_\theta$};
    \node[vol] (u-x)  at (\xb+60,21) {$\hat{\rvx}$};
    \draw[flow] (u-p) -- (u-f);  \draw[flow] (u-f) -- (u-x);
    \node[eq] at (\xb+74,21)
      {$\mathcal{L}_{\mathrm{MSE}}=\lVert\hat{\rvx}-\rvx\rVert_2^2$};

    \node[stitle, text=mP2P] at (\xb,3) {Pix2Pix3D};
    \node[vol] (g-p)  at (\xb+8,-12)  {$\rvp$};
    \node[net=mP2P] (g-G) at (\xb+34,-12) {$G$};
    \node[vol] (g-x)  at (\xb+60,-12) {$\hat{\rvx}$};
    \draw[flow] (g-p) -- (g-G);  \draw[flow] (g-G) -- (g-x);
    \node[net=mP2P, minimum width=22pt, minimum height=24pt] (g-D) at (\xb+150,-18) {$D$};
    \node[mlab, anchor=north] at (\xb+150,-31) {2 scales};
    \draw[flow] (g-x.east) -- node[mlab, above] {$\rvp\oplus\hat{\rvx}$ (fake)} (\xb+139,-12);
    \node[mlab, anchor=west] (g-real) at (\xb+68,-24) {$\rvp\oplus\rvx$ (real)};
    \draw[flow] (g-real.east) -- (\xb+139,-24);
    \draw[flow] (g-D.east) -- (\xb+172,-18);
    \node[mlab, anchor=west] at (\xb+173,-18) {real\,/\,fake};
    \node[eq] at (\xb,-42)
      {$\mathcal{L}_G=\mathrm{softplus}\bigl(-D(\rvp\oplus\hat{\rvx})\bigr)
        +\lambda_{L_1}\lVert\hat{\rvx}-\rvx\rVert_1$};
    \node[eq] at (\xb,-52)
      {$\mathcal{L}_D=\mathrm{softplus}\bigl(-D(\rvp\oplus\rvx)\bigr)
        +\mathrm{softplus}\bigl(D(\rvp\oplus\hat{\rvx})\bigr)
        +\mathrm{R1}+\mathrm{LeCam}$};

    \node[stitle, text=mCD] at (\xb,-68) {CELL-Diff3D};
    \node[eq] at (\xb+52,-68)
      {$\rvx_t=(1-t)\,\rvx_0+t\,\rvx_1,\quad t\sim\mathcal{U}(0,1)$};

    \newcommand{\fmtile}[3]{%
      \begin{scope}[shift={(#1,#2)}]
        \fill[white] (-9,-9) rectangle (9,9);
        \pgfmathsetmacro{\fmnoise}{1-#3}
        \fill[black!80, fill opacity=#3] (-3,2.2) ellipse (4.3 and 3.2);
        \fill[black!80, fill opacity=#3] (3.7,-3.6) ellipse (3.1 and 2.4);
        \pgfmathsetseed{11}
        \foreach \i in {1,...,60} {
          \pgfmathsetmacro{\px}{rnd*16-8}
          \pgfmathsetmacro{\py}{rnd*16-8}
          \pgfmathsetmacro{\pg}{20+rnd*65}
          \fill[black!\pg, fill opacity=\fmnoise] (\px,\py) circle (0.6);
        }
        \draw[black!55, line width=0.3pt] (-9,-9) rectangle (9,9);
      \end{scope}}
    \foreach \t/\x in {0/0, 0.25/1, 0.5/2, 0.75/3, 1/4}
      \fmtile{\xb+10+30*\x}{-88}{\t};
    \foreach \x in {0,1,2,3}
      \draw[flow, black!45] (\xb+21+30*\x,-88) -- (\xb+29+30*\x,-88);
    \node[mlab, anchor=north] at (\xb+10,-99)  {$\rvx_0$ (noise)};
    \node[mlab, anchor=north] at (\xb+40,-99)  {$t{=}0.25$};
    \node[mlab, anchor=north] at (\xb+70,-99)  {$0.5$};
    \node[mlab, anchor=north] at (\xb+100,-99) {$0.75$};
    \node[mlab, anchor=north] at (\xb+130,-99) {$\rvx_1$ (stain)};

    \node[mlab, anchor=west] (c-in) at (\xb,-117) {$\rvx_t,\ \rvp,\ t$};
    \node[net=mCD] (c-v) at (\xb+56,-117) {$\rvv_\theta$};
    \draw[flow] (c-in.east) -- (c-v.west);
    \draw[flow] (c-v.east) -- (\xb+86,-117);
    \node[mlab, anchor=west] at (\xb+87,-117) {regress onto $\rvx_1-\rvx_0$};
    \node[eq] at (\xb,-131)
      {$\mathcal{L}_{\mathrm{FM}}=\bigl\lVert\rvv_\theta(\rvx_t,\rvp,t)
        -(\rvx_1-\rvx_0)\bigr\rVert_2^2$};

  \end{tikzpicture}
  \caption{\textbf{Architectures and training objectives of the benchmark
    baselines.} \textbf{(A)} Three backbones cover the six trained models and
    differ mainly in how they handle the axial dimension. FNet3D
    \citep{ounkomol2018label} is a fully 3D U-Net of depth 4 (channels
    $32\!\to\!512$) with BatchNorm, ReLU, non-residual double convolutions, and a
    convolutional bottleneck; it halves $Z$ along with $Y$ and $X$, takes
    $32\!\times\!64\!\times\!64$ inputs, and is trained with MSE. VSCyto3D
    \citep{liu2025robust} folds the 15-plane input stack into channels at the stem,
    encodes it with a 2D ConvNeXt V2 (stages 3-3-9-3, channels $96\!\to\!768$) whose
    deepest stage feeds three decoder stages directly, and restores the stack with a
    pixel-shuffle head; UNeXt2 is the same network without FCMAE pretraining, and
    both take $15\!\times\!384^{2}$ inputs and are trained with MAE${+}$MS-SSIM. The
    third backbone preserves $Z$ with three levels of stride-$(1,2,2)$ convolutions
    (channels $64\!\to\!256$) around a vision-transformer bottleneck (patch 4, 8
    layers, width 512, 8 heads) and takes $8\!\times\!512^{2}$ inputs.
    \textbf{(B)} Three objectives on that backbone. UNetViT3D regresses the target
    with MSE. Pix2Pix3D adds a PatchGAN discriminator $D$ at two scales (2 input
    channels, base width 64, spectral norm) that scores the phase input
    concatenated with either the target or the generator output; the generator
    minimizes the non-saturating loss plus $\lambda_{L_1}{=}10$ times the $L_1$
    error, and the discriminator loss adds R1 and LeCam regularization.
    CELL-Diff3D draws $\rvx_0\sim\mathcal{N}(\mathbf{0},\rmI)$ and
    $t\sim\mathcal{U}(0,1)$, forms $\rvx_t$ on the linear path, and regresses the
    velocity $\rvv_\theta(\rvx_t,\rvp,t)$ onto $\rvx_1-\rvx_0$; the strip shows one
    noise draw at increasing $t$, and sampling integrates the ODE from $t{=}0$
    to 1. Dashed arrows denote skip connections and shaded blocks bottlenecks.
    Training all three objectives on one backbone makes
    App.~\ref{sec:matched-generator-ablation} a comparison of objectives rather
    than of capacity; hyperparameters are listed in
    Tab.~\ref{tab:benchmark_training}.}
  \label{fig:baseline-architectures}
\end{figure}

Fig.~\ref{fig:baseline-architectures} shows the architectures of the six trained models. Three backbones cover all of them: the fully 3D U-Net of FNet3D, the ConvNeXt V2 U-Net shared by VSCyto3D and UNeXt2, and the U-Net with a vision-transformer bottleneck shared by UNetViT3D, Pix2Pix3D, and CELL-Diff3D. Panel B illustrates how these three are trained.

\subsection{Shared training and augmentations}

The training, inference, and evaluation framework is built with open-source libraries and PyTorch Lightning. Experiments are tracked with Weights \& Biases. Training configurations are available at \url{https://github.com/Biohub/VisCy/tree/main/applications/dynacell}.

\paragraph{Auxiliary regression ablations.}
Tab.~\ref{tab:benchmark_training} lists two models beyond the four main baselines. Both are controls rather than benchmark entries: each is matched to one main baseline in everything except the single factor it isolates.
UNeXt2~\citep{liu2025robust} is the VSCyto3D ConvNeXt V2 U-Net trained from a random initialization. It matches VSCyto3D in parameter count ($32.1$M), schedule ($200$ epochs, batch $32$, learning rate $4\times10^{-4}$ with cosine decay), patch size ($15\times384\times384$), MAE$+$SSIM loss, and target normalization; the two differ only in that VSCyto3D initializes its encoder from the FCMAE checkpoint. UNeXt2 therefore isolates the contribution of encoder pretraining (App.~\ref{sec:vscyto3d-pretraining-ablation}).
UNetViT3D is trained with MSE on the same generator architecture that backs Pix2Pix3D and CELL-Diff3D ($77.8$M parameters, $20$ epochs, batch $4$, learning rate $3\times10^{-4}$ with cosine decay). These models hold the generator fixed while comparing reconstruction, adversarial, and flow-matching objectives~(App.~\ref{sec:matched-generator-ablation}). UNetViT3D also serves as the regression reference for CELL-Diff3D's single-pass conditional mean (Tab.~\ref{tab:gan-objective-ablation}).

\begin{table}[htbp]
\centering
\small
\caption{%
  Training pipeline comparison for the four main models and two auxiliary regression ablations.
  All models take 3D phase-contrast volumes as input and predict a single fluorescence channel.
}
\label{tab:benchmark_training}
\resizebox{\textwidth}{!}{
\begin{threeparttable}
\begin{tabular}{l c c c c c c c c c c}
\toprule
\multirow{2}{*}{\textbf{Model}}
  & \multicolumn{6}{c}{\textbf{Training}}
  & \multicolumn{4}{c}{\textbf{Data}} \\
\cmidrule(lr){2-7}\cmidrule(lr){8-11}
  & \textbf{Params} & \textbf{Pretrained} & \textbf{Duration} & \textbf{Batch} & \textbf{LR} & \textbf{LR schedule}
  & $\Delta z$ & $\Delta y/x$ & \textbf{Loss} & \textbf{Tgt.\ norm}\tnote{a} \\
\midrule
CELL-Diff3D & 93.4M & No & 20 ep & 4 & $3\times10^{-4}$ & Cosine & 8 & $512^{2}$ & Flow matching & min-max\tnote{c} \\
Pix2Pix3D & 77.8M+disc. & No & 40 ep & 4 & $2\times10^{-4}$ (G,D) & Warmup cosine & 8 & $512^{2}$ & Non-sat. GAN+$L_1$\tnote{d} & min-max\tnote{c} \\
UNetViT3D & 77.8M & No & 20 ep & 4 & $3\times10^{-4}$ & Cosine & 8 & $512^{2}$ & MSE & med/IQR \\
VSCyto3D  & 32.1M & Yes\tnote{b} & 200 ep & 32 & $4\times10^{-4}$ & Cosine & 15 & $384^{2}$ & MAE+SSIM & med/IQR \\
UNeXt2    & 32.1M & No & 200 ep & 32 & $4\times10^{-4}$ & Cosine & 15 & $384^{2}$ & MAE+SSIM & med/IQR \\
FNet3D    & 35.3M & No & 200k steps & 48 & $1\times10^{-3}$ & Constant & 32 & $64^{2}$ & MSE & mean/SD \\
\bottomrule
\end{tabular}
\begin{tablenotes}\footnotesize
  \item[a] med/IQR\,=\,(value\,$-$\,median)\,/\,IQR;\enspace
           mean/SD\,=\,(value\,$-$\,mean)\,/\,SD.
  \item[b] Encoder from VSCyto3D FCMAE checkpoint (400\,ep on HEK293, A549, iPSC phase);
           decoder randomly initialized.
  \item[c] min-max\,=\,clip to $[p_{1},p_{99}]$, then rescale to $[-1,1]$.
  \item[d] Pix2Pix3D checkpoints use the 40-epoch recipe: equal generator and discriminator learning rates, R1 and LeCam regularization, generator EMA, and $\lambda_{L_1}=10$.
\end{tablenotes}
\end{threeparttable}
}
\end{table}

\begin{table}[h]
\centering
\small
\caption{%
  Data augmentation summary for the four main models and two auxiliary regression ablations.
  \checkmark~indicates the augmentation is applied; $-$ indicates it is not used.
  Intensity augmentations are applied to the phase input only.
}
\label{tab:augmentation}
\resizebox{\linewidth}{!}{%
\begin{tabular}{l c c c c c c}
\toprule
\textbf{Augmentation} & \textbf{FNet3D} & \textbf{UNeXt2} & \textbf{VSCyto3D} & \textbf{UNetViT3D} & \textbf{Pix2Pix3D} & \textbf{CELL-Diff3D} \\
\midrule
Weighted random crop (CPU)               & \checkmark & \checkmark & \checkmark & \checkmark & \checkmark & \checkmark \\
Random flip, Y/X axes ($p=0.5$)          & \checkmark & $-$        & $-$        & $-$        & $-$        & $-$        \\
Random 3D affine ($p=0.8$)               & $-$        & \checkmark & \checkmark & \checkmark & \checkmark & \checkmark \\
\quad with safe-crop constraint          & $-$        & $-$        & $-$        & \checkmark & \checkmark & \checkmark \\
Gamma contrast adjustment ($p=0.5$)      & $-$        & \checkmark & \checkmark & \checkmark & \checkmark & \checkmark \\
Intensity scaling ($p=0.5$)              & $-$        & \checkmark & \checkmark & \checkmark & \checkmark & \checkmark \\
Gaussian noise ($p=0.5$)                 & $-$        & \checkmark & \checkmark & \checkmark & \checkmark & \checkmark \\
Gaussian smoothing ($p=0.5$)             & $-$        & \checkmark & \checkmark & \checkmark & \checkmark & \checkmark \\
\bottomrule
\end{tabular}
}
\end{table}
 
Augmentations are implemented using \texttt{viscy\_transforms}, a GPU-accelerated extension of MONAI~\citep{cardoso2022monai} transforms for batched 3D volumes.
The pipeline has two stages.

\paragraph{CPU stage (data loading).}
All models sample foreground-weighted random crops from each field of view, where the sampling probability at each voxel is proportional to the fluorescence target intensity.
FNet3D extracts eight small patches of size $32\times64\times64$ per FOV (matching the patch-based protocol of the original paper~\citep{ounkomol2018label}). UNeXt2 and VSCyto3D extract four oversized crops of $20\times600\times600$ to provide spatial margin for the subsequent affine transformation before center-cropping to $15\times384\times384$.
UNetViT3D, Pix2Pix3D, and CELL-Diff3D crop directly at the final patch size of $13\times512\times512$, using the affine safe-crop mechanism described below to avoid border artifacts.

\paragraph{GPU stage.}
For FNet3D, only independent Y- and X-axis random flips ($p=0.5$ each) are applied to both source (phase) and target (fluorescence).
For all other models, the GPU pipeline applies the following operations in sequence:
\begin{enumerate}[noitemsep]
  \item \textbf{Random 3D affine} ($p=0.8$): full $\pm\pi$ rotation about the Z-axis, shear $\pm0.05$ in Y/X, and anisotropic scaling ($0.7$--$1.3\times$ in Z, $0.5$--$1.5\times$ in Y/X), applied jointly to source and target. UNetViT3D, Pix2Pix3D, and CELL-Diff3D additionally enforce a safe-crop constraint that limits the scale factors so the rotated patch always covers the final crop, eliminating zero-filled corners. A center crop to the final patch size follows.
  \item \textbf{Gamma contrast adjustment} ($p=0.5$, $\gamma\in[0.8,1.2]$): applied to the phase input only.
  \item \textbf{Intensity scaling} ($p=0.5$, multiplicative factor $\in[0.5,1.5]$): applied to the phase input only.
  \item \textbf{Additive Gaussian noise} ($p=0.5$, $\sigma=0.3$): applied to the phase input only.
  \item \textbf{Gaussian smoothing} ($p=0.5$, $\sigma\in[0.25,0.75]$ isotropically in all three axes): applied to the phase input only.
\end{enumerate}
Operations (2)--(5) are applied only to the phase input, leaving the fluorescence target unchanged.
All reported Pix2Pix3D results use tiled inference with the EMA generator weights.

FNet3D follows the original publication~\citep{ounkomol2018label} and uses only random flips, without affine or intensity transformations. The nucleus ablation in~App.~\ref{sec:fnet_patch_aug_ablation} tests how this recipe contributes to its reduced OOD performance.

\subsection{FNet3D patch size and augmentation ablation}
\label{sec:fnet_patch_aug_ablation}

FNet3D differs from the other baselines in two ways: it trains on a $32\times64\times64$ patch, whereas the others use $384^2$ or $512^2$ in Y/X, and it applies only axis flips rather than the full augmentation sequence above. The cross-cell-type comparison therefore combines architecture and training-recipe effects. We test the recipe on the iPSC-trained nucleus model with two additional fits. Version A (\emph{$+$ big patch}) increases the patch to $32\times384\times384$, changes numerical precision from 32-bit to bf16 mixed, and retains flip-only augmentation. Version B (\emph{$+$ VSCyto3D aug}) adds the VSCyto3D augmentation sequence to version A. The versions otherwise share geometry, batch size, crop sampling, model, optimizer, loss, and step budget, so the A$\rightarrow$B comparison isolates augmentation. The baseline$\rightarrow$A comparison combines patch size and numerical precision.

\begin{figure}[!ht]
  \centering
  \includegraphics[width=1.0\linewidth]{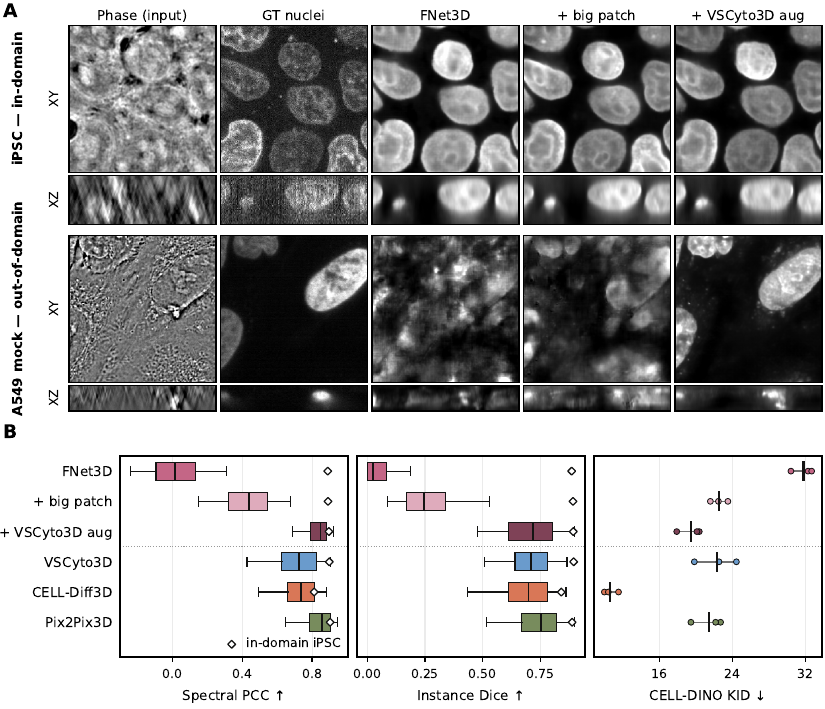}
  \caption{\textbf{FNet3D patch-size and augmentation ablation (nucleus, iPSC-trained).} \textbf{(A)} An in-focus XY plane with the orthogonal XZ slice directly below it at the same width, both shown at the physical aspect ratio, for the in-distribution iPSC test set (top) and OOD A549 Mock test set (bottom). Columns show the phase input, experimental fluorescence, and three FNet3D versions. Panels are clipped independently to the 1st--99.5th percentiles; each field is the case with median AP@$0.50$ for version B, not the best case. \textbf{(B)} Per-field Spectral PCC and instance Dice distributions pooled over the three A549 conditions (120 fields each, $n=360$): boxes show the median and interquartile range, whiskers show the 5th--95th percentiles, and diamonds show the in-distribution iPSC mean. CELL-DINO KID is a corpus-level scalar, so its panel shows the three per-condition values and their mean. Its in-distribution marker is omitted because KID uses a separate reference corpus for each dataset, making iPSC and A549 KID incomparable. Rows have the same order in all panels: the two additional versions above the dotted line and three published iPSC-trained baselines below.}
  \label{fig:patch_aug_ablation}
\end{figure}

Both versions preserve in-distribution scores and improve OOD scores (Fig.~\ref{fig:patch_aug_ablation}). On the held-out iPSC test set, the three versions have similar scores (Spectral PCC $0.890$, $0.892$, $0.897$; instance Dice $0.886$, $0.892$, $0.893$). Averaged over the three A549 conditions, Spectral PCC increases from $0.020$ (baseline) to $0.425$ (A) to $0.831$ (B). Over the same sequence, instance Dice increases from $0.052$ to $0.269$ to $0.701$, and CELL-DINO KID decreases from $31.8$ to $22.5$ to $19.4$. The baseline-to-A change and the A-to-B augmentation change contribute equally to Spectral PCC ($+0.405$ versus $+0.406$). For instance Dice, adding augmentations gives about twice the gain of the initial recipe change ($+0.432$ versus $+0.217$); for CELL-DINO KID, the initial change gives about three times the improvement. These results show that the original recipe contributes substantially to the transfer gap, but they do not rule out architectural effects. In the expanded appendix comparison of four main models, two regression ablations, and two FNet3D variants, version B ranks second on Spectral PCC, third on instance Dice, and third on CELL-DINO KID.

The ablation covers only the nucleus target and iPSC training set and is not a strict recipe match. The additional versions retain FNet3D's 32-plane input volume, compared with 15--20 planes for the other models, and use a batch size of $8$. This ablation estimates the importance of the training-recipe differences but does not establish FNet3D's best possible performance.

\subsection{FNet3D temporal-sampling ablation}
\label{sec:fnet-temporal-ablation}

\begin{figure}[ht]
  \centering
  \includegraphics[width=0.92\linewidth]{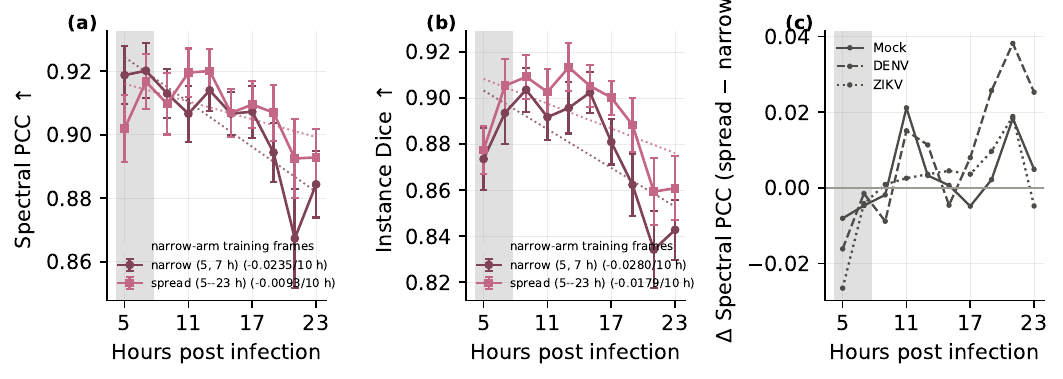}
  \caption{\textbf{FNet3D temporal-sampling ablation at a fixed frame budget.} A nucleus model trained on the narrow 5--7 hpi window is compared with one trained on frames spread over 5--23 hpi. \textbf{(a,b)} Spectral PCC and instance Dice over the paired A549 timepoints (mean $\pm$ 1 SEM across Mock, DENV, and ZIKV FOVs); the shaded region marks the narrow training window and dotted lines are least-squares slopes. \textbf{(c)} Per-condition Spectral-PCC difference between the spread and narrow models. Similar pooled scores mask the narrow model's steeper late-time decline.}
  \label{fig:fnet-temporal-sampling}
\end{figure}

At a fixed frame budget, we compare two A549-trained FNet3D nucleus models. One uses only the 5--7 hpi frames, while the other samples frames across the full time course. Their pooled Spectral PCC differs by only $0.0045$. However, the narrow-window model performs better within its training window, and its scores then decline $2.5\times$ faster in Spectral PCC and $1.6\times$ faster in instance Dice per 10 h~(Fig.~\ref{fig:fnet-temporal-sampling}). Thus, a pooled score can hide a failure to generalize to later timepoints.

\FloatBarrier
\subsection{Effect of FCMAE pretraining}
\label{sec:vscyto3d-pretraining-ablation}

UNeXt2 is the VSCyto3D architecture trained from scratch, whereas VSCyto3D initializes the same encoder with FCMAE pretraining; all other training choices are matched within each pool. Tab.~\ref{tab:vscyto3d-pretraining-ablation} shows that the effect is modest and organelle dependent. The clearest difference is iPSC membrane instance Dice, which increases from $0.589$ to $0.630$; results for nucleus, ER, and mitochondria are close or mixed.

\begin{table}[H]
\centering
\small
\caption{\textbf{VSCyto3D encoder-pretraining ablation.} Entries compare UNeXt2 (training from scratch) with VSCyto3D (FCMAE initialization) on the same training and test sets. A549 values average Mock, DENV, and ZIKV. Segmentation is instance Dice for nucleus and membrane and semantic Dice for ER and mitochondria; lower CELL-DINO KID is better.}
\label{tab:vscyto3d-pretraining-ablation}
\begin{tabular}{llccc}
\toprule
\textbf{Organelle} & \textbf{Domain} & \textbf{Spectral PCC} & \textbf{Dice} & \textbf{CELL-DINO KID} \\
\midrule
Membrane & iPSC & $0.878\rightarrow0.889$ & $0.589\rightarrow0.630$ & $44.98\rightarrow43.62$ \\
         & A549 & $0.874\rightarrow0.877$ & $0.724\rightarrow0.730$ & $28.28\rightarrow28.03$ \\
Nucleus  & iPSC & $0.901\rightarrow0.899$ & $0.888\rightarrow0.895$ & $24.68\rightarrow24.93$ \\
         & A549 & $0.897\rightarrow0.897$ & $0.895\rightarrow0.898$ & $32.17\rightarrow32.47$ \\
ER       & iPSC & $0.882\rightarrow0.883$ & $0.258\rightarrow0.262$ & $36.83\rightarrow36.64$ \\
         & A549 & $0.834\rightarrow0.838$ & $0.156\rightarrow0.147$ & $27.38\rightarrow28.34$ \\
Mitochondria & iPSC & $0.894\rightarrow0.893$ & $0.430\rightarrow0.428$ & $30.85\rightarrow30.45$ \\
         & A549 & $0.880\rightarrow0.884$ & $0.368\rightarrow0.368$ & $30.75\rightarrow30.64$ \\
\bottomrule
\end{tabular}
\end{table}
 
\subsection{Comparing objectives with the same generator}
\label{sec:matched-generator-ablation}

UNetViT3D is an auxiliary regression ablation that uses the same generator architecture as Pix2Pix3D and CELL-Diff3D. These controls compare regression, adversarial, and flow-matching objectives, including both single-pass and sampled CELL-Diff3D inference, while keeping the architecture and capacity fixed (Tab.~\ref{tab:gan-objective-ablation}).

\begin{table*}[t]
\centering
\scriptsize
\setlength{\tabcolsep}{2.5pt}
\caption{\textbf{Same-generator objective ablation.} Pix2Pix3D matches or exceeds its regression backbone on pixel and segmentation metrics, whereas sampled CELL-Diff3D has the lowest DINOv3 and CELL-DINO KID in every block. All models use the same UNetViT3D generator. Mean denotes the single-pass conditional-mean estimate at $t=0$ from the same flow-matching model used for sampling. Dice is instance Dice for nucleus and membrane and semantic Dice for ER and mitochondria. Within each organelle and train--test block, bold marks the best mean and values within one reported standard deviation of it. Pixel, segmentation, and GLCM+ values are mean~$\pm$~SD over image pairs; KID values are dataset-level means~$\pm$~SD over resampled subsets.}
\label{tab:gan-objective-ablation}
\resizebox{\textwidth}{!}{%
\begin{tabular}{lccccccc}
\toprule
\textbf{\shortstack{Organelle\\(train $\rightarrow$ test)}} & \textbf{Model} & \textbf{si-SSIM $\uparrow$} & \textbf{\shortstack{Spectral\\PCC $\uparrow$}} & \textbf{Dice $\uparrow$} & \textbf{\shortstack{DINOv3\\KID $\downarrow$}} & \textbf{\shortstack{CELL-DINO\\KID $\downarrow$}} & \textbf{\shortstack{GLCM+\\cos $\uparrow$}} \\
\midrule
\shortstack{\textbf{Nucleus}\\(iPSC $\rightarrow$ iPSC)} & UNetViT3D & \textbf{0.54 $\pm$ 0.16} & \textbf{0.89 $\pm$ 0.07} & \textbf{0.86 $\pm$ 0.08} & 2.01 $\pm$ 0.04 & 23.34 $\pm$ 0.17 & \textbf{-0.00 $\pm$ 0.43} \\
 & Pix2Pix3D & \textbf{0.58 $\pm$ 0.17} & \textbf{0.90 $\pm$ 0.06} & \textbf{0.89 $\pm$ 0.08} & 1.82 $\pm$ 0.06 & 23.99 $\pm$ 0.24 & \textbf{-0.01 $\pm$ 0.45} \\
 & \shortstack{CELL-Diff3D\\(mean)} & \textbf{0.54 $\pm$ 0.16} & \textbf{0.87 $\pm$ 0.07} & \textbf{0.85 $\pm$ 0.09} & 1.77 $\pm$ 0.03 & 27.71 $\pm$ 0.21 & \textbf{0.00 $\pm$ 0.43} \\
 & \shortstack{CELL-Diff3D\\(sampled)} & \textbf{0.47 $\pm$ 0.17} & 0.81 $\pm$ 0.09 & \textbf{0.84 $\pm$ 0.08} & \textbf{0.08 $\pm$ 0.02} & \textbf{1.72 $\pm$ 0.05} & \textbf{0.26 $\pm$ 0.41} \\
\shortstack{\textbf{Membrane}\\(iPSC $\rightarrow$ iPSC)} & UNetViT3D & \textbf{0.57 $\pm$ 0.12} & \textbf{0.88 $\pm$ 0.14} & \textbf{0.66 $\pm$ 0.10} & 2.57 $\pm$ 0.05 & 34.52 $\pm$ 0.24 & \textbf{0.22 $\pm$ 0.46} \\
 & Pix2Pix3D & \textbf{0.60 $\pm$ 0.12} & \textbf{0.88 $\pm$ 0.15} & \textbf{0.69 $\pm$ 0.11} & 2.62 $\pm$ 0.05 & 31.30 $\pm$ 0.24 & \textbf{0.26 $\pm$ 0.45} \\
 & \shortstack{CELL-Diff3D\\(mean)} & \textbf{0.54 $\pm$ 0.12} & \textbf{0.85 $\pm$ 0.14} & \textbf{0.60 $\pm$ 0.10} & 2.77 $\pm$ 0.04 & 41.50 $\pm$ 0.30 & \textbf{0.18 $\pm$ 0.45} \\
 & \shortstack{CELL-Diff3D\\(sampled)} & 0.47 $\pm$ 0.13 & \textbf{0.82 $\pm$ 0.14} & 0.52 $\pm$ 0.17 & \textbf{0.33 $\pm$ 0.02} & \textbf{4.22 $\pm$ 0.09} & \textbf{0.28 $\pm$ 0.35} \\
\shortstack{\textbf{ER}\\(iPSC $\rightarrow$ iPSC)} & UNetViT3D & \textbf{0.35 $\pm$ 0.08} & \textbf{0.88 $\pm$ 0.07} & \textbf{0.23 $\pm$ 0.05} & 3.84 $\pm$ 0.06 & 35.19 $\pm$ 0.20 & \textbf{0.16 $\pm$ 0.31} \\
 & Pix2Pix3D & \textbf{0.38 $\pm$ 0.08} & \textbf{0.90 $\pm$ 0.07} & \textbf{0.27 $\pm$ 0.04} & 3.99 $\pm$ 0.06 & 32.13 $\pm$ 0.26 & \textbf{0.16 $\pm$ 0.34} \\
 & \shortstack{CELL-Diff3D\\(mean)} & \textbf{0.31 $\pm$ 0.08} & \textbf{0.85 $\pm$ 0.08} & 0.22 $\pm$ 0.05 & 4.23 $\pm$ 0.06 & 37.87 $\pm$ 0.23 & \textbf{0.16 $\pm$ 0.28} \\
 & \shortstack{CELL-Diff3D\\(sampled)} & 0.23 $\pm$ 0.09 & 0.79 $\pm$ 0.09 & \textbf{0.23 $\pm$ 0.02} & \textbf{0.13 $\pm$ 0.02} & \textbf{3.16 $\pm$ 0.06} & \textbf{0.36 $\pm$ 0.35} \\
\shortstack{\textbf{Mitochondria}\\(iPSC $\rightarrow$ iPSC)} & UNetViT3D & \textbf{0.55 $\pm$ 0.06} & \textbf{0.92 $\pm$ 0.03} & \textbf{0.45 $\pm$ 0.04} & 5.66 $\pm$ 0.14 & 29.12 $\pm$ 0.17 & 0.15 $\pm$ 0.42 \\
 & Pix2Pix3D & \textbf{0.58 $\pm$ 0.06} & \textbf{0.92 $\pm$ 0.02} & \textbf{0.47 $\pm$ 0.04} & 4.03 $\pm$ 0.06 & 35.40 $\pm$ 0.22 & 0.11 $\pm$ 0.45 \\
 & \shortstack{CELL-Diff3D\\(mean)} & \textbf{0.52 $\pm$ 0.06} & \textbf{0.90 $\pm$ 0.03} & \textbf{0.44 $\pm$ 0.04} & 5.71 $\pm$ 0.13 & 27.65 $\pm$ 0.17 & 0.17 $\pm$ 0.38 \\
 & \shortstack{CELL-Diff3D\\(sampled)} & 0.43 $\pm$ 0.06 & 0.85 $\pm$ 0.03 & 0.32 $\pm$ 0.03 & \textbf{0.07 $\pm$ 0.01} & \textbf{3.13 $\pm$ 0.13} & \textbf{0.61 $\pm$ 0.28} \\
\shortstack{\textbf{ER}\\(A549 $\rightarrow$ A549 Mock)} & UNetViT3D & \textbf{0.51 $\pm$ 0.08} & \textbf{0.83 $\pm$ 0.06} & \textbf{0.12 $\pm$ 0.04} & 2.40 $\pm$ 0.07 & 27.43 $\pm$ 0.24 & \textbf{0.09 $\pm$ 0.46} \\
 & Pix2Pix3D & \textbf{0.53 $\pm$ 0.08} & \textbf{0.83 $\pm$ 0.06} & \textbf{0.12 $\pm$ 0.04} & 2.16 $\pm$ 0.06 & 19.97 $\pm$ 0.18 & \textbf{0.07 $\pm$ 0.47} \\
 & \shortstack{CELL-Diff3D\\(sampled)} & 0.42 $\pm$ 0.08 & 0.71 $\pm$ 0.10 & \textbf{0.14 $\pm$ 0.03} & \textbf{0.51 $\pm$ 0.06} & \textbf{4.26 $\pm$ 0.09} & \textbf{0.18 $\pm$ 0.39} \\
\shortstack{\textbf{Mitochondria}\\(A549 $\rightarrow$ A549 Mock)} & UNetViT3D & \textbf{0.72 $\pm$ 0.08} & \textbf{0.90 $\pm$ 0.04} & \textbf{0.36 $\pm$ 0.07} & 4.84 $\pm$ 0.11 & 21.86 $\pm$ 0.17 & \textbf{0.16 $\pm$ 0.45} \\
 & Pix2Pix3D & \textbf{0.76 $\pm$ 0.07} & \textbf{0.92 $\pm$ 0.03} & \textbf{0.38 $\pm$ 0.08} & 2.94 $\pm$ 0.05 & 32.08 $\pm$ 0.26 & \textbf{0.14 $\pm$ 0.51} \\
 & \shortstack{CELL-Diff3D\\(sampled)} & 0.64 $\pm$ 0.08 & 0.81 $\pm$ 0.06 & 0.25 $\pm$ 0.03 & \textbf{0.75 $\pm$ 0.05} & \textbf{6.72 $\pm$ 0.10} & \textbf{0.31 $\pm$ 0.47} \\
\bottomrule
\end{tabular}%
}
\end{table*}
 
\subsection{CELL-Diff3D flow-matching baseline}
\label{sec:celldiff_flow_matching}

\subsubsection{Training}
CELL-Diff3D is trained as a conditional flow-matching model~\citep{zheng2025bridging,lipman2023flow}. For each phase volume \(\rvp\) and paired fluorescence target \(\rvx\), let \(\rvx_1=\rvx\), sample noise \(\rvx_0 \sim \mathcal{N}(0,\rmI)\), and sample time \(t \sim \mathcal{U}(0,1)\). Linear flow-matching defines the interpolated state and target velocity as
\[
    \rvx_t = (1-t)\rvx_0 + t\rvx_1, \qquad \rvv_t = \rvx_1 - \rvx_0.
\]
The model learns a conditional velocity field \(\rvv_{\theta}(\rvx_t,t,\rvp)\) by minimizing
\begin{equation}
    \mathcal{L}_{\mathrm{FM}}
    = \E_{\rvp,\rvx_1,\rvx_0,t}
    \left[ \left\| \rvv_{\theta}(\rvx_t,t,\rvp) - \rvv_t \right\|_2^2 \right].
    \label{eq:flow_matching_loss}
\end{equation}
At inference, a noise volume is sampled and the learned probability-flow ODE is integrated from \(t=0\) to \(t=1\) to generate a fluorescence sample. Our 3D adaptation uses volumetric phase and fluorescence inputs and shares its backbone with the auxiliary UNetViT3D regression ablation, allowing regression-versus-generative comparisons that are not dominated by backbone differences.

\subsubsection{Iterative prediction with flow-matching}
\label{sec:iter_pred_fm}

\begin{figure}[h]
  \centering
  \includegraphics[width=\textwidth]{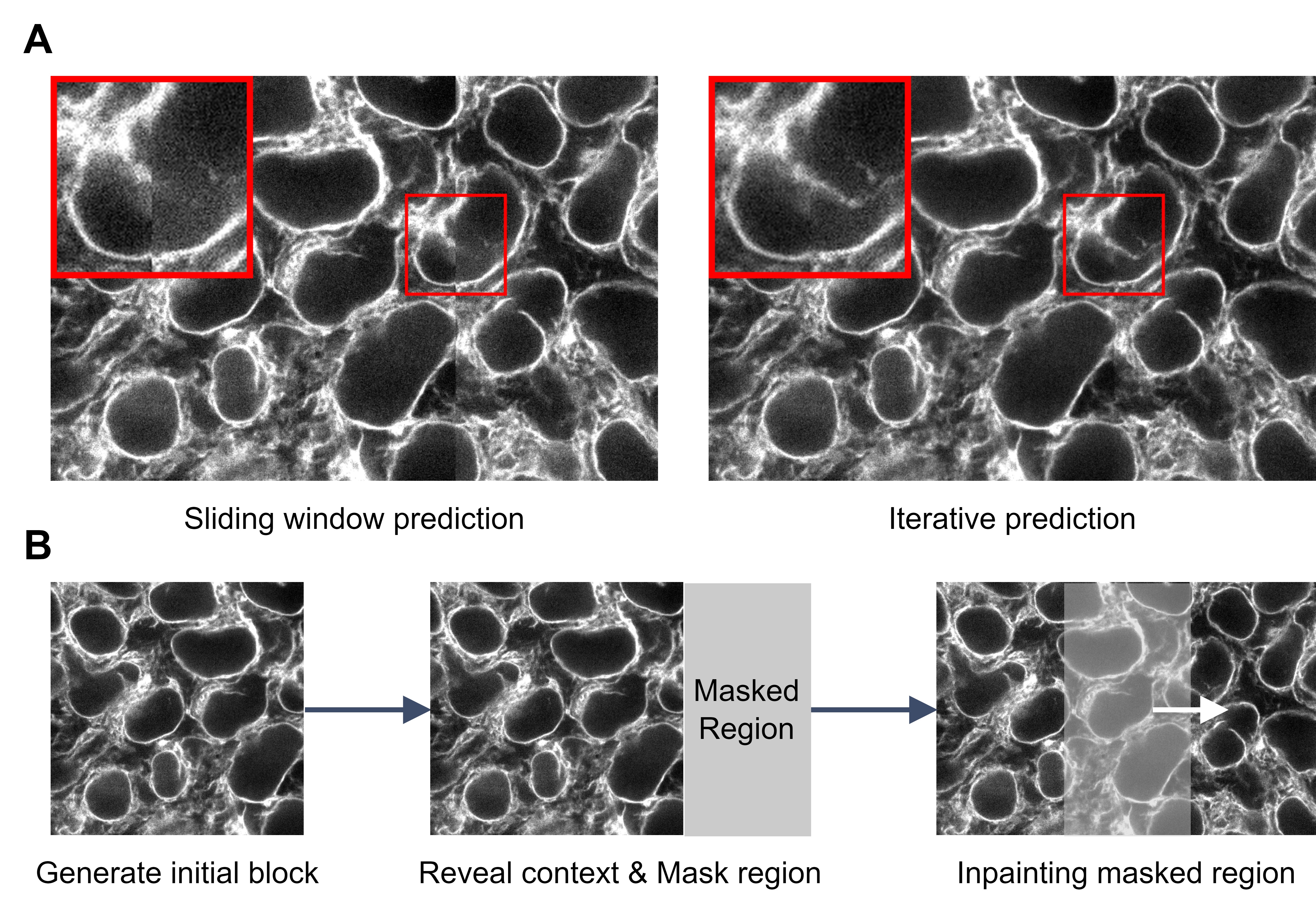}
  \caption{\textbf{Iterative flow-matching prediction for large-field virtual staining.}
  \textbf{(A)} Sliding-window prediction introduces a visible boundary artifact because neighboring tiles are sampled independently from the conditional distribution \(p(\rvx \mid \rvp)\), whereas iterative prediction preserves continuity across the overlap region. \textbf{(B)} Starting from an initial generated block, the window shifts to expose a new masked region while retaining the overlapping context. Flow-matching inpainting fills the masked region conditioned on this context and the phase input.}
  \label{fig:iterative_prediction_flow_matching}
\end{figure}

A standard way to process large microscopy fields is to apply a predictor to overlapping windows and stitch the outputs. This approach is less effective for conditional generative models because each window's prediction is an independent sample from \(p(\rvx \mid \rvp)\), rather than the output of a single-valued function such as a conditional mean. Neighboring windows may therefore represent different samples and produce visible seams when stitched (Fig.~\ref{fig:iterative_prediction_flow_matching}A, left).

We instead generate tiles sequentially with flow-matching inpainting. Each new prediction is conditioned on the phase input \(\rvp\) and the overlap with the previous fluorescence prediction, preserving continuity while filling the unseen region.

At each step, the local window contains an observed overlap and an unobserved region to be synthesized. Let \(\rvx_{\mathrm{obs}}\) denote the partially observed fluorescence volume and let \(\rvm \in \{0,1\}^{\Omega}\) be a binary mask of the observed voxels. The task is to sample from
\[
p(\rvx \mid \rvp, \rvx_{\mathrm{obs}}, \rvm).
\]

Both diffusion and flow-matching models define a time-dependent transformation from an initial noise distribution \(\rvx_0 \sim \gN(\mathbf{0}, \rmI)\) to a target sample \(\rvx_1 \sim p_{\mathrm{data}}(\rvx \mid \rvp)\). A broad family of such processes can be written as
\begin{equation}
    \rvx_t = \alpha_t \rvx_1 + \sigma_t \rvx_0,
\end{equation}
where \(\alpha_t\) increases from \(0\) to \(1\) and \(\sigma_t\) decreases from \(1\) to \(0\) as \(t\) goes from \(0\) to \(1\). For linear flow-matching,
\[
\alpha_t = t, \qquad \sigma_t = 1-t,
\]
so that
\[
\rvx_t = t\,\rvx_1 + (1-t)\,\rvx_0.
\]

The associated conditional probability-flow ODE is
\[
\frac{d\rvx_t}{dt} = \rvv(\rvx_t,t,\rvp), \qquad \rvx_{t=0} = \rvx_0,
\]
where the velocity field is
\begin{equation}
    \rvv(\rvx,t,\rvp)
    = \E\!\left[\dot{\rvx}_t \mid \rvx_t=\rvx,\rvp\right]
    = \dot{\alpha}_t \E\!\left[\rvx_1 \mid \rvx_t=\rvx,\rvp\right]
      + \dot{\sigma}_t \E\!\left[\rvx_0 \mid \rvx_t=\rvx,\rvp\right].
\end{equation}
Conditioning on \(\rvx_t=\rvx\) and \(\rvp\) also gives
\begin{equation}
    \rvx
    = \E\!\left[\rvx_t \mid \rvx_t=\rvx,\rvp\right]
    = \alpha_t \E\!\left[\rvx_1 \mid \rvx_t=\rvx,\rvp\right]
      + \sigma_t \E\!\left[\rvx_0 \mid \rvx_t=\rvx,\rvp\right].
\end{equation}

Therefore, solving the linear system
\[
\begin{cases}
\rvx &= \alpha_t \E[\rvx_1 \mid \rvx_t=\rvx,\rvp] + \sigma_t \E[\rvx_0 \mid \rvx_t=\rvx,\rvp], \\[4pt]
\rvv(\rvx,t,\rvp) &= \dot{\alpha}_t \E[\rvx_1 \mid \rvx_t=\rvx,\rvp] + \dot{\sigma}_t \E[\rvx_0 \mid \rvx_t=\rvx,\rvp],
\end{cases}
\]
yields
\begin{equation}
    \begin{cases}
        \E\!\left[\rvx_0 \mid \rvx_t=\rvx,\rvp\right]
        =
        \dfrac{\alpha_t \,\rvv(\rvx,t,\rvp) - \dot{\alpha}_t \,\rvx}
        {\alpha_t \dot{\sigma}_t - \sigma_t \dot{\alpha}_t}, \\[10pt]
        \E\!\left[\rvx_1 \mid \rvx_t=\rvx,\rvp\right]
        =
        \dfrac{\dot{\sigma}_t \,\rvx - \sigma_t \,\rvv(\rvx,t,\rvp)}
        {\alpha_t \dot{\sigma}_t - \sigma_t \dot{\alpha}_t}.
    \end{cases}
    \label{eq:conditional_x0_x1_p}
\end{equation}

We now use this identity to construct conditional inpainting dynamics. In the unknown region, we keep the original pretrained velocity field \(\rvv(\rvx,t,\rvp)\) unchanged. In the known region, we enforce the endpoint constraint
\[
\E[\rvx_1 \mid \rvx_t=\rvx,\rvp] = \rvx_{\mathrm{obs}},
\]
so that the terminal state matches the observed fluorescence context. Let \(\rvv^\star\) denote the calibrated velocity field in the observed region. Using
\[
\rvv^\star(\rvx,t,\rvp)
= \dot{\alpha}_t \rvx_{\mathrm{obs}}
+ \dot{\sigma}_t \E[\rvx_0 \mid \rvx_t=\rvx,\rvp],
\]
together with Eq.~\eqref{eq:conditional_x0_x1_p}, we obtain
\begin{equation}
    \rvv^\star(\rvx,t,\rvp)
    =
    \dot{\alpha}_t \rvx_{\mathrm{obs}}
    +
    \dot{\sigma}_t
    \frac{\alpha_t \,\rvv(\rvx,t,\rvp) - \dot{\alpha}_t \,\rvx}
         {\alpha_t \dot{\sigma}_t - \sigma_t \dot{\alpha}_t}.
    \label{eq:general_calibrated_velocity_p}
\end{equation}

For linear flow-matching, where \(\alpha_t=t\) and \(\sigma_t=1-t\), we have \(\dot{\alpha}_t=1\) and \(\dot{\sigma}_t=-1\). Eq.~\eqref{eq:conditional_x0_x1_p} then simplifies to
\begin{equation}
    \begin{cases}
        \E\!\left[\rvx_0 \mid \rvx_t=\rvx,\rvp\right] = \rvx - t\,\rvv(\rvx,t,\rvp), \\[4pt]
        \E\!\left[\rvx_1 \mid \rvx_t=\rvx,\rvp\right] = \rvx + (1-t)\,\rvv(\rvx,t,\rvp).
    \end{cases}
    \label{eq:conditional_x0_x1_p_linear_path}
\end{equation}
Substituting these expressions into Eq.~\eqref{eq:general_calibrated_velocity_p} gives
\begin{equation}
    \rvv^\star(\rvx,t,\rvp)
    = \rvx_{\mathrm{obs}} - \bigl(\rvx - t\,\rvv(\rvx,t,\rvp)\bigr)
    = t\,\rvv(\rvx,t,\rvp) + \rvx_{\mathrm{obs}} - \rvx.
\end{equation}

Finally, we combine the known and unknown regions with a binary mask \(\rvm \in \{0,1\}^{\Omega}\) and define the calibrated velocity field elementwise as
\begin{equation}
\rvv^\star(\rvx,t,\rvp)_q =
\begin{cases}
\rvv(\rvx,t,\rvp)_q, & \text{if } \rvm_q = 0, \\[4pt]
t\,\rvv(\rvx,t,\rvp)_q + {\rvx_{\mathrm{obs}}}_q - \rvx_q, & \text{if } \rvm_q = 1,
\end{cases}
\label{eq:masked_calibrated_velocity_p}
\end{equation}
for each voxel \(q \in \Omega\).

Starting from \(\rvx_0 \sim \gN(\mathbf{0}, \rmI)\), we integrate the conditional ODE
\[
\frac{d\rvx_t}{dt} = \rvv^\star(\rvx_t,t,\rvp), \qquad \rvx_{t=0} = \rvx_0,
\]
from \(t=0\) to \(t=1\) to obtain the final sample \(\rvx_1\). The dynamics follow the original conditional generative flow \(p(\rvx \mid \rvp)\) in unobserved regions. In observed regions, they are constrained to reach the prescribed context \(\rvx_{\mathrm{obs}}\) at \(t=1\) (Alg.~\ref{alg:iterative_prediction_flow_matching}). Applied sequentially to overlapping windows, this procedure extends prediction beyond the native input size while maintaining spatial continuity (Fig.~\ref{fig:iterative_prediction_flow_matching}).

\begin{algorithm}
\caption{Iterative prediction with flow matching}
\label{alg:iterative_prediction_flow_matching}
\begin{algorithmic}[1]
\Require Phase input \(\rvp\), pretrained conditional velocity field \(\rvv_{\theta}(\cdot,\cdot,\rvp)\), initial window, stride, overlap mask rule
\Ensure Large-field fluorescence prediction \(\hat{\rvx}\)

\State Generate the first tile by sampling from \(p_{\theta}(\rvx \mid \rvp)\) in the initial window
\State Place this tile into the output canvas \(\hat{\rvx}\)

\For{each subsequent window along the traversal path}
    \State Extract the local phase crop \(\rvp_{\mathrm{loc}}\)
    \State Construct a partial observation \(\hat{\rvx}_{\mathrm{loc}}\) from the overlap with previously generated tiles
    \State Build a binary mask \(\rvm\) indicating observed overlap voxels
    \State Sample an initial noise state \(\rvx_0 \sim \gN(\mathbf{0}, \rmI)\)
    \State Define the calibrated velocity field
    \[
    \rvv^\star(\rvx,t,\rvp_{\mathrm{loc}})_q =
    \begin{cases}
    \rvv_{\theta}(\rvx,t,\rvp_{\mathrm{loc}})_q, & \text{if } \rvm_q = 0, \\[4pt]
    t\,\rvv_{\theta}(\rvx,t,\rvp_{\mathrm{loc}})_q + \hat{\rvx}_{\mathrm{loc},q} - \rvx_q, & \text{if } \rvm_q = 1
    \end{cases}
    \]
    \State Solve
    \[
    \frac{d\rvx_t}{dt} = \rvv^\star(\rvx_t,t,\rvp_{\mathrm{loc}}), \qquad \rvx_{t=0}=\rvx_0
    \]
    from \(t=0\) to \(t=1\) to obtain the next tile
    \State Write the newly generated non-overlap region into the output canvas \(\hat{\rvx}\)
\EndFor

\State \Return \(\hat{\rvx}\)
\end{algorithmic}
\end{algorithm}

Tab.~\ref{tab:iterative_vs_sliding} compares the two inference strategies across four organelles on the iPSC test set. Iterative prediction gives small improvements in Spectral PCC and Dice on all four organelles, whereas CELL-DINO KID is better for sliding-window inference. The model is trained on \(8 \times 512 \times 512\) patches, while the iPSC test volumes are \(40 \times 512 \times 512\). Because the in-plane field matches training, stitching artifacts are mostly limited to the axial direction in this evaluation. We nevertheless use iterative prediction because it reduces visible seams by conditioning each block on previously generated context (Fig.~\ref{fig:iterative_prediction_flow_matching}a).

\begin{table}[ht]
\centering
\small
\caption{\textbf{Comparison between CELL-Diff3D Iterative and Sliding-Window inference.}
  Spectral PCC, Dice (instance Dice for nucleus and membrane; semantic
  Dice for ER and mitochondria), and CELL-DINO KID across four
  organelles on the iPSC held-out test set (100 FOVs).}
\label{tab:iterative_vs_sliding}
\begin{tabular}{llccc}
\toprule
\textbf{Organelle} & \textbf{Inference} & \textbf{Spectral PCC} $\uparrow$ & \textbf{Dice} $\uparrow$ & \textbf{CELL-DINO KID} $\downarrow$ \\
\midrule
\multirow{2}{*}{Cell Membrane} & Iterative & $\mathbf{0.819}\pm0.14$ & $\mathbf{0.516}\pm0.17$ & $4.220\pm0.09$ \\
 & Sliding-Window & $0.771\pm0.17$ & $0.467\pm0.20$ & $\mathbf{2.747}\pm0.07$ \\
\midrule
\multirow{2}{*}{Nucleus} & Iterative & $\mathbf{0.812}\pm0.09$ & $\mathbf{0.841}\pm0.08$ & $1.720\pm0.05$ \\
 & Sliding-Window & $0.797\pm0.09$ & $0.826\pm0.09$ & $\mathbf{1.042}\pm0.03$ \\
\midrule
\multirow{2}{*}{ER (SEC61B)} & Iterative & $\mathbf{0.790}\pm0.09$ & $\mathbf{0.233}\pm0.02$ & $3.164\pm0.06$ \\
 & Sliding-Window & $0.765\pm0.11$ & $0.230\pm0.02$ & $\mathbf{2.204}\pm0.05$ \\
\midrule
\multirow{2}{*}{Mito.\ (TOMM20)} & Iterative & $\mathbf{0.850}\pm0.03$ & $\mathbf{0.319}\pm0.03$ & $3.128\pm0.13$ \\
 & Sliding-Window & $0.838\pm0.04$ & $0.314\pm0.03$ & $\mathbf{1.975}\pm0.07$ \\
\bottomrule
\end{tabular}
\end{table}
 
\subsubsection{Mean prediction with flow-matching}
\label{sec:mean_pred_fm}

CELL-Diff3D can also produce a mean prediction. The most direct approach is Monte Carlo averaging: draw multiple samples
\[
\rvx^{(1)},\ldots,\rvx^{(K)} \sim p_{\theta}(\rvx \mid \rvp),
\]
and estimate the conditional mean by
\[
\bar{\rvx}
=
\frac{1}{K}\sum_{k=1}^{K}\rvx^{(k)}.
\]
As \(K \to \infty\), this converges to \(\E[\rvx \mid \rvp]\). Each sample, however, requires solving the probability-flow ODE from noise to data, making repeated sampling substantially slower than a single forward pass through a point predictor.

Eq.~\eqref{eq:conditional_x0_x1_p} shows that the posterior conditional mean of the target image can be recovered directly from the learned velocity field. For a general interpolation path,
\begin{equation}
    \E\!\left[\rvx_1 \mid \rvx_t=\rvx,\rvp\right]
    =
    \dfrac{\dot{\sigma}_t \,\rvx - \sigma_t \,\rvv(\rvx,t,\rvp)}
    {\alpha_t \dot{\sigma}_t - \sigma_t \dot{\alpha}_t}.
    \label{eq:mean_prediction_general}
\end{equation}
In particular, at \(t=0\), where \(\rvx_t=\rvx_0\), this becomes
\begin{equation}
    \E\!\left[\rvx_1 \mid \rvx_0=\rvx,\rvp\right]
    =
    \dfrac{\dot{\sigma}_0 \,\rvx - \sigma_0 \,\rvv(\rvx,0,\rvp)}
    {\alpha_0 \dot{\sigma}_0 - \sigma_0 \dot{\alpha}_0}.
    \label{eq:mean_prediction_t0_general}
\end{equation}

For the linear path used in our flow-matching model,
\[
\alpha_t = t, \qquad \sigma_t = 1-t,
\]
so that
\[
\alpha_0 = 0,\quad \sigma_0 = 1,\quad \dot{\alpha}_0 = 1,\quad \dot{\sigma}_0 = -1.
\]
Substituting these into Eq.~\eqref{eq:mean_prediction_t0_general} yields
\begin{equation}
    \E\!\left[\rvx_1 \mid \rvx_0=\rvx,\rvp\right]
    =
    \rvx + \rvv(\rvx,0,\rvp).
    \label{eq:mean_prediction_linear_t0}
\end{equation}
Because \(\rvx_0\) is independent of \((\rvx_1,\rvp)\), this conditional expectation equals \(\E[\rvx_1\mid\rvp]\) for the population-optimal field. Thus, one velocity-field evaluation at \(t=0\) gives a conditional-mean estimator and a single-pass alternative to Monte Carlo averaging (Alg.~\ref{alg:mean_prediction_flow_matching}).

\begin{algorithm}
\caption{Mean prediction with flow matching}
\label{alg:mean_prediction_flow_matching}
\begin{algorithmic}[1]
\Require Phase input \(\rvp\), pretrained velocity field \(\rvv_{\theta}\).
\Ensure Mean prediction \(\hat{\rvx}_{\mathrm{mean}}\)

\State Sample an initial noise state \(\rvx_0 \sim \gN(\mathbf{0}, \rmI)\)
\State Compute
\[
\hat{\rvx}_{\mathrm{mean}}
\gets
\rvx_0 + \rvv_{\theta}(\rvx_0,0,\rvp)
\]
\State \Return \(\hat{\rvx}_{\mathrm{mean}}\)
\end{algorithmic}
\end{algorithm}

\begin{figure}
    \centering
    \includegraphics[width=1.0\linewidth]{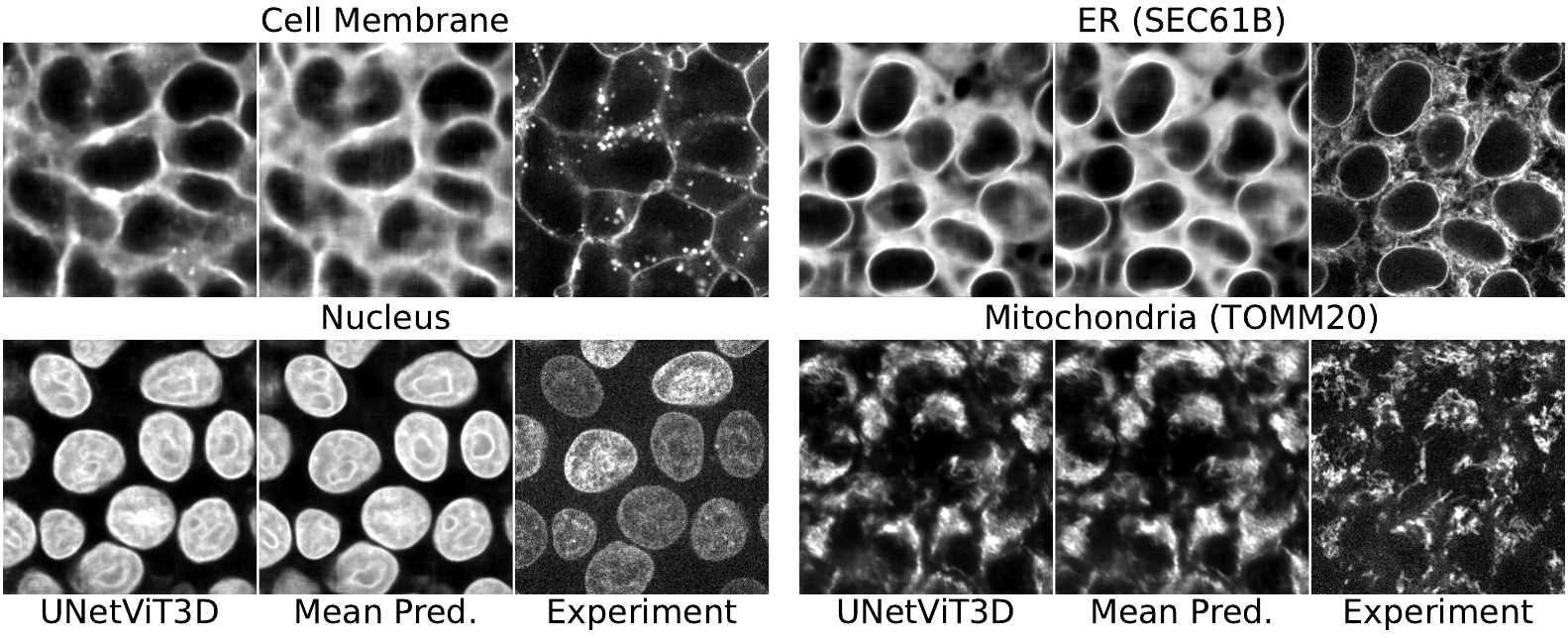}
    \caption{Visual comparison between UNetViT3D and CELL-Diff3D mean prediction.}
    \label{fig:celldiff_mean_pred_vs_unetvit3d}
\end{figure}

The iPSC results in Tab.~\ref{tab:gan-objective-ablation} compare the auxiliary UNetViT3D regression ablation with CELL-Diff3D mean prediction across all four organelles~(Fig.~\ref{fig:celldiff_mean_pred_vs_unetvit3d}). Their pixel and Dice scores are similar, with the largest Dice difference for membrane. UNetViT3D is trained directly with MSE, which encourages conditional-mean prediction, whereas CELL-Diff3D obtains the mean indirectly from the \(t=0\) velocity field. The same CELL-Diff3D model can produce either this single-pass estimate or diverse stochastic samples through ODE integration, without additional training.

\subsubsection{CELL-Diff3D sampling and denoising trajectories}
\label{sec:celldiff_traj}

CELL-Diff3D generates a fluorescence volume by integrating the learned flow-matching velocity field as an ordinary differential equation (ODE). This transports a Gaussian-noise sample \(\rvx_0\sim\gN(\mathbf{0},\rmI)\) at \(t=0\) to a predicted image \(\rvx_1\) at \(t=1\). We refer to the intermediate ODE states \(\{\rvx_t\}_{t\in[0,1]}\) produced during this integration as the \emph{sampling trajectory}.

{\bf Denoising trajectory.} At any intermediate time \(t\), the velocity field also yields the model's current estimate of the clean target, obtained by extrapolating the state to the endpoint \(t=1\). Specializing the endpoint estimator of Eq.~\eqref{eq:conditional_x0_x1_p_linear_path} to the learned velocity gives:

\begin{equation}
    \hat{\rvx}_1(\rvx_t,t)
    \;=\;
    \E\!\left[\rvx_1 \mid \rvx_t,\rvp\right]
    \;=\;
    \rvx_t + (1-t)\,\rvv_{\theta}(\rvx_t,t,\rvp).
    \label{eq:denoising_trajectory}
\end{equation}
We call the sequence \(\{\hat{\rvx}_1(\rvx_t,t)\}_{t\in[0,1]}\), evaluated along the sampling trajectory, the \emph{denoising trajectory}. Unlike the raw state \(\rvx_t\), each \(\hat{\rvx}_1\) is a full-resolution estimate of the fluorescence image: at \(t=0\) it reduces to the single-pass mean predictor of Eq.~\eqref{eq:mean_prediction_linear_t0}, and at \(t=1\) it coincides with the final sample \(\rvx_1\).

\begin{figure}
    \centering
    \includegraphics[width=1.0\linewidth]{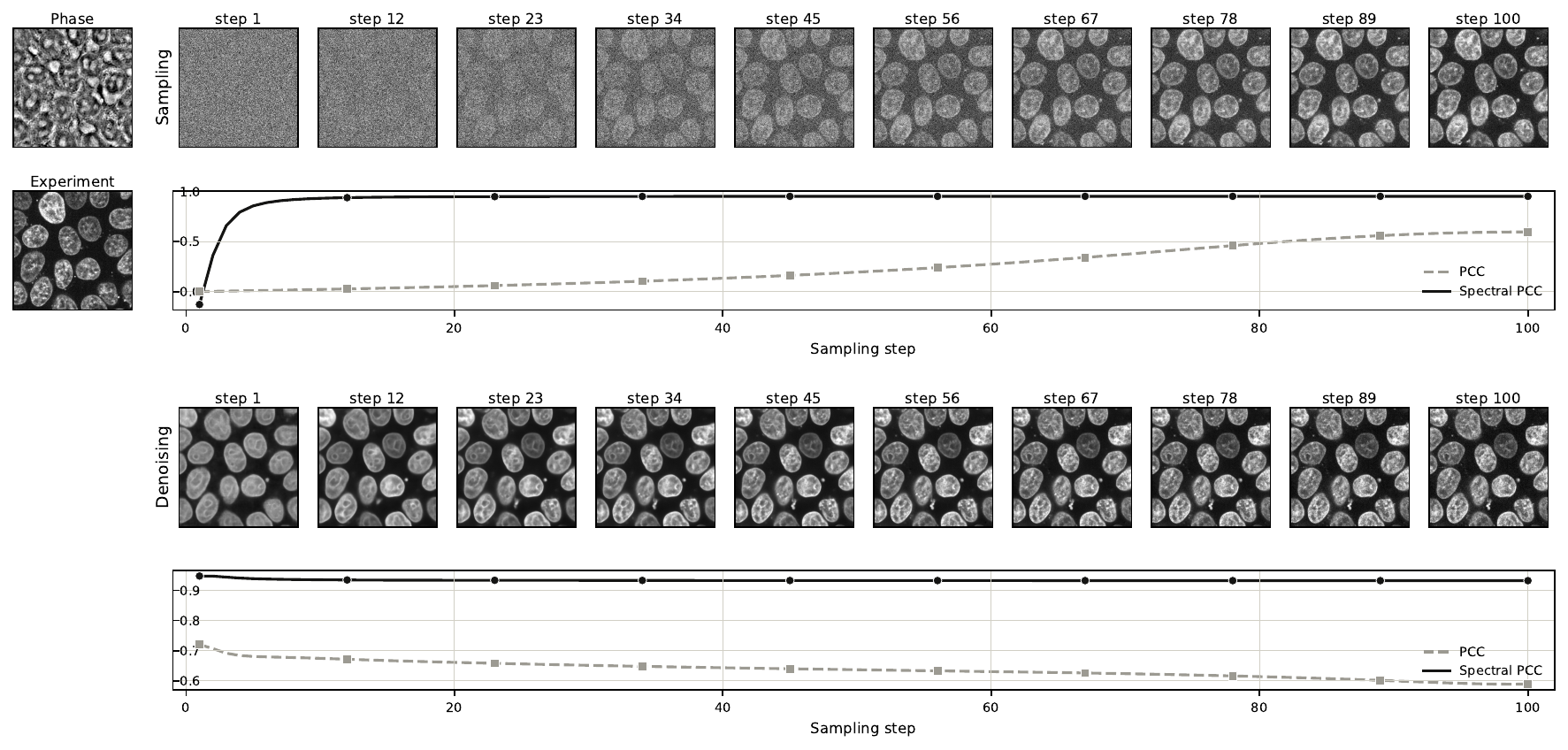}
    \caption{CELL-Diff3D sampling and denoising trajectories for nucleus virtual staining on an iPSC confocal field of view. \emph{Row 1:} phase input followed by the sampling trajectory at ten uniformly spaced steps of the $100$-step flow-matching ODE. \emph{Row 2:} experimental target and the evolution of pixel metrics (PCC and Spectral PCC) over the sampling trajectory. \emph{Row 3:} the denoising trajectory at the same steps. \emph{Row 4:} pixel metrics over the denoising trajectory.}
    \label{fig:celldiff_sample_traj}
\end{figure}

We save every integration step and evaluate the sampling state \(\rvx_t\) and denoised estimate \(\hat{\rvx}_1(\rvx_t,t)\) against the experimental target. Fig.~\ref{fig:celldiff_sample_traj} shows nucleus trajectories for a representative iPSC FOV over the \(100\)-step ODE. The sampling trajectory begins as noise and resolves into individual nuclei, with PCC and Spectral PCC increasing during integration. The denoising trajectory already recovers coherent nuclei at the first step, where it attains its highest PCC and Spectral PCC in this example. As integration proceeds, \(\hat{\rvx}_1\) becomes increasingly conditioned on the sampled Gaussian noise and synthesizes sample-specific high-frequency detail. This plausible detail need not match the paired experimental realization, so both pixel metrics decline; Spectral PCC declines more gradually because it downweights noise-dominated high frequencies. The sampling and denoising trajectories converge as \(t\to 1\), where \(\hat{\rvx}_1\) and \(\rvx_1\) coincide.

\subsection{Compute costs of training the models}
\label{sec:compute-cost}

All reported models and ablations were trained on the Biohub HPC cluster under SLURM, using NVIDIA H100, H200, L40S, A6000, A40, and A100 accelerators. We report accelerator-hours (wall time multiplied by the recorded accelerator count), so multi-GPU runs are not counted as single-device jobs. As of August 5, 2026, the accessible Weights \& Biases project contained 309 training attempts across the iPSC-only, A549-only, and joint pools, including Pix2Pix3D and superseded or restarted runs. These attempts account for \textbf{17{,}506.4 accelerator-hours}: 15{,}122.7 in runs whose logger terminated cleanly, 1{,}282.9 in crashed runs, 444.8 in failed runs, and 656.0 in seven four-GPU continuations that were still running at the audit snapshot.

\paragraph{Completion audit and lower-bound scope.}
The W\&B state \texttt{finished} records clean logger termination, not necessarily completion of the configured training goal. Cross-checking configured epoch targets and SLURM accounting identifies runs that stopped early, including canceled and timed-out jobs; later reruns are used where available. Separately, 839 prediction and re-prediction SLURM jobs submitted between March 1 and August 5, 2026, account for another 5{,}584.7 accelerator-hours. The combined tracked project cost is therefore at least \textbf{23{,}091.2 accelerator-hours}. This is a lower bound: a legacy W\&B workspace is no longer queryable, and the audit omits jobs submitted by other users and evaluation or preprocessing jobs whose names do not identify them as prediction runs.

\paragraph{Inference and evaluation compute.}
Inference and evaluation use cpdino-ViT-L for nucleus and membrane instances, Allen Institute organelle workflows for ER and mitochondria, GLCM+ for handcrafted 3D features, and DINOv3, DynaCLR, CELL-DINO, and MorphEm for learned embeddings. Their costs are not all logged in W\&B; the prediction jobs captured above provide a lower bound, while segmentation and feature extraction are omitted where SLURM job names do not identify them reliably.

\FloatBarrier

\section{Extended results}

\subsection{Results across cell types and microscopes}
\label{sec:training-pool-matrix}

Tab.~\ref{tab:training-matrix-three-tiers} reports iPSC-only, A549-only, and joint training results on the iPSC and A549 test sets for VSCyto3D, Pix2Pix3D, and CELL-Diff3D. Each row combines representative pixel, segmentation, and phenotype metrics from the same final evaluation.

\definecolor{tierpixel}{RGB}{46,139,131}
\definecolor{tierseg}{RGB}{201,162,39}
\definecolor{tierpheno}{RGB}{74,90,115}
\begin{table*}[h]
\centering
\scriptsize
\caption{\textbf{Results for three training sets and two test sets.} Rows show Spectral PCC $\uparrow$, Dice $\uparrow$, and CELL-DINO KID $\downarrow$ for each organelle. Model columns group versions trained on iPSC, A549, or both; panels separate the iPSC and A549 Mock test sets. Dice is instance Dice for nucleus and membrane and semantic Dice for ER and mitochondria. For each training--test condition, bold entries are within one reported standard deviation of the best of the three models. Colored squares denote the metric tier: \textcolor{tierpixel}{\rule{1.1ex}{1.1ex}}~pixel, \textcolor{tierseg}{\rule{1.1ex}{1.1ex}}~segmentation, \textcolor{tierpheno}{\rule{1.1ex}{1.1ex}}~phenotype.}
\label{tab:training-matrix-three-tiers}
\resizebox{\textwidth}{!}{%
\begin{tabular}{l ccc ccc ccc}
\toprule
\multicolumn{10}{l}{\textbf{(A) iPSC test set}} \\
 & \multicolumn{3}{c}{\textbf{VSCyto3D}} & \multicolumn{3}{c}{\textbf{Pix2Pix3D}} & \multicolumn{3}{c}{\textbf{CELL-Diff3D}} \\
\cmidrule(lr){2-4}\cmidrule(lr){5-7}\cmidrule(lr){8-10}
\textbf{Metric} & \textbf{\shortstack{iPSC$\rightarrow$\\iPSC}} & \textbf{\shortstack{A549$\rightarrow$\\iPSC}} & \textbf{\shortstack{Joint$\rightarrow$\\iPSC}} & \textbf{\shortstack{iPSC$\rightarrow$\\iPSC}} & \textbf{\shortstack{A549$\rightarrow$\\iPSC}} & \textbf{\shortstack{Joint$\rightarrow$\\iPSC}} & \textbf{\shortstack{iPSC$\rightarrow$\\iPSC}} & \textbf{\shortstack{A549$\rightarrow$\\iPSC}} & \textbf{\shortstack{Joint$\rightarrow$\\iPSC}} \\
\midrule
\multicolumn{10}{l}{\textbf{Nucleus}} \\
  \textcolor{tierpixel}{\rule{1.1ex}{1.1ex}}~Spectral PCC ($\uparrow$) & \textbf{0.899 $\pm$ 0.06} & \textbf{0.557 $\pm$ 0.15} & \textbf{0.904 $\pm$ 0.06} & \textbf{0.904 $\pm$ 0.06} & \textbf{0.592 $\pm$ 0.14} & \textbf{0.892 $\pm$ 0.06} & 0.812 $\pm$ 0.09 & 0.377 $\pm$ 0.15 & 0.805 $\pm$ 0.09 \\
  \textcolor{tierseg}{\rule{1.1ex}{1.1ex}}~Dice ($\uparrow$) & \textbf{0.895 $\pm$ 0.07} & \textbf{0.631 $\pm$ 0.12} & \textbf{0.891 $\pm$ 0.07} & \textbf{0.888 $\pm$ 0.08} & \textbf{0.635 $\pm$ 0.11} & \textbf{0.888 $\pm$ 0.07} & \textbf{0.841 $\pm$ 0.08} & 0.394 $\pm$ 0.09 & \textbf{0.834 $\pm$ 0.08} \\
  \textcolor{tierpheno}{\rule{1.1ex}{1.1ex}}~CELL-DINO KID ($\downarrow$) & 24.93 $\pm$ 0.22 & 30.62 $\pm$ 0.19 & 24.91 $\pm$ 0.24 & 23.99 $\pm$ 0.24 & 29.14 $\pm$ 0.23 & 16.40 $\pm$ 0.14 & \textbf{1.720 $\pm$ 0.05} & \textbf{12.28 $\pm$ 0.11} & \textbf{1.652 $\pm$ 0.05} \\
\midrule
\multicolumn{10}{l}{\textbf{Membrane}} \\
  \textcolor{tierpixel}{\rule{1.1ex}{1.1ex}}~Spectral PCC ($\uparrow$) & \textbf{0.889 $\pm$ 0.11} & 0.256 $\pm$ 0.33 & \textbf{0.895 $\pm$ 0.10} & \textbf{0.880 $\pm$ 0.15} & \textbf{0.530 $\pm$ 0.15} & \textbf{0.882 $\pm$ 0.14} & \textbf{0.819 $\pm$ 0.14} & \textbf{0.486 $\pm$ 0.20} & \textbf{0.829 $\pm$ 0.14} \\
  \textcolor{tierseg}{\rule{1.1ex}{1.1ex}}~Dice ($\uparrow$) & \textbf{0.630 $\pm$ 0.11} & \textbf{0.386 $\pm$ 0.16} & \textbf{0.714 $\pm$ 0.10} & \textbf{0.689 $\pm$ 0.11} & \textbf{0.262 $\pm$ 0.15} & \textbf{0.694 $\pm$ 0.10} & 0.516 $\pm$ 0.17 & \textbf{0.258 $\pm$ 0.16} & 0.538 $\pm$ 0.17 \\
  \textcolor{tierpheno}{\rule{1.1ex}{1.1ex}}~CELL-DINO KID ($\downarrow$) & 43.62 $\pm$ 0.30 & 36.80 $\pm$ 0.27 & 36.76 $\pm$ 0.28 & 31.30 $\pm$ 0.24 & 34.94 $\pm$ 0.23 & 27.11 $\pm$ 0.22 & \textbf{4.220 $\pm$ 0.09} & \textbf{16.13 $\pm$ 0.15} & \textbf{3.583 $\pm$ 0.09} \\
\midrule
\multicolumn{10}{l}{\textbf{ER}} \\
  \textcolor{tierpixel}{\rule{1.1ex}{1.1ex}}~Spectral PCC ($\uparrow$) & \textbf{0.883 $\pm$ 0.08} & \textbf{0.470 $\pm$ 0.18} & \textbf{0.882 $\pm$ 0.09} & \textbf{0.895 $\pm$ 0.07} & \textbf{0.526 $\pm$ 0.15} & \textbf{0.884 $\pm$ 0.07} & 0.790 $\pm$ 0.09 & \textbf{0.507 $\pm$ 0.15} & 0.780 $\pm$ 0.12 \\
  \textcolor{tierseg}{\rule{1.1ex}{1.1ex}}~Dice ($\uparrow$) & \textbf{0.262 $\pm$ 0.04} & 0.013 $\pm$ 0.02 & \textbf{0.261 $\pm$ 0.04} & \textbf{0.269 $\pm$ 0.04} & 0.027 $\pm$ 0.02 & \textbf{0.272 $\pm$ 0.04} & \textbf{0.233 $\pm$ 0.02} & \textbf{0.102 $\pm$ 0.02} & 0.228 $\pm$ 0.02 \\
  \textcolor{tierpheno}{\rule{1.1ex}{1.1ex}}~CELL-DINO KID ($\downarrow$) & 36.64 $\pm$ 0.23 & 40.07 $\pm$ 0.22 & 37.04 $\pm$ 0.22 & 32.13 $\pm$ 0.26 & 38.81 $\pm$ 0.26 & 26.63 $\pm$ 0.26 & \textbf{3.164 $\pm$ 0.06} & \textbf{16.86 $\pm$ 0.15} & \textbf{2.375 $\pm$ 0.06} \\
\midrule
\multicolumn{10}{l}{\textbf{Mitochondria}} \\
  \textcolor{tierpixel}{\rule{1.1ex}{1.1ex}}~Spectral PCC ($\uparrow$) & 0.893 $\pm$ 0.03 & 0.526 $\pm$ 0.13 & 0.895 $\pm$ 0.03 & \textbf{0.924 $\pm$ 0.02} & \textbf{0.651 $\pm$ 0.08} & \textbf{0.920 $\pm$ 0.02} & 0.850 $\pm$ 0.03 & 0.552 $\pm$ 0.08 & 0.838 $\pm$ 0.04 \\
  \textcolor{tierseg}{\rule{1.1ex}{1.1ex}}~Dice ($\uparrow$) & 0.428 $\pm$ 0.03 & 0.067 $\pm$ 0.04 & \textbf{0.432 $\pm$ 0.03} & \textbf{0.472 $\pm$ 0.04} & \textbf{0.186 $\pm$ 0.04} & \textbf{0.457 $\pm$ 0.04} & 0.319 $\pm$ 0.03 & 0.124 $\pm$ 0.02 & 0.304 $\pm$ 0.03 \\
  \textcolor{tierpheno}{\rule{1.1ex}{1.1ex}}~CELL-DINO KID ($\downarrow$) & 30.45 $\pm$ 0.19 & 37.70 $\pm$ 0.21 & 31.25 $\pm$ 0.19 & 35.40 $\pm$ 0.22 & 32.57 $\pm$ 0.19 & 28.05 $\pm$ 0.18 & \textbf{3.128 $\pm$ 0.13} & \textbf{21.84 $\pm$ 0.20} & \textbf{6.038 $\pm$ 0.19} \\
\midrule
\addlinespace[0.6em]
\multicolumn{10}{l}{\textbf{(B) A549 Mock test set}} \\
 & \multicolumn{3}{c}{\textbf{VSCyto3D}} & \multicolumn{3}{c}{\textbf{Pix2Pix3D}} & \multicolumn{3}{c}{\textbf{CELL-Diff3D}} \\
\cmidrule(lr){2-4}\cmidrule(lr){5-7}\cmidrule(lr){8-10}
\textbf{Metric} & \textbf{\shortstack{A549$\rightarrow$\\A549}} & \textbf{\shortstack{iPSC$\rightarrow$\\A549}} & \textbf{\shortstack{Joint$\rightarrow$\\A549}} & \textbf{\shortstack{A549$\rightarrow$\\A549}} & \textbf{\shortstack{iPSC$\rightarrow$\\A549}} & \textbf{\shortstack{Joint$\rightarrow$\\A549}} & \textbf{\shortstack{A549$\rightarrow$\\A549}} & \textbf{\shortstack{iPSC$\rightarrow$\\A549}} & \textbf{\shortstack{Joint$\rightarrow$\\A549}} \\
\midrule
\multicolumn{10}{l}{\textbf{Nucleus}} \\
  \textcolor{tierpixel}{\rule{1.1ex}{1.1ex}}~Spectral PCC ($\uparrow$) & \textbf{0.892 $\pm$ 0.06} & 0.699 $\pm$ 0.13 & \textbf{0.900 $\pm$ 0.06} & \textbf{0.911 $\pm$ 0.06} & \textbf{0.855 $\pm$ 0.07} & \textbf{0.913 $\pm$ 0.06} & \textbf{0.875 $\pm$ 0.07} & 0.722 $\pm$ 0.11 & \textbf{0.891 $\pm$ 0.06} \\
  \textcolor{tierseg}{\rule{1.1ex}{1.1ex}}~Dice ($\uparrow$) & \textbf{0.904 $\pm$ 0.05} & \textbf{0.694 $\pm$ 0.10} & \textbf{0.909 $\pm$ 0.05} & \textbf{0.918 $\pm$ 0.05} & \textbf{0.731 $\pm$ 0.12} & \textbf{0.917 $\pm$ 0.05} & 0.763 $\pm$ 0.09 & \textbf{0.669 $\pm$ 0.15} & 0.844 $\pm$ 0.08 \\
  \textcolor{tierpheno}{\rule{1.1ex}{1.1ex}}~CELL-DINO KID ($\downarrow$) & 30.88 $\pm$ 0.09 & 19.80 $\pm$ 0.06 & 29.53 $\pm$ 0.09 & 26.04 $\pm$ 0.11 & 19.43 $\pm$ 0.08 & 20.10 $\pm$ 0.09 & \textbf{4.163 $\pm$ 0.03} & \textbf{9.947 $\pm$ 0.04} & \textbf{4.039 $\pm$ 0.03} \\
\midrule
\multicolumn{10}{l}{\textbf{Membrane}} \\
  \textcolor{tierpixel}{\rule{1.1ex}{1.1ex}}~Spectral PCC ($\uparrow$) & \textbf{0.883 $\pm$ 0.04} & \textbf{0.567 $\pm$ 0.11} & \textbf{0.889 $\pm$ 0.04} & \textbf{0.890 $\pm$ 0.04} & \textbf{0.566 $\pm$ 0.10} & \textbf{0.891 $\pm$ 0.04} & 0.833 $\pm$ 0.05 & \textbf{0.632 $\pm$ 0.10} & 0.828 $\pm$ 0.06 \\
  \textcolor{tierseg}{\rule{1.1ex}{1.1ex}}~Dice ($\uparrow$) & \textbf{0.713 $\pm$ 0.11} & \textbf{0.363 $\pm$ 0.11} & \textbf{0.698 $\pm$ 0.14} & \textbf{0.693 $\pm$ 0.13} & \textbf{0.394 $\pm$ 0.13} & \textbf{0.695 $\pm$ 0.13} & \textbf{0.629 $\pm$ 0.13} & 0.166 $\pm$ 0.11 & \textbf{0.622 $\pm$ 0.14} \\
  \textcolor{tierpheno}{\rule{1.1ex}{1.1ex}}~CELL-DINO KID ($\downarrow$) & 24.96 $\pm$ 0.06 & 39.36 $\pm$ 0.07 & 25.83 $\pm$ 0.06 & 19.97 $\pm$ 0.05 & 19.86 $\pm$ 0.06 & 18.64 $\pm$ 0.05 & \textbf{1.853 $\pm$ 0.02} & \textbf{8.818 $\pm$ 0.03} & \textbf{2.141 $\pm$ 0.02} \\
\midrule
\multicolumn{10}{l}{\textbf{ER}} \\
  \textcolor{tierpixel}{\rule{1.1ex}{1.1ex}}~Spectral PCC ($\uparrow$) & \textbf{0.819 $\pm$ 0.06} & \textbf{0.675 $\pm$ 0.11} & \textbf{0.828 $\pm$ 0.06} & \textbf{0.827 $\pm$ 0.06} & \textbf{0.721 $\pm$ 0.09} & \textbf{0.832 $\pm$ 0.06} & 0.714 $\pm$ 0.10 & \textbf{0.696 $\pm$ 0.10} & 0.731 $\pm$ 0.09 \\
  \textcolor{tierseg}{\rule{1.1ex}{1.1ex}}~Dice ($\uparrow$) & \textbf{0.140 $\pm$ 0.03} & 0.095 $\pm$ 0.02 & \textbf{0.147 $\pm$ 0.03} & \textbf{0.125 $\pm$ 0.04} & 0.098 $\pm$ 0.02 & \textbf{0.141 $\pm$ 0.05} & \textbf{0.141 $\pm$ 0.03} & \textbf{0.148 $\pm$ 0.03} & \textbf{0.140 $\pm$ 0.04} \\
  \textcolor{tierpheno}{\rule{1.1ex}{1.1ex}}~CELL-DINO KID ($\downarrow$) & 28.33 $\pm$ 0.22 & 23.82 $\pm$ 0.24 & 28.61 $\pm$ 0.23 & 19.97 $\pm$ 0.18 & 19.07 $\pm$ 0.19 & 22.04 $\pm$ 0.22 & \textbf{4.259 $\pm$ 0.09} & \textbf{5.803 $\pm$ 0.08} & \textbf{4.731 $\pm$ 0.10} \\
\midrule
\multicolumn{10}{l}{\textbf{Mitochondria}} \\
  \textcolor{tierpixel}{\rule{1.1ex}{1.1ex}}~Spectral PCC ($\uparrow$) & 0.887 $\pm$ 0.04 & \textbf{0.793 $\pm$ 0.05} & \textbf{0.898 $\pm$ 0.04} & \textbf{0.919 $\pm$ 0.03} & \textbf{0.821 $\pm$ 0.05} & \textbf{0.919 $\pm$ 0.03} & 0.814 $\pm$ 0.06 & 0.754 $\pm$ 0.05 & 0.841 $\pm$ 0.06 \\
  \textcolor{tierseg}{\rule{1.1ex}{1.1ex}}~Dice ($\uparrow$) & \textbf{0.366 $\pm$ 0.06} & \textbf{0.222 $\pm$ 0.06} & \textbf{0.380 $\pm$ 0.07} & \textbf{0.380 $\pm$ 0.08} & \textbf{0.247 $\pm$ 0.05} & \textbf{0.390 $\pm$ 0.08} & 0.246 $\pm$ 0.03 & \textbf{0.203 $\pm$ 0.04} & 0.265 $\pm$ 0.03 \\
  \textcolor{tierpheno}{\rule{1.1ex}{1.1ex}}~CELL-DINO KID ($\downarrow$) & 30.24 $\pm$ 0.23 & 33.81 $\pm$ 0.23 & 29.51 $\pm$ 0.23 & 32.08 $\pm$ 0.26 & 33.59 $\pm$ 0.25 & 27.13 $\pm$ 0.24 & \textbf{6.724 $\pm$ 0.10} & \textbf{12.06 $\pm$ 0.17} & \textbf{5.675 $\pm$ 0.11} \\
\bottomrule
\end{tabular}%
}
\end{table*}
 
\paragraph{HEK293T evaluation.}
\label{sec:hek-ablation-detail}
Tabs.~\ref{tab:hek-cell-type-decomposition} and~\ref{tab:hek-cell-type-summary} evaluate FNet3D and VSCyto3D on HEK293T without using HEK293T for training or checkpoint selection. When both cell types are imaged on Mantis, the mean Spectral PCC decrease from A549 to HEK293T is 0.180. This compares with a decrease of 0.514 from A549 to iPSC, when both cell type and microscope change. These decreases compare absolute scores across test sets, but the Spectral PCC of the raw phase input, a no-model floor, differs by at most $0.07$ between the three test sets (Tab.~\ref{tab:hek-cell-type-summary}). This comparison suggests that both factors contribute to the OOD gap, although the different fluorescence markers prevent a complete separation of their effects. Joint training gives the highest HEK293T Spectral PCC for both organelles.

\begin{table}[t]
\centering
\small
\caption{\textbf{Two-model decomposition of the iPSC--A549 transfer gap.} For A549-trained FNet3D and VSCyto3D, the mean Spectral PCC drop is 0.180 from A549 to HEK with the microscope held fixed and 0.514 from A549 to iPSC with cell type and microscope both changed. The 35\% value is based on these two models; estimating it across all four main models requires results for the remaining models.}
\label{tab:hek-cell-type-decomposition}
\resizebox{0.72\linewidth}{!}{%
\begin{tabular}{lccc}
\toprule
\textbf{Organelle} & \textbf{Model} & \textbf{\shortstack{Drop to HEK\\(cell type)}} & \textbf{\shortstack{Drop to iPSC\\(cell type + scope)}} \\
\midrule
Cell membrane & FNet3D & +0.302 & +0.479 \\
Cell membrane & VSCyto3D & +0.175 & +0.627 \\
Mitochondria & FNet3D & +0.143 & +0.589 \\
Mitochondria & VSCyto3D & +0.099 & +0.360 \\
\bottomrule
\end{tabular}%
}
\end{table}
\begin{table*}[t]
\centering
\scriptsize
\caption{\textbf{HEK293T cell-type ablation with a fixed microscope.} Spectral PCC is shown for models used as trained, without HEK data or checkpoint reselection. HEK and A549 share the Mantis microscope; iPSC images were acquired by confocal microscopy. HEK comprises four $64\times960\times1184$ ZYX volumes per organelle (about 120 cells), reused from S-BIAD1287. Marker changes (KRAS versus CAAX; TOMM70A versus TOMM20) make the HEK gap an upper bound on the cell-type effect. Bold marks training domains.}
\label{tab:hek-cell-type-summary}
\resizebox{\textwidth}{!}{%
\begin{tabular}{lcccccccc}
\toprule
\textbf{Organelle} & \textbf{Test set} & \textbf{\shortstack{FNet3D\\iPSC}} & \textbf{\shortstack{FNet3D\\A549}} & \textbf{\shortstack{FNet3D\\Joint}} & \textbf{\shortstack{VSCyto3D\\iPSC}} & \textbf{\shortstack{VSCyto3D\\A549}} & \textbf{\shortstack{VSCyto3D\\Joint}} & \textbf{\shortstack{Raw phase\\no model}} \\
\midrule
\shortstack{\textbf{Cell membrane}\\KRAS on HEK,\\CAAX elsewhere} & iPSC (confocal) & \textbf{0.894} & 0.399 & \textbf{0.876} & \textbf{0.889} & 0.256 & \textbf{0.895} & \emph{-0.001} \\
 & A549 (Mantis) & 0.594 & \textbf{0.878} & \textbf{0.882} & 0.567 & \textbf{0.883} & \textbf{0.889} & \emph{0.015} \\
 & HEK (Mantis) & 0.642 & 0.576 & 0.692 & 0.709 & 0.708 & 0.765 & \emph{0.044} \\
\shortstack{\textbf{Mitochondria}\\TOMM70A on HEK,\\TOMM20 elsewhere} & iPSC (confocal) & \textbf{0.910} & 0.331 & \textbf{0.903} & \textbf{0.893} & 0.526 & \textbf{0.895} & \emph{0.139} \\
 & A549 (Mantis) & 0.223 & \textbf{0.921} & \textbf{0.922} & 0.793 & \textbf{0.887} & \textbf{0.898} & \emph{0.176} \\
 & HEK (Mantis) & 0.090 & 0.777 & 0.782 & 0.807 & 0.787 & 0.818 & \emph{0.110} \\
\bottomrule
\end{tabular}%
}
\end{table*}
 
\FloatBarrier

\subsection{A 2D variant of the benchmark}
\label{sec:twod-benchmark}

The 2D variant pairs each 3D model family with a 2D counterpart that keeps its objective: FNet2D, UNeXt2-2D, VSCyto2D, Pix2Pix2D, and CELL-Diff2D for FNet3D, UNeXt2, VSCyto3D, Pix2Pix3D, and CELL-Diff3D. VSCyto2D initializes its encoder from the public 2D FCMAE checkpoint~\citep{liu2025robust}. Each model is trained on the five in-focus planes of every training field of view and predicts the three-plane slab around a focus plane anchored on the ground-truth nuclei, or on the input phase for A549 ER and mitochondria, which have no nuclear channel. Pixel metrics, GLCM+, and ER and mitochondria Dice are computed on that plane, and nucleus and membrane segmentation and the learned feature extractors use the slab; test sets and the five selected metrics are those of the 3D benchmark. Tab.~\ref{tab:twod-values} reports the in-distribution values; CELL-Diff2D has the best CELL-DINO KID for every organelle and test set.

Tab.~\ref{tab:twod-rankings} compares the family rankings of the two benchmarks. Over all contexts the median Kendall $\tau$ is $+0.60$, against $+0.80$ between the three A549 test conditions within either benchmark. Where a second training seed exists (four families; nucleus and membrane, iPSC-trained), the 2D ranking agrees with itself more closely than with the 3D one (median $\tau = +1.00$ versus $+0.67$), except on si-SSIM and GLCM+ cos, where the two agree equally. Tier-level conclusions carry over: CELL-Diff ranks first on CELL-DINO KID in all 48 contexts in both benchmarks; FNet and CELL-Diff are the two lowest-ranked families on si-SSIM, Spectral PCC, and Dice in both; and FNet remains the OOD failure case, ranking last on Spectral PCC in 16 of 16 OOD contexts in 2D and 14 of 16 in 3D. The best-ranked family carries over only in part: Pix2Pix3D is the best 3D family on si-SSIM in 13 of 16 in-distribution contexts and VSCyto2D the best 2D family in 12 of 16, and Dice shifts the same way, whereas Pix2Pix ranks first on Spectral PCC in both. GLCM+ cos agrees least ($\tau = +0.30$), although its A549 conditions agree only at $+0.60$ within the 2D benchmark. The 2D variant therefore screens for coarse model quality at a fraction of the cost but does not replace the 3D benchmark for selecting the best model.

Training the five 2D families took 433 accelerator-hours, against 11{,}893 for the 3D families (final fits on mixed GPU types). The 2D recipes keep the 3D objectives but not their step counts: FNet2D is matched, Pix2Pix2D and CELL-Diff2D take fewer steps, and UNeXt2-2D and VSCyto2D take more. Pix2Pix2D warms up for the same fraction of training as Pix2Pix3D; with the 3D absolute warmup length, five of its twelve fits ended still warming up, and correcting this improved it slightly, left the median agreement unchanged (best family agreeing in 122 rather than 123 of 231 contexts), and raised the 2D seed agreement from $+0.67$ to $+1.00$. VSCyto2D, which ranks first in 2D, has no second seed. A549 ER Dice stays below 0.18 for every model in both benchmarks and is left out of the ranking statistics.

\definecolor{tierpixel}{RGB}{46,139,131}
\definecolor{tierseg}{RGB}{201,162,39}
\definecolor{tierpheno}{RGB}{74,90,115}
\begin{table}[ht]
\centering
\small
\caption{\textbf{2D benchmark: in-distribution metrics of the five 2D families.} Pixel metrics, GLCM+ and ER and mitochondria Dice are computed on one focus plane per field of view and timepoint, anchored on the ground-truth nuclei (on the input phase for A549 ER and mitochondria); nucleus and membrane segmentation and the learned feature extractors use the three-plane slab around it. iPSC test: iPSC-trained models; A549 test: A549-trained models on the mock condition. Dice denotes instance Dice for nucleus and membrane and semantic Dice for ER and mitochondria. SPCC denotes Spectral PCC; CD KID denotes CELL-DINO KID. Colored squares denote the metric tier: \textcolor{tierpixel}{\rule{1.1ex}{1.1ex}}~pixel, \textcolor{tierseg}{\rule{1.1ex}{1.1ex}}~segmentation, \textcolor{tierpheno}{\rule{1.1ex}{1.1ex}}~phenotype. Bold marks the best mean and means within its SD, per organelle, metric and test set.}
\label{tab:twod-values}
\resizebox{\textwidth}{!}{%
\begin{tabular}{l cccccccccc}
\toprule
 & \multicolumn{5}{c}{\textbf{iPSC test}} & \multicolumn{5}{c}{\textbf{A549 test}} \\
\cmidrule(lr){2-6}\cmidrule(lr){7-11}
\textbf{Metric} & \textbf{FNet2D} & \textbf{UNeXt2-2D} & \textbf{VSCyto2D} & \textbf{Pix2Pix2D} & \textbf{CELL-Diff2D} & \textbf{FNet2D} & \textbf{UNeXt2-2D} & \textbf{VSCyto2D} & \textbf{Pix2Pix2D} & \textbf{CELL-Diff2D} \\
\midrule
\multicolumn{11}{l}{\textbf{Nucleus}} \\
  \textcolor{tierpixel}{\rule{1.1ex}{1.1ex}}~si-SSIM ($\uparrow$) & $\mathbf{0.32}$ & $\mathbf{0.34}$ & $\mathbf{0.35}$ & $\mathbf{0.35}$ & $\mathbf{0.23}$ & $\mathbf{0.87}$ & $\mathbf{0.88}$ & $\mathbf{0.89}$ & $\mathbf{0.88}$ & $0.81$ \\
  \textcolor{tierpixel}{\rule{1.1ex}{1.1ex}}~SPCC ($\uparrow$) & $\mathbf{0.80}$ & $\mathbf{0.81}$ & $\mathbf{0.83}$ & $\mathbf{0.82}$ & $0.66$ & $\mathbf{0.89}$ & $\mathbf{0.88}$ & $\mathbf{0.89}$ & $\mathbf{0.90}$ & $0.83$ \\
  \textcolor{tierseg}{\rule{1.1ex}{1.1ex}}~Dice ($\uparrow$) & $\mathbf{0.75}$ & $\mathbf{0.82}$ & $\mathbf{0.84}$ & $\mathbf{0.83}$ & $0.74$ & $\mathbf{0.84}$ & $\mathbf{0.89}$ & $\mathbf{0.90}$ & $\mathbf{0.89}$ & $0.60$ \\
  \textcolor{tierpheno}{\rule{1.1ex}{1.1ex}}~CD KID ($\downarrow$) & $27.32$ & $27.95$ & $27.14$ & $28.33$ & $\mathbf{3.69}$ & $22.19$ & $23.90$ & $25.28$ & $24.37$ & $\mathbf{4.87}$ \\
  \textcolor{tierpheno}{\rule{1.1ex}{1.1ex}}~GLCM+ cos ($\uparrow$) & $0.04$ & $0.03$ & $0.02$ & $0.04$ & $\mathbf{0.31}$ & $0.51$ & $0.51$ & $0.52$ & $0.51$ & $\mathbf{0.68}$ \\
\midrule
\multicolumn{11}{l}{\textbf{Membrane}} \\
  \textcolor{tierpixel}{\rule{1.1ex}{1.1ex}}~si-SSIM ($\uparrow$) & $\mathbf{0.42}$ & $\mathbf{0.32}$ & $\mathbf{0.34}$ & $\mathbf{0.45}$ & $0.30$ & $\mathbf{0.45}$ & $\mathbf{0.48}$ & $\mathbf{0.51}$ & $\mathbf{0.48}$ & $0.30$ \\
  \textcolor{tierpixel}{\rule{1.1ex}{1.1ex}}~SPCC ($\uparrow$) & $\mathbf{0.69}$ & $0.49$ & $0.54$ & $\mathbf{0.70}$ & $0.37$ & $\mathbf{0.73}$ & $\mathbf{0.76}$ & $\mathbf{0.79}$ & $\mathbf{0.78}$ & $0.46$ \\
  \textcolor{tierseg}{\rule{1.1ex}{1.1ex}}~Dice ($\uparrow$) & $\mathbf{0.61}$ & $0.26$ & $0.38$ & $\mathbf{0.60}$ & $0.23$ & $\mathbf{0.62}$ & $\mathbf{0.68}$ & $\mathbf{0.69}$ & $\mathbf{0.65}$ & $0.38$ \\
  \textcolor{tierpheno}{\rule{1.1ex}{1.1ex}}~CD KID ($\downarrow$) & $38.40$ & $34.35$ & $37.49$ & $40.70$ & $\mathbf{8.95}$ & $24.39$ & $26.11$ & $26.98$ & $18.76$ & $\mathbf{4.93}$ \\
  \textcolor{tierpheno}{\rule{1.1ex}{1.1ex}}~GLCM+ cos ($\uparrow$) & $0.01$ & $-0.00$ & $-0.00$ & $0.04$ & $\mathbf{0.13}$ & $0.42$ & $0.45$ & $\mathbf{0.49}$ & $0.43$ & $0.24$ \\
\midrule
\multicolumn{11}{l}{\textbf{ER}} \\
  \textcolor{tierpixel}{\rule{1.1ex}{1.1ex}}~si-SSIM ($\uparrow$) & $\mathbf{0.30}$ & $\mathbf{0.29}$ & $\mathbf{0.31}$ & $\mathbf{0.32}$ & $0.21$ & $\mathbf{0.34}$ & $\mathbf{0.36}$ & $\mathbf{0.39}$ & $\mathbf{0.37}$ & $0.20$ \\
  \textcolor{tierpixel}{\rule{1.1ex}{1.1ex}}~SPCC ($\uparrow$) & $\mathbf{0.87}$ & $\mathbf{0.85}$ & $\mathbf{0.87}$ & $\mathbf{0.87}$ & $0.72$ & $\mathbf{0.69}$ & $\mathbf{0.72}$ & $\mathbf{0.76}$ & $\mathbf{0.74}$ & $0.44$ \\
  \textcolor{tierseg}{\rule{1.1ex}{1.1ex}}~Dice ($\uparrow$) & $\mathbf{0.28}$ & $\mathbf{0.31}$ & $\mathbf{0.31}$ & $\mathbf{0.30}$ & $0.26$ & $0.06$ & $0.09$ & $\mathbf{0.11}$ & $0.08$ & $\mathbf{0.14}$ \\
  \textcolor{tierpheno}{\rule{1.1ex}{1.1ex}}~CD KID ($\downarrow$) & $46.07$ & $46.85$ & $46.58$ & $46.68$ & $\mathbf{6.89}$ & $30.39$ & $26.84$ & $26.36$ & $24.44$ & $\mathbf{4.55}$ \\
  \textcolor{tierpheno}{\rule{1.1ex}{1.1ex}}~GLCM+ cos ($\uparrow$) & $0.18$ & $0.18$ & $0.19$ & $0.21$ & $\mathbf{0.32}$ & $0.19$ & $0.22$ & $\mathbf{0.23}$ & $0.20$ & $0.20$ \\
\midrule
\multicolumn{11}{l}{\textbf{Mitochondria}} \\
  \textcolor{tierpixel}{\rule{1.1ex}{1.1ex}}~si-SSIM ($\uparrow$) & $\mathbf{0.45}$ & $\mathbf{0.46}$ & $\mathbf{0.46}$ & $\mathbf{0.48}$ & $0.34$ & $\mathbf{0.60}$ & $\mathbf{0.60}$ & $\mathbf{0.65}$ & $\mathbf{0.65}$ & $0.45$ \\
  \textcolor{tierpixel}{\rule{1.1ex}{1.1ex}}~SPCC ($\uparrow$) & $\mathbf{0.82}$ & $\mathbf{0.80}$ & $\mathbf{0.82}$ & $\mathbf{0.83}$ & $0.65$ & $\mathbf{0.84}$ & $0.80$ & $\mathbf{0.86}$ & $\mathbf{0.85}$ & $0.61$ \\
  \textcolor{tierseg}{\rule{1.1ex}{1.1ex}}~Dice ($\uparrow$) & $\mathbf{0.39}$ & $\mathbf{0.38}$ & $\mathbf{0.39}$ & $\mathbf{0.37}$ & $0.27$ & $\mathbf{0.35}$ & $0.29$ & $\mathbf{0.36}$ & $\mathbf{0.34}$ & $0.24$ \\
  \textcolor{tierpheno}{\rule{1.1ex}{1.1ex}}~CD KID ($\downarrow$) & $36.59$ & $38.91$ & $37.60$ & $39.13$ & $\mathbf{9.31}$ & $29.15$ & $24.00$ & $24.59$ & $26.26$ & $\mathbf{6.58}$ \\
  \textcolor{tierpheno}{\rule{1.1ex}{1.1ex}}~GLCM+ cos ($\uparrow$) & $0.19$ & $0.16$ & $0.20$ & $0.16$ & $\mathbf{0.35}$ & $0.12$ & $0.13$ & $0.18$ & $\mathbf{0.20}$ & $\mathbf{0.20}$ \\
\bottomrule
\end{tabular}}
\end{table}
\definecolor{tierpixel}{RGB}{46,139,131}
\definecolor{tierseg}{RGB}{201,162,39}
\definecolor{tierpheno}{RGB}{74,90,115}
\begin{table}[ht]
\centering
\small
\caption{\textbf{Family rankings, 2D benchmark versus 3D benchmark.} Median Kendall $\tau$ between the five families' ranks in the two benchmarks, over (organelle, training set, test set) contexts, all and by domain; \emph{same best}: contexts whose best family agrees. Ceilings, both inside the 2D benchmark: the three A549 test conditions ranked against each other, and a second training seed (four seeded families, nucleus and membrane, iPSC-trained), shown next to the 2D-versus-3D $\tau$ on the same families and contexts. A549 ER Dice, below 0.18 for every model, is not ranked. SPCC denotes Spectral PCC; CD KID denotes CELL-DINO KID. Colored squares denote the metric tier: \textcolor{tierpixel}{\rule{1.1ex}{1.1ex}}~pixel, \textcolor{tierseg}{\rule{1.1ex}{1.1ex}}~segmentation, \textcolor{tierpheno}{\rule{1.1ex}{1.1ex}}~phenotype.}
\label{tab:twod-rankings}
\resizebox{\textwidth}{!}{%
\begin{tabular}{l cccc c c cc}
\toprule
 & \multicolumn{4}{c}{\textbf{2D vs 3D, median $\tau$}} & & \textbf{A549 conditions} & \multicolumn{2}{c}{\textbf{Seed contexts, median $\tau$}} \\
\cmidrule(lr){2-5}\cmidrule(lr){7-7}\cmidrule(lr){8-9}
\textbf{Metric} & \textbf{All} & \textbf{In-distribution} & \textbf{OOD} & \textbf{Joint} & \textbf{Same best} & \textbf{2D vs 2D} & \textbf{2D seed vs seed} & \textbf{2D vs 3D} \\
\midrule
  \textcolor{tierpixel}{\rule{1.1ex}{1.1ex}}~si-SSIM & $+0.80$ & $+0.70$ & $+0.60$ & $+0.80$ & $23/48$ & $+0.80$ & $+0.83$ & $+0.83$ \\
  \textcolor{tierpixel}{\rule{1.1ex}{1.1ex}}~SPCC & $+0.60$ & $+0.70$ & $+0.50$ & $+0.60$ & $14/48$ & $+0.80$ & $+1.00$ & $+0.67$ \\
  \textcolor{tierseg}{\rule{1.1ex}{1.1ex}}~Dice & $+0.60$ & $+0.80$ & $+0.60$ & $+0.60$ & $12/39$ & $+1.00$ & $+0.83$ & $+0.67$ \\
  \textcolor{tierpheno}{\rule{1.1ex}{1.1ex}}~CD KID & $+0.60$ & $+0.80$ & $+0.60$ & $+0.50$ & $48/48$ & $+1.00$ & $+1.00$ & $+0.67$ \\
  \textcolor{tierpheno}{\rule{1.1ex}{1.1ex}}~GLCM+ cos & $+0.30$ & $+0.20$ & $+0.50$ & $+0.40$ & $25/48$ & $+0.60$ & $+0.67$ & $+0.67$ \\
\midrule
  All five & $+0.60$ & $+0.60$ & $+0.60$ & $+0.60$ & $122/231$ & $+0.80$ & $+1.00$ & $+0.67$ \\
\bottomrule
\end{tabular}}
\end{table}
 
\FloatBarrier

\subsection{A lite test split}
\label{sec:lite-split}

For faster model iteration, the released code also defines a lite split of the 3D test sets: all 12 A549 FOVs per organelle and condition at 5 of the 10 timepoints, spread over the time course, and a fixed random 50 of the 100 iPSC FOVs per organelle. It holds half of the test volumes and roughly halves single-pass prediction on A549 ($0.44\times$ the accelerator-hours). Subsetting the full benchmark's per-FOV scores, the lite split keeps at least 89.5\% of the pairs of trained models that the full benchmark resolves (paired FOV bootstrap, difference above twice its SD) on si-SSIM, Spectral PCC, Dice, and CELL-DINO KID, and resolves none of them in the opposite order. On GLCM+ cos it keeps at least 71\%.

\subsection{Per-organelle prediction overview across training sets}
\label{sec:pred_overview_appendix}

\begin{figure}[!ht]
    \centering
    \includegraphics[width=\linewidth]{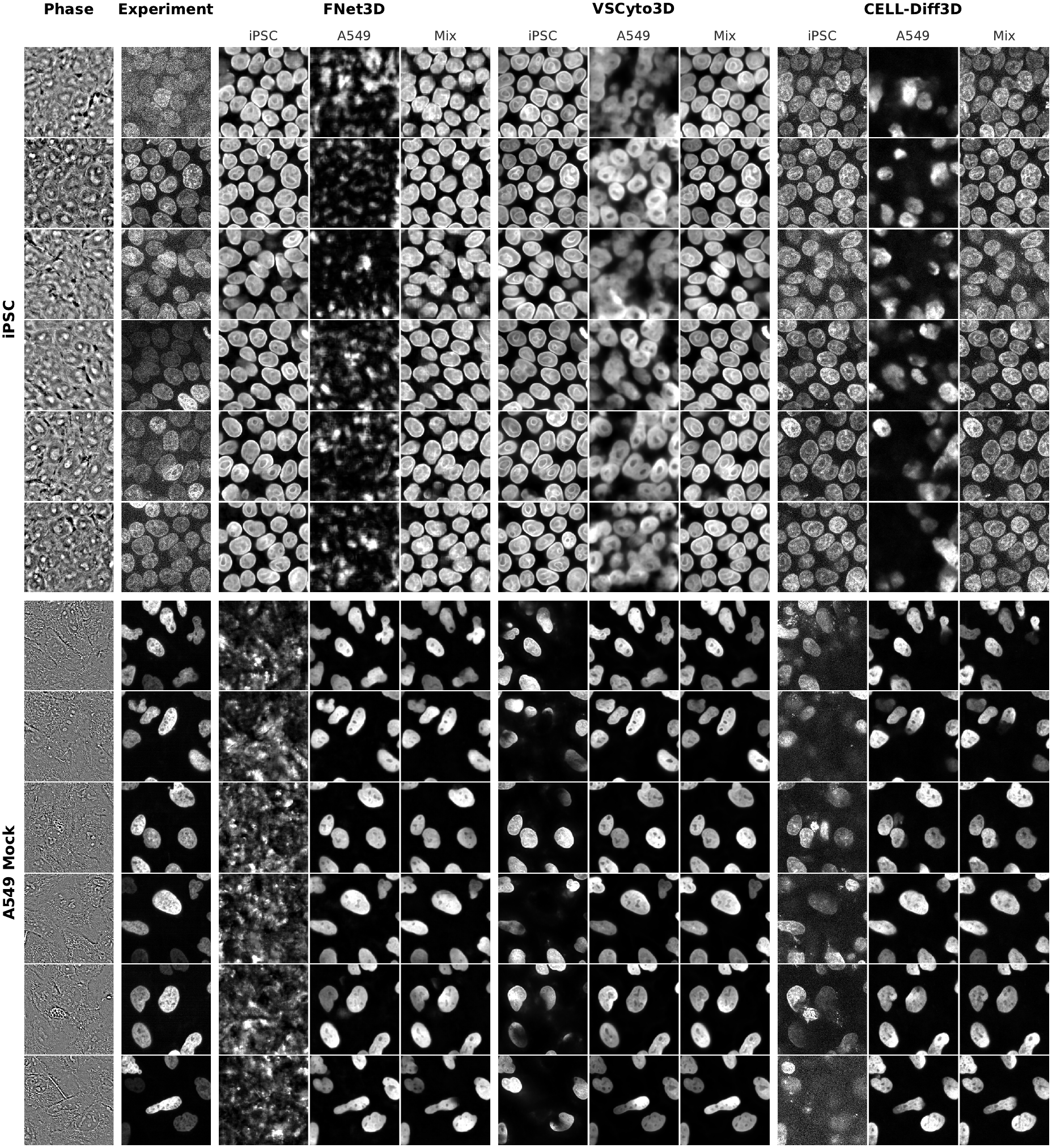}
    \caption{\textbf{Nucleus predictions across training sets.} Twelve FOVs are shown (top six: iPSC; bottom six: A549 Mock). Columns show the phase input, experimental target, and predictions from FNet3D, VSCyto3D, and CELL-Diff3D models trained on iPSC, A549, or both datasets. The maximum-nuclear-area plane is shown.}
    \label{fig:appendix-pred-overview-nucleus}
\end{figure}

\begin{figure}[!ht]
    \centering
    \includegraphics[width=\linewidth]{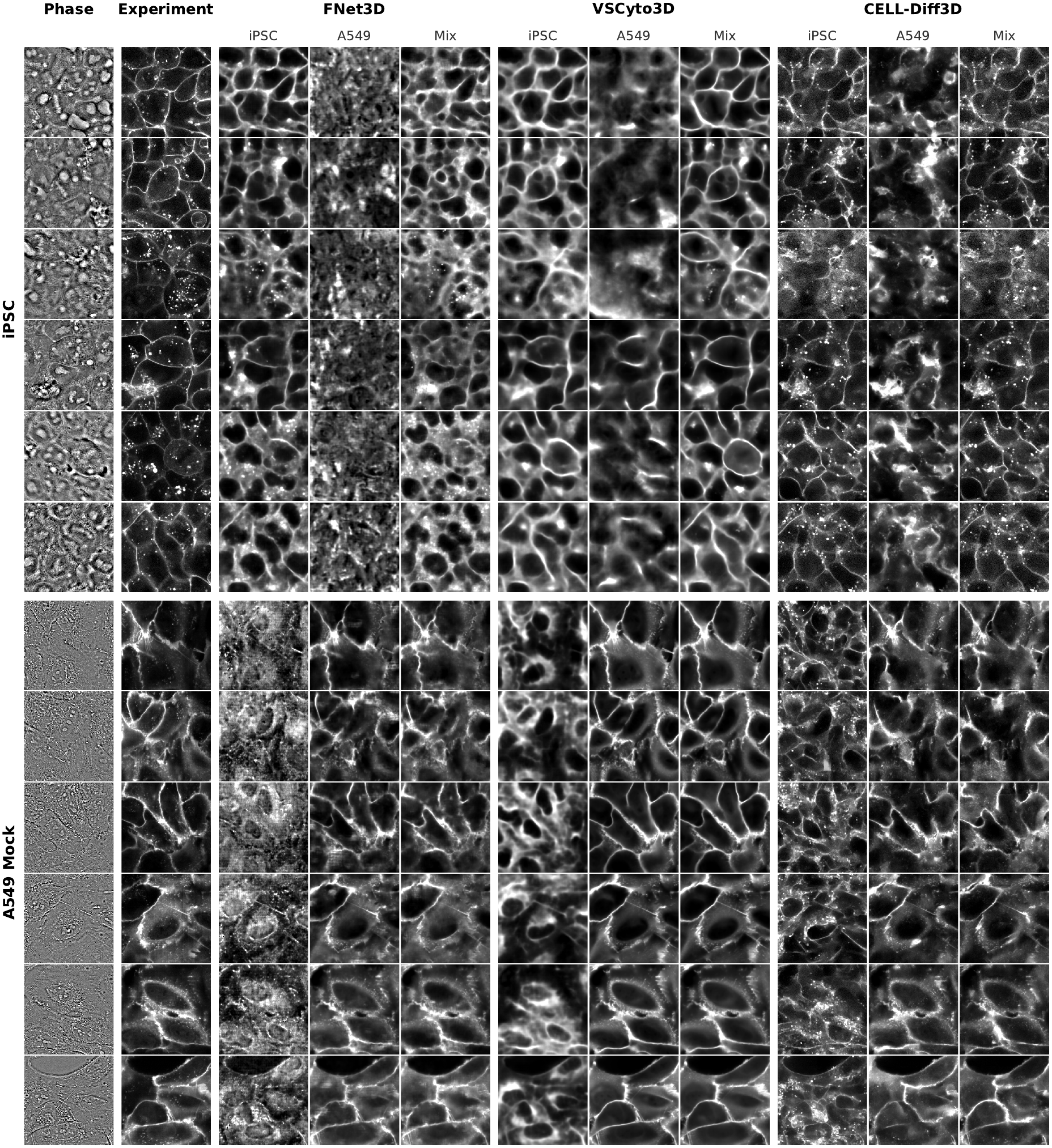}
    \caption{\textbf{Cell membrane predictions across training sets.} Twelve FOVs are shown (top six: iPSC; bottom six: A549 Mock). Columns show the phase input, experimental target, and predictions from FNet3D, VSCyto3D, and CELL-Diff3D models trained on iPSC, A549, or both datasets. The maximum-nuclear-area plane is shown.}
    \label{fig:appendix-pred-overview-membrane}
\end{figure}

\begin{figure}[!ht]
    \centering
    \includegraphics[width=\linewidth]{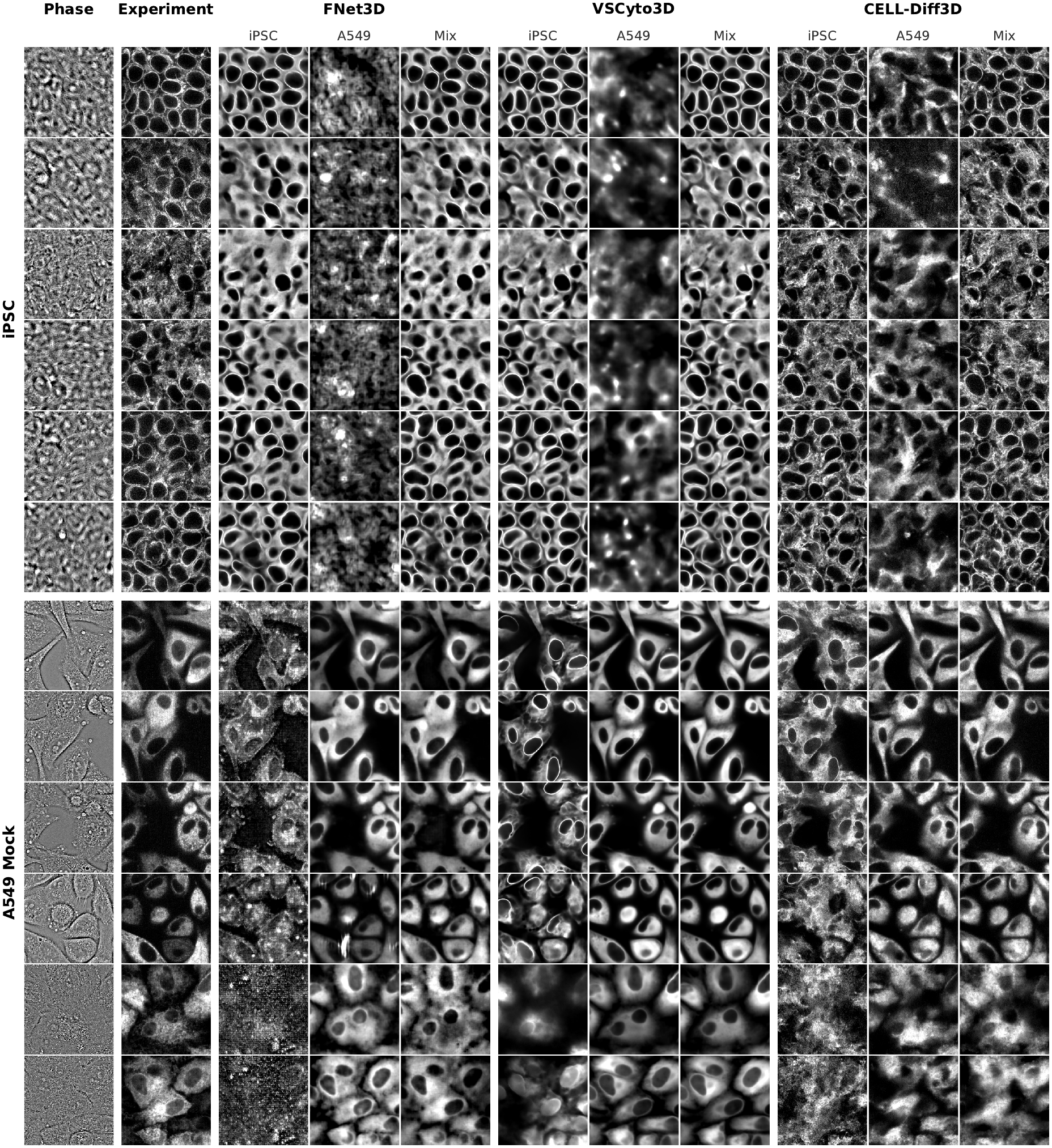}
    \caption{\textbf{ER (SEC61B) predictions across training sets.} Twelve FOVs are shown (top six: iPSC; bottom six: A549 Mock). Columns show the phase input, experimental target, and predictions from FNet3D, VSCyto3D, and CELL-Diff3D models trained on iPSC, A549, or both datasets. The phase-midband plane is shown.}
    \label{fig:appendix-pred-overview-er}
\end{figure}

\begin{figure}[!ht]
    \centering
    \includegraphics[width=\linewidth]{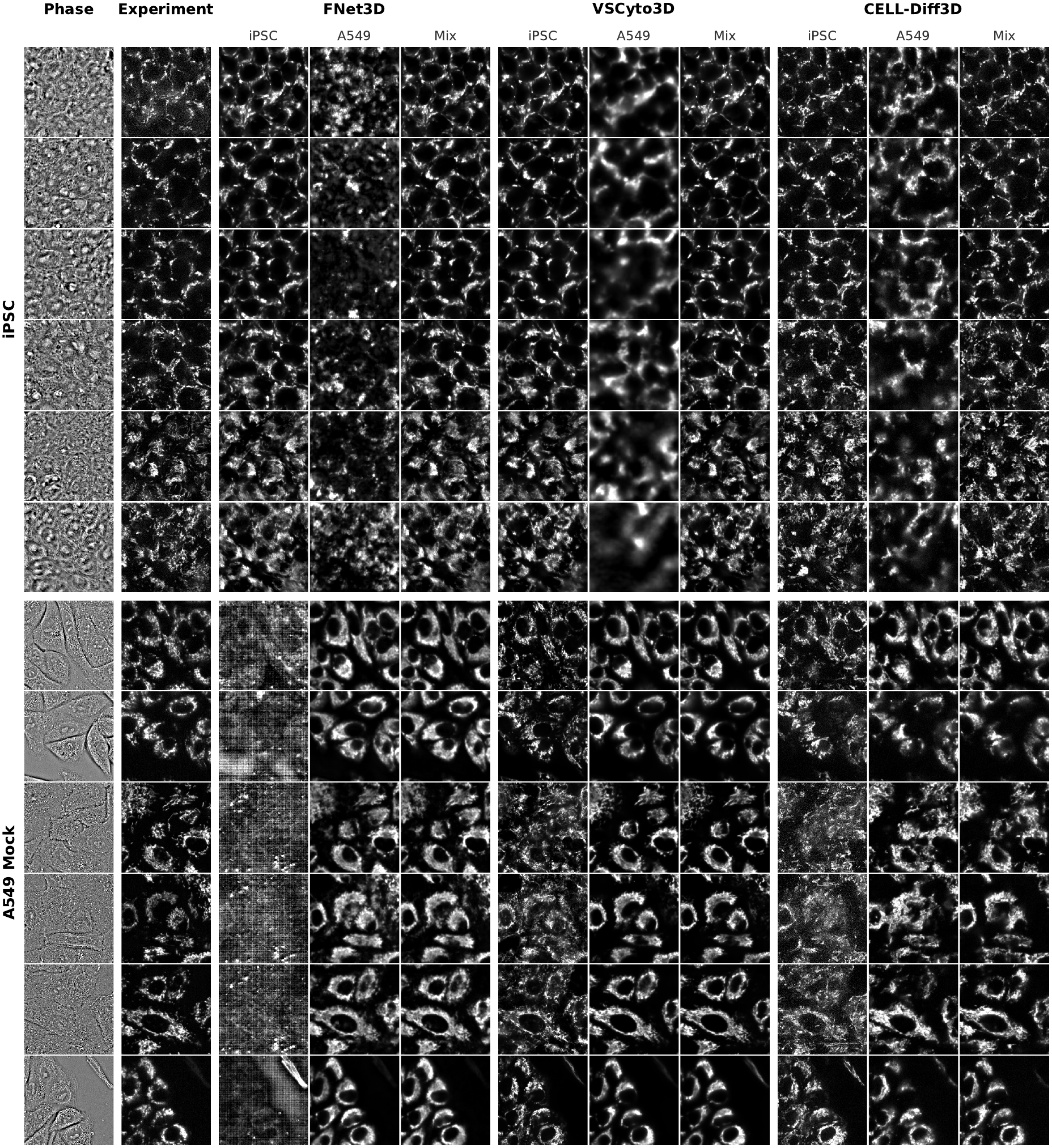}
    \caption{\textbf{Mitochondria (TOMM20) predictions across training sets.} Twelve FOVs are shown (top six: iPSC; bottom six: A549 Mock). Columns show the phase input, experimental target, and predictions from FNet3D, VSCyto3D, and CELL-Diff3D models trained on iPSC, A549, or both datasets. The phase-midband plane is shown.}
    \label{fig:appendix-pred-overview-mito}
\end{figure}

Figs.~\ref{fig:appendix-pred-overview-nucleus}--\ref{fig:appendix-pred-overview-mito} compare virtual-staining predictions for each organelle. Each figure shows twelve FOVs (top six rows: iPSC; bottom six rows: A549 Mock). Columns show the phase input, experimental fluorescence target, and predictions from FNet3D, VSCyto3D, and CELL-Diff3D models trained on iPSC, A549, or both datasets. A single in-focus plane is shown: the maximum-nuclear-area plane for nucleus and cell membrane, and the phase-midband plane for ER and mitochondria.

\subsection{Phenotype Precision and Recall across organelles}
\label{sec:precision_recall}

KID can remain low when paired spatial or instance-level scores are weak, so we also use a phenotype metric that separates fidelity from diversity. Following~\citet{kynkaanniemi2019improved}, we compute Precision ($P$) and Recall ($R$) on DynaCLR single-cell embeddings. Precision is the fraction of predicted cells inside the experimental $k$-NN manifold, and Recall is the fraction of experimental cells covered by the predicted $k$-NN manifold, with $k=5$. We also report $F_1 = 2PR/(P+R)$. $F_1$ penalizes models with high Precision but low Recall, or vice versa.

Tabs.~\ref{tab:precision-recall-membrane-updated}--\ref{tab:precision-recall-mitochondria-updated} report per-organelle $P/R/F_1$ using the DynaCLR embeddings. They include the four main models and the UNeXt2 and UNetViT3D regression ablations, with all three training sets where available. We report only the model--training-set combinations that were evaluated.

For jointly trained membrane models, CELL-Diff3D reaches $F_1=0.87$, $0.90$, and $0.88$ on A549 Mock, DENV, and ZIKV, respectively, with both high fidelity ($P=0.95$--$0.98$) and diversity ($R=0.78$--$0.83$). The regression and adversarial models reach only $F_1=0.06$--$0.31$ under the same conditions, with either fidelity or diversity low. Fewer combinations are available for the other organelles, and the results vary by organelle. We therefore report $P$ and $R$ separately rather than reducing phenotype agreement to one score.

\begin{table}[ht]
\centering
\small
\caption{\textbf{Precision / Recall / $F_1$ on DynaCLR single-cell embeddings: Cell Membrane.} We report $P$ (fidelity, fraction of predicted cells inside the experimental $k$-NN manifold), $R$ (diversity, fraction of experimental cells covered by the predicted $k$-NN manifold), and $F_1=2PR/(P+R)$ on per-cell DynaCLR embeddings ($k=5$). Results are computed directly from the saved single-cell embeddings.}
\label{tab:precision-recall-membrane-updated}
\resizebox{\textwidth}{!}{%
\begin{tabular}{l l cccccccccccccccc}
\toprule
\multirow{2}{*}{\textbf{Cell type}} & \multirow{2}{*}{\textbf{Metric}} & \multicolumn{3}{c}{\textbf{FNet3D}} & \multicolumn{3}{c}{\textbf{UNeXt2}} & \multicolumn{3}{c}{\textbf{VSCyto3D}} & \multicolumn{1}{c}{\textbf{UNetViT3D}} & \multicolumn{3}{c}{\textbf{Pix2Pix3D}} & \multicolumn{3}{c}{\textbf{CELL-Diff3D}} \\
\cmidrule(lr){3-5} \cmidrule(lr){6-8} \cmidrule(lr){9-11} \cmidrule(lr){12-12} \cmidrule(lr){13-15} \cmidrule(lr){16-18}
&  & \textbf{iPSC} & \textbf{A549} & \textbf{Joint} & \textbf{iPSC} & \textbf{A549} & \textbf{Joint} & \textbf{iPSC} & \textbf{A549} & \textbf{Joint} & \textbf{iPSC} & \textbf{iPSC} & \textbf{A549} & \textbf{Joint} & \textbf{iPSC} & \textbf{A549} & \textbf{Joint} \\
\midrule
\multirow{3}{*}{iPSC} & $P$ $\uparrow$ & $0.26$ & $0.20$ & $0.21$ & $0.27$ & $0.37$ & $0.30$ & $0.30$ & $0.38$ & $0.37$ & $0.26$ & $0.40$ & $0.29$ & $0.53$ & $0.85$ & $0.48$ & $0.90$ \\
 & $R$ $\uparrow$ & $0.02$ & $0.05$ & $0.01$ & $0.19$ & $0.13$ & $0.20$ & $0.30$ & $0.11$ & $0.27$ & $0.03$ & $0.01$ & $0.04$ & $0.02$ & $0.69$ & $0.34$ & $0.75$ \\
 & $F_1$ $\uparrow$ & $0.03$ & $0.09$ & $0.02$ & $0.22$ & $0.19$ & $0.24$ & $0.30$ & $0.17$ & $0.31$ & $0.06$ & $0.02$ & $0.07$ & $0.03$ & $0.76$ & $0.40$ & $0.82$ \\
\midrule
\multirow{3}{*}{A549 Mock} & $P$ $\uparrow$ & $0.33$ & $0.19$ & $0.22$ & $0.12$ & $0.06$ & $0.07$ & $0.04$ & $0.05$ & $0.04$ & $0.38$ & $0.14$ & $0.11$ & $0.12$ & $0.77$ & $0.97$ & $0.98$ \\
 & $R$ $\uparrow$ & $0.14$ & $0.15$ & $0.25$ & $0.16$ & $0.12$ & $0.17$ & $0.10$ & $0.15$ & $0.11$ & $0.22$ & $0.16$ & $0.30$ & $0.29$ & $0.35$ & $0.82$ & $0.78$ \\
 & $F_1$ $\uparrow$ & $0.20$ & $0.17$ & $0.24$ & $0.14$ & $0.08$ & $0.10$ & $0.05$ & $0.07$ & $0.06$ & $0.28$ & $0.15$ & $0.16$ & $0.17$ & $0.48$ & $0.89$ & $0.87$ \\
\midrule
\multirow{3}{*}{A549 DENV} & $P$ $\uparrow$ & $0.32$ & $0.53$ & $0.49$ & $0.33$ & $0.21$ & $0.26$ & $0.24$ & $0.21$ & $0.19$ & $0.64$ & $0.67$ & $0.32$ & $0.35$ & $0.65$ & $0.98$ & $0.97$ \\
 & $R$ $\uparrow$ & $0.14$ & $0.17$ & $0.23$ & $0.21$ & $0.11$ & $0.10$ & $0.17$ & $0.11$ & $0.09$ & $0.56$ & $0.41$ & $0.12$ & $0.13$ & $0.30$ & $0.82$ & $0.83$ \\
 & $F_1$ $\uparrow$ & $0.20$ & $0.26$ & $0.31$ & $0.26$ & $0.14$ & $0.15$ & $0.20$ & $0.15$ & $0.13$ & $0.60$ & $0.51$ & $0.18$ & $0.19$ & $0.41$ & $0.89$ & $0.90$ \\
\midrule
\multirow{3}{*}{A549 ZIKV} & $P$ $\uparrow$ & $0.35$ & $0.22$ & $0.26$ & $0.23$ & $0.15$ & $0.19$ & $0.11$ & $0.16$ & $0.15$ & $0.39$ & $0.19$ & $0.22$ & $0.22$ & $0.68$ & $0.96$ & $0.95$ \\
 & $R$ $\uparrow$ & $0.15$ & $0.16$ & $0.39$ & $0.09$ & $0.07$ & $0.61$ & $0.44$ & $0.07$ & $0.08$ & $0.29$ & $0.21$ & $0.38$ & $0.49$ & $0.33$ & $0.83$ & $0.82$ \\
 & $F_1$ $\uparrow$ & $0.21$ & $0.19$ & $0.31$ & $0.13$ & $0.10$ & $0.29$ & $0.17$ & $0.10$ & $0.11$ & $0.33$ & $0.20$ & $0.27$ & $0.31$ & $0.44$ & $0.89$ & $0.88$ \\
\bottomrule
\end{tabular}%
}
\end{table}

\begin{table}[ht]
\centering
\small
\caption{\textbf{Precision / Recall / $F_1$ on DynaCLR single-cell embeddings: Nucleus.} We report $P$ (fidelity, fraction of predicted cells inside the experimental $k$-NN manifold), $R$ (diversity, fraction of experimental cells covered by the predicted $k$-NN manifold), and $F_1=2PR/(P+R)$ on per-cell DynaCLR embeddings ($k=5$). Results are computed directly from the saved single-cell embeddings.}
\label{tab:precision-recall-nucleus-updated}
\resizebox{\textwidth}{!}{%
\begin{tabular}{l l cccccccccccccccc}
\toprule
\multirow{2}{*}{\textbf{Cell type}} & \multirow{2}{*}{\textbf{Metric}} & \multicolumn{3}{c}{\textbf{FNet3D}} & \multicolumn{3}{c}{\textbf{UNeXt2}} & \multicolumn{3}{c}{\textbf{VSCyto3D}} & \multicolumn{1}{c}{\textbf{UNetViT3D}} & \multicolumn{3}{c}{\textbf{Pix2Pix3D}} & \multicolumn{3}{c}{\textbf{CELL-Diff3D}} \\
\cmidrule(lr){3-5} \cmidrule(lr){6-8} \cmidrule(lr){9-11} \cmidrule(lr){12-12} \cmidrule(lr){13-15} \cmidrule(lr){16-18}
&  & \textbf{iPSC} & \textbf{A549} & \textbf{Joint} & \textbf{iPSC} & \textbf{A549} & \textbf{Joint} & \textbf{iPSC} & \textbf{A549} & \textbf{Joint} & \textbf{iPSC} & \textbf{iPSC} & \textbf{A549} & \textbf{Joint} & \textbf{iPSC} & \textbf{A549} & \textbf{Joint} \\
\midrule
\multirow{3}{*}{iPSC} & $P$ $\uparrow$ & $0.00$ & $0.00$ & $0.00$ & $0.00$ & $0.00$ & $0.00$ & $0.00$ & $0.00$ & $0.00$ & $0.00$ & $0.00$ & $0.01$ & $0.01$ & $0.63$ & $0.18$ & $0.63$ \\
 & $R$ $\uparrow$ & $0.01$ & $0.00$ & $0.00$ & $0.45$ & $0.02$ & $0.09$ & $0.03$ & $0.01$ & $0.24$ & $0.01$ & $0.02$ & $0.00$ & $0.04$ & $0.82$ & $0.08$ & $0.83$ \\
 & $F_1$ $\uparrow$ & $0.00$ & $0.00$ & $0.00$ & $0.00$ & $0.00$ & $0.00$ & $0.00$ & $0.00$ & $0.00$ & $0.00$ & $0.00$ & $0.00$ & $0.02$ & $0.71$ & $0.11$ & $0.71$ \\
\midrule
\multirow{3}{*}{A549 Mock} & $P$ $\uparrow$ & $0.05$ & $0.24$ & $0.30$ & $0.10$ & $0.32$ & $0.30$ & $0.09$ & $0.32$ & $0.32$ & $0.09$ & $0.12$ & $0.38$ & $0.49$ & $0.49$ & $0.89$ & $0.93$ \\
 & $R$ $\uparrow$ & $0.06$ & $0.02$ & $0.11$ & $0.22$ & $0.01$ & $0.02$ & $0.24$ & $0.01$ & $0.01$ & $0.17$ & $0.10$ & $0.46$ & $0.18$ & $0.15$ & $0.77$ & $0.83$ \\
 & $F_1$ $\uparrow$ & $0.05$ & $0.04$ & $0.16$ & $0.14$ & $0.03$ & $0.03$ & $0.13$ & $0.02$ & $0.02$ & $0.12$ & $0.11$ & $0.41$ & $0.26$ & $0.23$ & $0.83$ & $0.88$ \\
\midrule
\multirow{3}{*}{A549 DENV} & $P$ $\uparrow$ & $0.03$ & $0.20$ & $0.23$ & $0.09$ & $0.20$ & $0.18$ & $0.09$ & $0.19$ & $0.21$ & $0.05$ & $0.08$ & $0.26$ & $0.40$ & $0.30$ & $0.90$ & $0.90$ \\
 & $R$ $\uparrow$ & $0.12$ & $0.04$ & $0.06$ & $0.21$ & $0.02$ & $0.02$ & $0.10$ & $0.02$ & $0.02$ & $0.13$ & $0.08$ & $0.19$ & $0.13$ & $0.40$ & $0.76$ & $0.84$ \\
 & $F_1$ $\uparrow$ & $0.05$ & $0.06$ & $0.10$ & $0.13$ & $0.03$ & $0.03$ & $0.10$ & $0.04$ & $0.04$ & $0.07$ & $0.08$ & $0.22$ & $0.20$ & $0.34$ & $0.82$ & $0.87$ \\
\midrule
\multirow{3}{*}{A549 ZIKV} & $P$ $\uparrow$ & $0.06$ & $0.19$ & $0.25$ & $0.12$ & $0.31$ & $0.27$ & $0.10$ & $0.28$ & $0.29$ & $0.09$ & $0.16$ & $0.31$ & $0.41$ & $0.32$ & $0.89$ & $0.91$ \\
 & $R$ $\uparrow$ & $0.03$ & $0.05$ & $0.06$ & $0.10$ & $0.48$ & $0.16$ & $0.06$ & $0.03$ & $0.28$ & $0.11$ & $0.08$ & $0.10$ & $0.15$ & $0.20$ & $0.80$ & $0.83$ \\
 & $F_1$ $\uparrow$ & $0.04$ & $0.08$ & $0.10$ & $0.11$ & $0.38$ & $0.20$ & $0.08$ & $0.06$ & $0.28$ & $0.10$ & $0.11$ & $0.15$ & $0.22$ & $0.25$ & $0.85$ & $0.87$ \\
\bottomrule
\end{tabular}%
}
\end{table}

\begin{table}[ht]
\centering
\small
\caption{\textbf{Precision / Recall / $F_1$ on DynaCLR single-cell embeddings: ER (SEC61B).} We report $P$ (fidelity, fraction of predicted cells inside the experimental $k$-NN manifold), $R$ (diversity, fraction of experimental cells covered by the predicted $k$-NN manifold), and $F_1=2PR/(P+R)$ on per-cell DynaCLR embeddings ($k=5$). Results are computed directly from the saved single-cell embeddings.}
\label{tab:precision-recall-er-updated}
\resizebox{\textwidth}{!}{%
\begin{tabular}{l l cccccccccccccccccc}
\toprule
\multirow{2}{*}{\textbf{Cell type}} & \multirow{2}{*}{\textbf{Metric}} & \multicolumn{3}{c}{\textbf{FNet3D}} & \multicolumn{3}{c}{\textbf{UNeXt2}} & \multicolumn{3}{c}{\textbf{VSCyto3D}} & \multicolumn{3}{c}{\textbf{UNetViT3D}} & \multicolumn{3}{c}{\textbf{Pix2Pix3D}} & \multicolumn{3}{c}{\textbf{CELL-Diff3D}} \\
\cmidrule(lr){3-5} \cmidrule(lr){6-8} \cmidrule(lr){9-11} \cmidrule(lr){12-14} \cmidrule(lr){15-17} \cmidrule(lr){18-20}
&  & \textbf{iPSC} & \textbf{A549} & \textbf{Joint} & \textbf{iPSC} & \textbf{A549} & \textbf{Joint} & \textbf{iPSC} & \textbf{A549} & \textbf{Joint} & \textbf{iPSC} & \textbf{A549} & \textbf{Joint} & \textbf{iPSC} & \textbf{A549} & \textbf{Joint} & \textbf{iPSC} & \textbf{A549} & \textbf{Joint} \\
\midrule
\multirow{3}{*}{iPSC} & $P$ $\uparrow$ & $0.00$ & $0.00$ & $0.00$ & $0.00$ & $0.00$ & $0.00$ & $0.00$ & $0.00$ & $0.00$ & $0.00$ & $0.01$ & $0.00$ & $0.00$ & $0.01$ & $0.00$ & $0.86$ & $0.20$ & $0.88$ \\
 & $R$ $\uparrow$ & $0.00$ & $0.02$ & $0.00$ & $0.01$ & $0.08$ & $0.00$ & $0.00$ & $0.01$ & $0.00$ & $0.00$ & $0.03$ & $0.00$ & $0.00$ & $0.10$ & $0.00$ & $0.83$ & $0.11$ & $0.86$ \\
 & $F_1$ $\uparrow$ & $0.00$ & $0.01$ & \textemdash & $0.00$ & $0.00$ & \textemdash & $0.00$ & $0.00$ & $0.00$ & $0.00$ & $0.01$ & $0.00$ & $0.00$ & $0.01$ & $0.00$ & $0.85$ & $0.14$ & $0.87$ \\
\midrule
\multirow{3}{*}{A549 Mock} & $P$ $\uparrow$ & $0.09$ & $0.01$ & $0.01$ & $0.01$ & $0.00$ & $0.00$ & $0.02$ & $0.00$ & $0.00$ & $0.11$ & $0.02$ & $0.01$ & $0.02$ & $0.00$ & $0.01$ & $0.60$ & $0.75$ & $0.73$ \\
 & $R$ $\uparrow$ & $0.04$ & $0.01$ & $0.08$ & $0.03$ & $0.01$ & $0.00$ & $0.34$ & $0.00$ & $0.00$ & $0.28$ & $0.14$ & $0.06$ & $0.19$ & $0.09$ & $0.03$ & $0.30$ & $0.61$ & $0.59$ \\
 & $F_1$ $\uparrow$ & $0.06$ & $0.01$ & $0.02$ & $0.02$ & $0.00$ & $0.00$ & $0.04$ & $0.00$ & $0.00$ & $0.16$ & $0.03$ & $0.02$ & $0.04$ & $0.01$ & $0.02$ & $0.40$ & $0.67$ & $0.65$ \\
\midrule
\multirow{3}{*}{A549 DENV} & $P$ $\uparrow$ & $0.15$ & $0.01$ & $0.01$ & $0.01$ & $0.00$ & $0.00$ & $0.01$ & $0.00$ & $0.00$ & $0.05$ & $0.02$ & $0.01$ & $0.02$ & $0.00$ & $0.01$ & $0.57$ & $0.82$ & $0.81$ \\
 & $R$ $\uparrow$ & $0.11$ & $0.01$ & $0.78$ & $0.24$ & $0.05$ & $0.03$ & $0.15$ & $0.16$ & $0.10$ & $0.04$ & $0.34$ & $0.47$ & $0.35$ & $0.69$ & $0.26$ & $0.33$ & $0.65$ & $0.60$ \\
 & $F_1$ $\uparrow$ & $0.13$ & $0.01$ & $0.02$ & $0.02$ & $0.00$ & $0.00$ & $0.01$ & $0.00$ & $0.00$ & $0.05$ & $0.04$ & $0.02$ & $0.04$ & $0.00$ & $0.01$ & $0.42$ & $0.73$ & $0.69$ \\
\midrule
\multirow{3}{*}{A549 ZIKV} & $P$ $\uparrow$ & $0.13$ & $0.14$ & $0.15$ & $0.18$ & $0.13$ & $0.13$ & $0.17$ & $0.13$ & $0.13$ & $0.20$ & $0.13$ & $0.15$ & $0.12$ & $0.11$ & $0.12$ & $0.59$ & $0.74$ & $0.65$ \\
 & $R$ $\uparrow$ & $0.16$ & $0.05$ & $0.14$ & $0.86$ & $0.01$ & $0.01$ & $0.38$ & $0.01$ & $0.00$ & $0.30$ & $0.10$ & $0.02$ & $0.22$ & $0.01$ & $0.04$ & $0.33$ & $0.68$ & $0.67$ \\
 & $F_1$ $\uparrow$ & $0.14$ & $0.07$ & $0.14$ & $0.29$ & $0.02$ & $0.02$ & $0.24$ & $0.02$ & $0.01$ & $0.24$ & $0.11$ & $0.03$ & $0.16$ & $0.01$ & $0.06$ & $0.42$ & $0.71$ & $0.66$ \\
\bottomrule
\end{tabular}%
}
\end{table}

\begin{table}[ht]
\centering
\small
\caption{\textbf{Precision / Recall / $F_1$ on DynaCLR single-cell embeddings: Mito.\ (TOMM20).} We report $P$ (fidelity, fraction of predicted cells inside the experimental $k$-NN manifold), $R$ (diversity, fraction of experimental cells covered by the predicted $k$-NN manifold), and $F_1=2PR/(P+R)$ on per-cell DynaCLR embeddings ($k=5$). Results are computed directly from the saved single-cell embeddings.}
\label{tab:precision-recall-mitochondria-updated}
\resizebox{\textwidth}{!}{%
\begin{tabular}{l l cccccccccccccccccc}
\toprule
\multirow{2}{*}{\textbf{Cell type}} & \multirow{2}{*}{\textbf{Metric}} & \multicolumn{3}{c}{\textbf{FNet3D}} & \multicolumn{3}{c}{\textbf{UNeXt2}} & \multicolumn{3}{c}{\textbf{VSCyto3D}} & \multicolumn{3}{c}{\textbf{UNetViT3D}} & \multicolumn{3}{c}{\textbf{Pix2Pix3D}} & \multicolumn{3}{c}{\textbf{CELL-Diff3D}} \\
\cmidrule(lr){3-5} \cmidrule(lr){6-8} \cmidrule(lr){9-11} \cmidrule(lr){12-14} \cmidrule(lr){15-17} \cmidrule(lr){18-20}
&  & \textbf{iPSC} & \textbf{A549} & \textbf{Joint} & \textbf{iPSC} & \textbf{A549} & \textbf{Joint} & \textbf{iPSC} & \textbf{A549} & \textbf{Joint} & \textbf{iPSC} & \textbf{A549} & \textbf{Joint} & \textbf{iPSC} & \textbf{A549} & \textbf{Joint} & \textbf{iPSC} & \textbf{A549} & \textbf{Joint} \\
\midrule
\multirow{3}{*}{iPSC} & $P$ $\uparrow$ & $0.00$ & $0.00$ & $0.00$ & $0.00$ & $0.00$ & $0.00$ & $0.00$ & $0.00$ & $0.00$ & $0.00$ & $0.00$ & $0.00$ & $0.00$ & $0.00$ & $0.00$ & $0.93$ & $0.04$ & $0.82$ \\
 & $R$ $\uparrow$ & $0.00$ & $0.15$ & $0.00$ & $0.01$ & $0.12$ & $0.00$ & $0.00$ & $0.03$ & $0.00$ & $0.00$ & $0.07$ & $0.00$ & $0.00$ & $0.26$ & $0.06$ & $0.87$ & $0.05$ & $0.84$ \\
 & $F_1$ $\uparrow$ & $0.00$ & $0.00$ & $0.00$ & $0.00$ & $0.00$ & $0.00$ & $0.00$ & $0.00$ & $0.00$ & $0.00$ & $0.00$ & $0.00$ & $0.00$ & $0.00$ & $0.00$ & $0.89$ & $0.05$ & $0.83$ \\
\midrule
\multirow{3}{*}{A549 Mock} & $P$ $\uparrow$ & $0.52$ & $0.01$ & $0.01$ & $0.08$ & $0.00$ & $0.00$ & $0.06$ & $0.00$ & $0.00$ & $0.10$ & $0.01$ & $0.01$ & $0.03$ & $0.00$ & $0.01$ & $0.44$ & $0.67$ & $0.64$ \\
 & $R$ $\uparrow$ & $0.07$ & $0.31$ & $0.69$ & $0.17$ & $0.01$ & $0.05$ & $0.02$ & $0.07$ & $0.02$ & $0.37$ & $0.33$ & $0.36$ & $0.66$ & $0.01$ & $0.35$ & $0.23$ & $0.59$ & $0.64$ \\
 & $F_1$ $\uparrow$ & $0.13$ & $0.02$ & $0.01$ & $0.11$ & $0.00$ & $0.01$ & $0.02$ & $0.00$ & $0.00$ & $0.16$ & $0.02$ & $0.02$ & $0.06$ & $0.01$ & $0.01$ & $0.30$ & $0.62$ & $0.64$ \\
\midrule
\multirow{3}{*}{A549 DENV} & $P$ $\uparrow$ & $0.29$ & $0.00$ & $0.00$ & $0.00$ & $0.00$ & $0.00$ & $0.01$ & $0.00$ & $0.00$ & $0.01$ & $0.01$ & $0.01$ & $0.01$ & $0.00$ & $0.00$ & $0.42$ & $0.57$ & $0.59$ \\
 & $R$ $\uparrow$ & $0.07$ & $0.09$ & $0.07$ & $0.00$ & $0.01$ & $0.01$ & $0.00$ & $0.03$ & $0.01$ & $0.26$ & $0.18$ & $0.24$ & $0.01$ & $0.42$ & $0.35$ & $0.23$ & $0.59$ & $0.61$ \\
 & $F_1$ $\uparrow$ & $0.11$ & $0.01$ & $0.01$ & $0.00$ & $0.00$ & $0.00$ & $0.00$ & $0.00$ & $0.00$ & $0.02$ & $0.02$ & $0.02$ & $0.01$ & $0.00$ & $0.01$ & $0.30$ & $0.58$ & $0.60$ \\
\midrule
\multirow{3}{*}{A549 ZIKV} & $P$ $\uparrow$ & $0.17$ & $0.01$ & $0.01$ & $0.05$ & $0.00$ & $0.00$ & $0.03$ & $0.00$ & $0.01$ & $0.11$ & $0.01$ & $0.01$ & $0.03$ & $0.00$ & $0.01$ & $0.47$ & $0.70$ & $0.64$ \\
 & $R$ $\uparrow$ & $0.14$ & $0.15$ & $0.08$ & $0.15$ & $0.01$ & $0.02$ & $0.15$ & $0.00$ & $0.02$ & $0.23$ & $0.06$ & $0.05$ & $0.05$ & $0.02$ & $0.06$ & $0.16$ & $0.54$ & $0.59$ \\
 & $F_1$ $\uparrow$ & $0.15$ & $0.02$ & $0.01$ & $0.08$ & $0.00$ & $0.01$ & $0.05$ & $0.00$ & $0.01$ & $0.15$ & $0.02$ & $0.02$ & $0.03$ & $0.01$ & $0.02$ & $0.24$ & $0.61$ & $0.61$ \\
\bottomrule
\end{tabular}%
}
\end{table}

\subsection{Temporal and perturbation-value analyses}
\label{sec:temporal-perturbation-appendix}

We examine how measured phenotypes, model rankings, paired prediction quality, and temporal ordering change over time. These analyses use the paired A549 test set on its 2-hour grid, $t=1,\dots,10$ spanning $5$--$23$ hpi, with 12 fields per organelle--condition combination. The adjacent-frame analysis instead uses the unpaired 10-min movie described below.

\paragraph{Paired prediction quality over time.}
We measure Spectral PCC against the paired fluorescence on the A549 Mock series. In the Mock condition, Spectral PCC varies modestly over time, although the leading model changes for ER. Across five A549-trained models and four organelles, the temporal coefficient of variation ranges from $0.9\%$ to $2.7\%$, and the change from the first to the last timepoint ranges from $-0.044$ to $+0.017$. Pix2Pix3D remains the leading model for nucleus and membrane, and FNet3D remains the leader for mitochondria; for ER, the leader changes from VSCyto3D to Pix2Pix3D (Tab.~\ref{tab:temporal-paired-robustness}).

\begin{table}[t]
\centering
\small
\caption{\textbf{Paired prediction quality over the A549 Mock time course.} At each of ten 2-h timepoints from 5 to 23 hpi, Spectral PCC is first averaged over 12 FOVs. Ranges then summarize five A549-trained models across time. CV is the temporal coefficient of variation, and score changes are T10 minus T1. This analysis measures agreement with paired fluorescence, not frame-to-frame consistency.}
\label{tab:temporal-paired-robustness}
\resizebox{0.88\linewidth}{!}{%
\begin{tabular}{lccc}
\toprule
\textbf{Organelle} & \textbf{CV range (\%)} & \textbf{T10 - T1 range} & \textbf{\shortstack{Leader\\T1 $\rightarrow$ T10}} \\
\midrule
\textbf{Nucleus} & 1.0--1.9 & -0.039 to -0.019 & Pix2Pix3D $\rightarrow$ Pix2Pix3D \\
\textbf{Membrane} & 1.0--2.5 & -0.004 to +0.006 & Pix2Pix3D $\rightarrow$ Pix2Pix3D \\
\textbf{ER} & 0.9--2.7 & -0.029 to +0.017 & VSCyto3D $\rightarrow$ Pix2Pix3D \\
\textbf{Mitochondria} & 1.2--1.9 & -0.044 to -0.033 & FNet3D $\rightarrow$ FNet3D \\
\bottomrule
\end{tabular}%
}
\end{table}
 
Fix a feature extractor and let $\mathbf{f}$ denote one single-cell embedding. Embeddings are standardized coordinatewise against the pooled real-fluorescence distribution $\mathcal{F}_g$, and the perturbation at timepoint $t$ is the difference between the infected and Mock population means in that standardized space:
\begin{equation}
  \tilde{\mathbf{f}} = \frac{\mathbf{f} - \operatorname{med}(\mathcal{F}_g)}
                            {1.4826\operatorname{MAD}(\mathcal{F}_g)},
  \qquad
  \mathbf{d}_{c,t} = \bar{\tilde{\mathbf{f}}}_{c,t} - \bar{\tilde{\mathbf{f}}}_{\mathrm{mock},t},
  \qquad c \in \{\mathrm{DENV}, \mathrm{ZIKV}\}.
  \label{eq:perturbation-shift}
\end{equation}
The scale $1.4826\operatorname{MAD}$ is the median absolute deviation rescaled to estimate the standard deviation of a Gaussian, so no coordinate dominates $\mathbf{d}$ through its units. Real and predicted embeddings are standardized with the same real-fluorescence median and scale, which keeps the two shifts in one frame of reference.

\paragraph{Probe-transfer decay (Tab.~\ref{tab:temporal-probe-transfer}).}
A linear probe is fitted on frame $t$ to separate Mock from infected cells and evaluated on frame $t'$, with lag $\ell = |t - t'|$ and elapsed time $2\ell$ hours. Fits are FOV-stratified, so the training and evaluation cells come from disjoint fields and the probe cannot memorize a field. Performance is the area under the ROC curve in its Mann--Whitney form,
\begin{equation}
  \mathrm{AUROC} = \frac{1}{n_+ n_-}\sum_{i=1}^{n_+}\sum_{j=1}^{n_-}
  \Big[\mathbf{1}\{s^+_i > s^-_j\} + \tfrac{1}{2}\mathbf{1}\{s^+_i = s^-_j\}\Big],
  \label{eq:auroc}
\end{equation}
the probability that a randomly drawn infected cell receives a higher score $s$ than a randomly drawn Mock cell, with $0.5$ as chance. There are $10$ ordered frame pairs at zero lag and $2(10-\ell)$ at lag $\ell$. We fit a probe for each pair in each of the $24$ organelle--virus--feature-space series. The last column reports these counts, from $24 \times 10 = 240$ fits at zero lag down to $24 \times 2 = 48$ at $18$ h. Pooled AUROC falls monotonically from $0.644$ to $0.523$, within $0.023$ of chance at the longest lag. The infection signature is therefore specific to the frame it was learned on, and collecting more fields at a single timepoint cannot recover what elapsed time removes.

\begin{table*}[t]
\centering
\normalsize
\caption{\textbf{Probe-transfer decay across the paired 2-h time grid.} Each row groups independently fitted, FOV-stratified probes by train--test frame lag. Intervals are standard errors of row means. Pooled AUROC falls monotonically from 0.644 at zero lag to 0.523 at 18 h; the far-lag deficit is 0.091 in all 24 organelle--virus--space series. More fields at one frame cannot remove this elapsed-time effect.}
\label{tab:temporal-probe-transfer}
\resizebox{\ifdim\width>\textwidth \textwidth\else\width\fi}{!}{%
\begin{tabular}{lccccc}
\toprule
\textbf{train/test frame gap} & \textbf{elapsed} & \textbf{AUROC, DENV $\uparrow$} & \textbf{AUROC, ZIKV $\uparrow$} & \textbf{pooled $\uparrow$} & \textbf{probe fits averaged} \\
\midrule
same frame & 0 h & 0.639 $\pm$ 0.009 & 0.649 $\pm$ 0.009 & \textbf{0.644} $\pm$ 0.007 & 240 \\
$\pm$ 1 & 2 h & 0.628 $\pm$ 0.006 & 0.638 $\pm$ 0.006 & \textbf{0.633} $\pm$ 0.004 & 432 \\
$\pm$ 2 & 4 h & 0.609 $\pm$ 0.006 & 0.629 $\pm$ 0.006 & \textbf{0.619} $\pm$ 0.004 & 384 \\
$\pm$ 3 & 6 h & 0.596 $\pm$ 0.006 & 0.607 $\pm$ 0.007 & \textbf{0.601} $\pm$ 0.005 & 336 \\
$\pm$ 4 & 8 h & 0.579 $\pm$ 0.006 & 0.598 $\pm$ 0.007 & \textbf{0.588} $\pm$ 0.005 & 288 \\
$\pm$ 5 & 10 h & 0.560 $\pm$ 0.006 & 0.592 $\pm$ 0.007 & \textbf{0.576} $\pm$ 0.004 & 240 \\
$\pm$ 6 & 12 h & 0.540 $\pm$ 0.006 & 0.565 $\pm$ 0.007 & \textbf{0.553} $\pm$ 0.005 & 192 \\
$\pm$ 7 & 14 h & 0.540 $\pm$ 0.007 & 0.551 $\pm$ 0.008 & \textbf{0.546} $\pm$ 0.005 & 144 \\
$\pm$ 8 & 16 h & 0.522 $\pm$ 0.009 & 0.529 $\pm$ 0.010 & \textbf{0.526} $\pm$ 0.007 & 96 \\
$\pm$ 9 & 18 h & 0.519 $\pm$ 0.014 & 0.528 $\pm$ 0.015 & \textbf{0.523} $\pm$ 0.010 & 48 \\
\bottomrule
\end{tabular}%
}
\end{table*}
 
\paragraph{Phenotype progression (Tab.~\ref{tab:temporal-phenotype-progression}).}
The same probe evaluated at zero lag measures how separable the two conditions are within a frame. From the first to the last timepoint, within-frame AUROC rises in all $24$ organelle--virus--feature-space combinations, from $0.559$ to $0.743$ on average. The shift of Eq.~\eqref{eq:perturbation-shift} grows over the same window by $g_c = \|\mathbf{d}_{c,10}\|_2 / \|\mathbf{d}_{c,1}\|_2 - 1$, which is $26\%$ for DENV and $22\%$ for ZIKV. The classifier and embedding-distance analyses therefore show the same increase in infection-associated phenotype separation.

\begin{table*}[t]
\centering
\normalsize
\caption{\textbf{Real-fluorescence infection phenotype from frame 1 to frame 10.} Within-frame AUROC rises in all 24 organelle--virus--feature-space combinations, from 0.559 to 0.743 on average. Independently, the real mock-to-infected mean-embedding shift grows by 24\% (DENV 26\%, ZIKV 22\%). The increasing underestimation of response amplitude is specific to DENV and is not pooled across viruses here.}
\label{tab:temporal-phenotype-progression}
\begin{tabular}{lcccc}
\toprule
\textbf{Organelle} & \textbf{Virus} & \textbf{DINOv3 $\uparrow$} & \textbf{CELL-DINO $\uparrow$} & \textbf{MorphEm $\uparrow$} \\
\midrule
Membrane & DENV & 0.49 $\rightarrow$ \textbf{0.54} & 0.55 $\rightarrow$ \textbf{0.76} & 0.53 $\rightarrow$ \textbf{0.72} \\
Membrane & ZIKV & 0.47 $\rightarrow$ \textbf{0.77} & 0.55 $\rightarrow$ \textbf{0.92} & 0.47 $\rightarrow$ \textbf{0.87} \\
Nucleus & DENV & 0.53 $\rightarrow$ \textbf{0.71} & 0.54 $\rightarrow$ \textbf{0.66} & 0.45 $\rightarrow$ \textbf{0.58} \\
Nucleus & ZIKV & 0.51 $\rightarrow$ \textbf{0.67} & 0.64 $\rightarrow$ \textbf{0.79} & 0.56 $\rightarrow$ \textbf{0.67} \\
ER & DENV & 0.56 $\rightarrow$ \textbf{0.70} & 0.66 $\rightarrow$ \textbf{0.83} & 0.64 $\rightarrow$ \textbf{0.79} \\
ER & ZIKV & 0.51 $\rightarrow$ \textbf{0.72} & 0.64 $\rightarrow$ \textbf{0.82} & 0.51 $\rightarrow$ \textbf{0.76} \\
Mito & DENV & 0.54 $\rightarrow$ \textbf{0.76} & 0.72 $\rightarrow$ \textbf{0.88} & 0.56 $\rightarrow$ \textbf{0.87} \\
Mito & ZIKV & 0.50 $\rightarrow$ \textbf{0.60} & 0.70 $\rightarrow$ \textbf{0.77} & 0.59 $\rightarrow$ \textbf{0.68} \\
\bottomrule
\end{tabular}
\end{table*}
 
\paragraph{Frame-dependent ranking (Tab.~\ref{tab:temporal-ranking-sensitivity}).}
Within each evaluation setting, we retain FOVs with valid scores at every timepoint for every compared model. The metric value $M_{i,t}$ of model $i$ at frame $t$ decomposes as
\begin{equation}
  M_{i,t} = \mu + \alpha_i + \beta_t + \gamma_{i,t} + \varepsilon_{i,t},
  \label{eq:frame-decomposition}
\end{equation}
with $\alpha_i$ the model effect, $\beta_t$ the shared frame effect, and $\gamma_{i,t}$ their interaction. The shared frame effect changes every model equally and cannot change their order; the interaction can. Tab.~\ref{tab:temporal-ranking-sensitivity} reports the standard deviation of both terms, split-half agreement of the model means, and the size of the interaction relative to the gap between the top two models. Its final row reports how often the best model at one frame differs from the best model over the full window. The analysis pools all four organelles. Phenotype is measured with FID rather than KID because a single frame supplies too few cells for subset-resampling KID~(App.~\ref{sec:feat_simi}).

\begin{table*}[t]
\centering
\normalsize
\caption{\textbf{Frame-dependent model ranking.} The model--frame interaction is the component that can reorder models; the shared frame effect cancels. Across the full grid, a one-frame winner differs from the window winner in 443/1800 (25\%) pixel/segmentation comparisons and 63/720 (9\%) phenotype comparisons. Dice is instance Dice for nucleus and membrane and semantic Dice for ER and mitochondria. Phenotype uses FID because per-frame KID coverage is insufficient; this table measures only ranking movement, not which model is best.}
\label{tab:temporal-ranking-sensitivity}
\resizebox{\ifdim\width>\textwidth \textwidth\else\width\fi}{!}{%
\begin{tabular}{@{}lccccccc@{}}
\toprule
 & \multicolumn{4}{c}{\textbf{Pixel}} & \textbf{Seg.} & \multicolumn{2}{c}{\textbf{Phenotype (FID)}} \\
\cmidrule(lr){2-5}\cmidrule(lr){6-6}\cmidrule(lr){7-8}
\textbf{Statistic} & \textbf{\shortstack{Spectral\\PCC}} & \textbf{PCC} & \textbf{SSIM} & \textbf{si-SSIM} & \textbf{Dice} & \textbf{\shortstack{CELL-\\DINO}} & \textbf{DINOv3} \\
\midrule
Frame effect SD & 0.012 & 0.020 & 0.025 & 0.019 & 0.018 & 84.0 & 36.8 \\
Model $\times$ frame SD $\downarrow$ & \textbf{0.005} & \textbf{0.007} & \textbf{0.017} & \textbf{0.005} & \textbf{0.006} & \textbf{41.2} & \textbf{17.1} \\
Split-half r $\uparrow$ & +0.27 & +0.44 & +0.42 & +0.49 & +0.33 & +0.45 & +0.19 \\
Vs. top-two gap $\downarrow$ & \textbf{0.80$\times$} & \textbf{0.57$\times$} & \textbf{0.31$\times$} & \textbf{0.72$\times$} & \textbf{0.70$\times$} & \textbf{0.03$\times$} & \textbf{0.08$\times$} \\
1-frame $\ne$ window winner $\downarrow$ & \shortstack{113/360\\(31\%)} & \shortstack{85/360\\(24\%)} & \shortstack{62/360\\(17\%)} & \shortstack{75/360\\(21\%)} & \shortstack{108/360\\(30\%)} & \shortstack{16/360\\(4\%)} & \shortstack{47/360\\(13\%)} \\
\bottomrule
\end{tabular}%
}
\end{table*}
 
\begin{figure}[ht]
  \centering
  \includegraphics[width=0.98\linewidth]{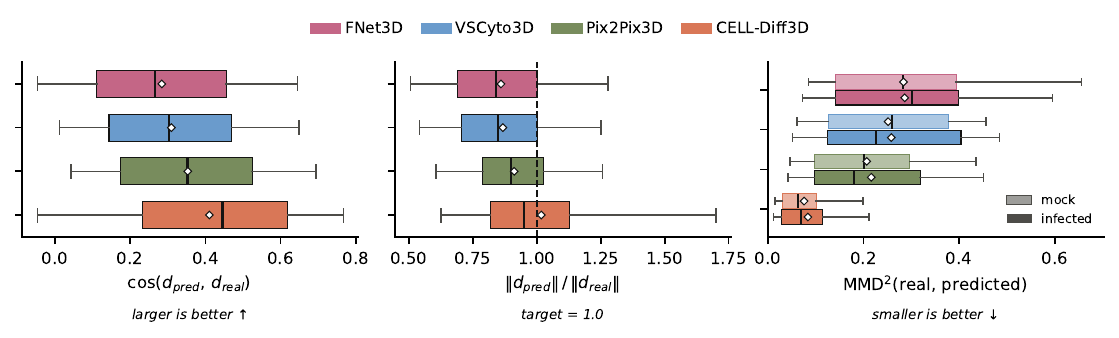}
  \caption{\textbf{Agreement with the A549 infection response.} For each organelle, training set, virus, feature space, and frame, $d$ is the infected-minus-Mock difference between mean single-cell embeddings. Before computing $d$, real and predicted embeddings are standardized by the pooled real-fluorescence median and robust scale ($1.4826$ times the median absolute deviation). The left panel reports the cosine between the predicted and real shifts; the middle panel reports their norm ratio, for which 1 is the target. The right panel reports unbiased RBF MMD$^2$ between predicted and real embeddings within Mock and infected conditions; DENV and ZIKV are pooled only for display. Each box contains 720 settings (4 organelles $\times$ 3 training sets $\times$ 2 viruses $\times$ 3 learned feature spaces $\times$ 10 frames). Boxes show the median and interquartile range, whiskers the 5th--95th percentiles, and diamonds the mean. CELL-Diff3D has the greatest mean direction agreement and the lowest distributional distance. Because the marginal distributions overlap, model comparisons use the paired tests in Tab.~\ref{tab:perturbation-response}.}
  \label{fig:perturbation-response-axis}
\end{figure}

\paragraph{Perturbation response (Tab.~\ref{tab:perturbation-response}).}
Let $\mathbf{d}^{g}_{c,t}$ denote the experimental shift of Eq.~\eqref{eq:perturbation-shift} and $\mathbf{d}^{m}_{c,t}$ the same quantity computed from model $m$'s predictions. The table reports
\begin{equation}
  \cos\theta = \frac{\langle \mathbf{d}^{m}_{c,t},\, \mathbf{d}^{g}_{c,t}\rangle}
                    {\|\mathbf{d}^{m}_{c,t}\|_2\,\|\mathbf{d}^{g}_{c,t}\|_2},
  \qquad
  a = \frac{\|\mathbf{d}^{m}_{c,t}\|_2}{\|\mathbf{d}^{g}_{c,t}\|_2},
  \label{eq:perturbation-cosine}
\end{equation}
where $\cos\theta$ measures direction agreement and $a$ is the amplitude ratio, with a target of $1$. Because each population supplies its own shift, an offset common to a model's Mock and infected predictions leaves $\mathbf{d}^{m}$ unchanged. The comparison therefore measures whether a model reproduces the infection-associated change rather than whether individual predictions are accurate. Pooled over all four organelles, CELL-Diff3D leads under both viruses and wins $61$--$74\%$ of $480$ matched comparisons against every regression and adversarial baseline (two-sided Wilcoxon signed-rank test, $p<10^{-5}$). These $480$ comparisons are the subset of the $720$ settings in Fig.~\ref{fig:perturbation-response-axis} with results for all six models, including the UNeXt2 and UNetViT3D ablations. The $\mathrm{MMD}^2$ in that figure uses an unbiased RBF-kernel estimator rather than the polynomial-kernel KID of App.~\ref{sec:feat_simi}, so their absolute values are not comparable.
\begin{table*}[t]
\centering
\normalsize
\caption{\textbf{Perturbation-response agreement.} Cosine similarity compares the direction of the predicted Mock-to-infected mean-embedding shift with that measured in real fluorescence. An amplitude ratio of 1 indicates equal shift magnitudes. CELL-Diff3D leads for both viruses and wins 61--74\% of 480 paired comparisons against each regression or adversarial baseline (two-sided Wilcoxon $p<10^{-5}$).}
\label{tab:perturbation-response}
\resizebox{\ifdim\width>\textwidth \textwidth\else\width\fi}{!}{%
\begin{tabular}{lccccc}
\toprule
\textbf{Model} & \textbf{cos, DENV $\uparrow$} & \textbf{cos, ZIKV $\uparrow$} & \textbf{amp, DENV $\rightarrow$1} & \textbf{amp, ZIKV $\rightarrow$1} & \textbf{CELL-Diff3D cos win, pooled $\uparrow$} \\
\midrule
FNet3D & 0.270 & 0.298 & 0.876 & 0.845 & 74\% (4e-29) \\
UNeXt2 & 0.314 & 0.329 & 0.857 & 0.889 & 67\% (5e-14) \\
VSCyto3D & 0.295 & 0.324 & 0.858 & 0.878 & 69\% (4e-17) \\
UNetViT3D & 0.274 & 0.245 & 0.890 & 0.939 & 66\% (9e-13) \\
Pix2Pix3D & 0.333 & 0.373 & 0.901 & 0.923 & 61\% (7e-06) \\
\textbf{CELL-Diff3D} & \textbf{0.401} & \textbf{0.420} & 0.996 & 1.040 & -- \\
\bottomrule
\end{tabular}%
}
\end{table*}
 
\paragraph{Temporal-order retention (Tab.~\ref{tab:temporal-order-retention}).}
A FOV-stratified ridge probe regresses the timepoint index on single-cell embeddings. Let $\rho$ denote the Spearman rank correlation between predicted and true timepoints on held-out fields. Retention is
\begin{equation}
  R = \frac{\rho_{\mathrm{pred}}}{\rho_{\mathrm{real}}},
  \label{eq:order-retention}
\end{equation}
normalized against the real-fluorescence ceiling measured in the same condition. Thus, $R=1$ means the predictions carry as much temporal ordering as the experimental images, and a temporally flat prediction gives $R=0$. After pooling over DINOv3, CELL-DINO, and MorphEm on the common model subset, we find that every model loses temporal signal under infection, by $0.129$ to $0.232$, with CELL-Diff3D losing the most.

\begin{table}[t]
\centering
\normalsize
\caption{\textbf{Temporal-order retention.} A FOV-stratified ridge probe predicts timepoint from single-cell embeddings. Retention is the probe's Spearman $\rho$ on predicted fluorescence divided by its value on real fluorescence. The summary pools DINOv3, CELL-DINO, and MorphEm over the common model subset. In the infected conditions, all models' predictions retain less of the time-dependent change seen in experimental fluorescence: every model loses temporal signal under infection (decrease 0.129--0.232); a temporally flat prediction scores zero.}
\label{tab:temporal-order-retention}
\resizebox{\ifdim\width>\linewidth \linewidth\else\width\fi}{!}{%
\begin{tabular}{lccc}
\toprule
\textbf{retention} & \textbf{FNet3D} & \textbf{VSCyto3D} & \textbf{CELL-Diff3D} \\
\midrule
\textbf{mock} & 0.868 & 0.760 & 0.849 \\
\textbf{infected} & 0.739 & 0.592 & 0.616 \\
\textbf{drop (mock $-$ infected)} & +0.129 & +0.167 & +0.232 \\
\bottomrule
\end{tabular}%
}
\end{table}
 
\begin{figure}[ht]
  \centering
  \includegraphics[width=0.96\linewidth]{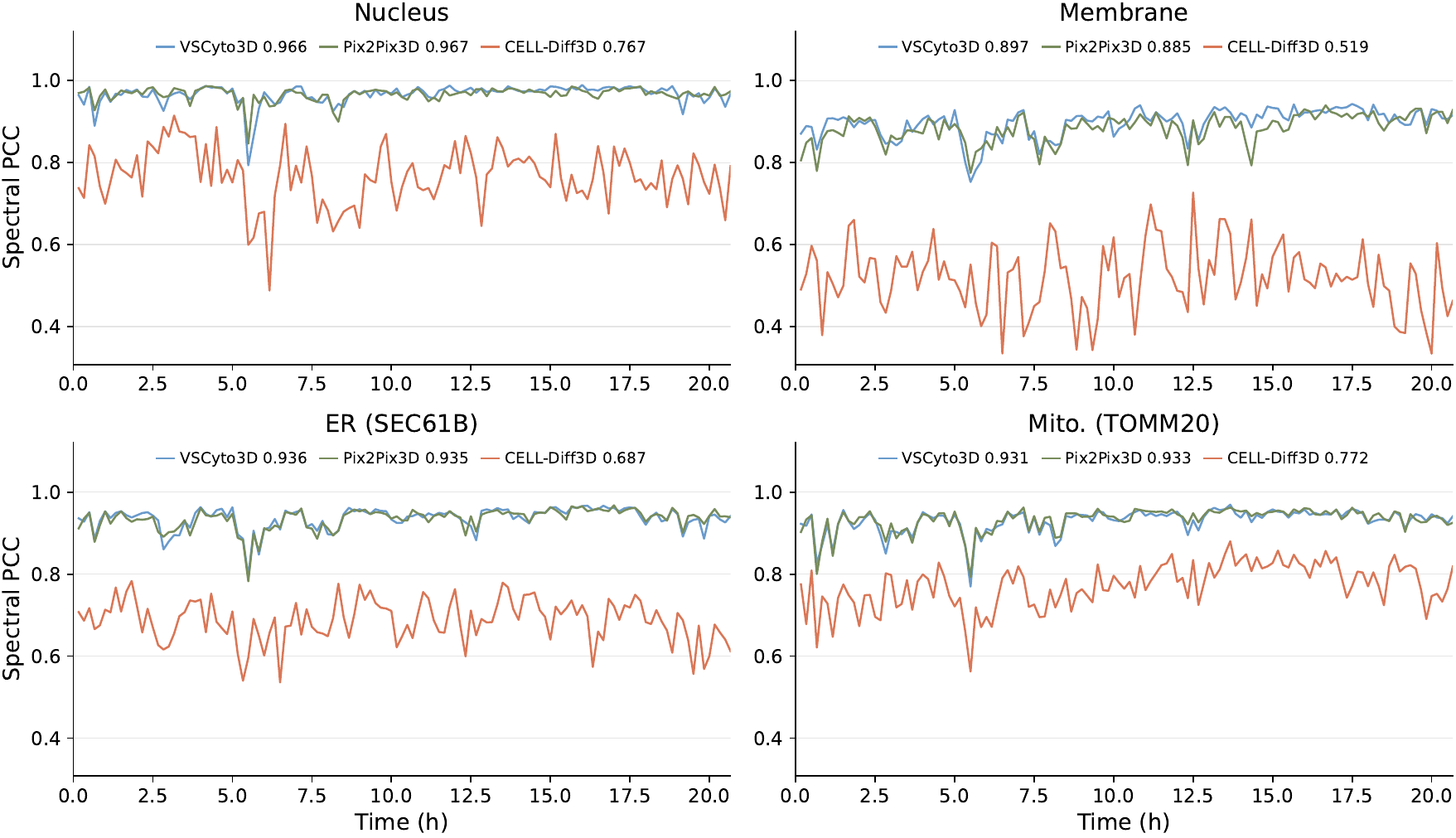}
  \caption{\textbf{Reference-free frame-to-frame self-consistency on one dense A549 movie.} The same uninfected A549 field was processed by A549-trained models for each of the four organelles. Lines show Spectral PCC between each predicted 3D volume and the preceding prediction for 124 adjacent pairs from 125 frames acquired every 10 min; the number beside each model in the legend is its mean over those pairs. The mean adjacent-frame Spectral PCC of the regression (VSCyto3D) and adversarial (Pix2Pix3D) predictions lies between $0.885$ and $0.967$ across the four organelles, whereas that of the generative CELL-Diff3D predictions ranges from $0.519$ to $0.772$. Because the field has no paired fluorescence, this comparison measures prediction self-consistency rather than temporal accuracy.}
  \label{fig:temporal-frame-to-frame}
\end{figure}

The diagnostic uses one uninfected Mantis A549 field with $125$ phase volumes of $48\times512\times512$ voxels spanning 20 h 40 min. CELL-Diff3D is the least self-consistent model in all four organelles, while VSCyto3D and Pix2Pix3D differ by less than $0.013$ in each organelle~(Fig.~\ref{fig:temporal-frame-to-frame}). These results identify output instability in this field but cannot measure agreement with the unobserved fluorescence trajectory.

\FloatBarrier

\begin{figure}[!ht]
  \centering
  \includegraphics[width=\linewidth]{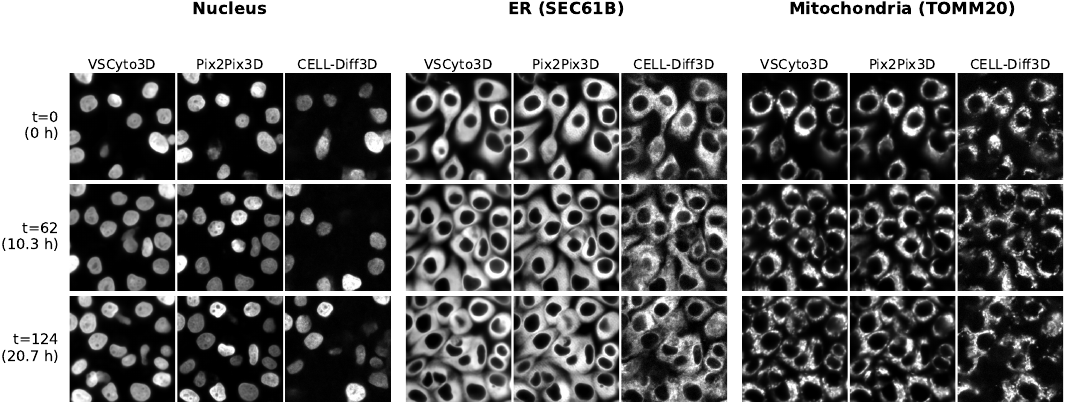}
  \caption{\textbf{Representative predictions from the dense unpaired A549 movie.} Central XY slices at the beginning, midpoint, and end of the 20 h 40 min movie are shown for nucleus, ER, and mitochondria. Membrane predictions from the same field appear in Fig.~\ref{fig:a549-robustness}C. Each image is clipped independently to its 1st--99th percentiles. The field has no paired fluorescence, so the panels show prediction appearance over time rather than accuracy.}
  \label{fig:movie-selected-organelles}
\end{figure}

\section{Detailed limitations and future work}
\label{sec:limitations-detail}

\paragraph{Metrics.}
All three metric tiers have limitations. Spectral PCC suppresses noise-dominated frequencies and is therefore less sensitive to blur. The segmentation workflows have not been validated against manual annotations of infected cells. The phenotype tier uses 2D projections for DINOv3, DynaCLR, CELL-DINO, and MorphEm, while GLCM+ remains 3D. Because these feature extractors do not always rank models in the same order, our claims apply to the measured feature spaces rather than to phenotype in general. A future infection-classifier evaluation should calibrate conditions jointly because separate scaling produces a same-condition null AUROC of $0.576$.

\paragraph{Data and time.}
The released test movies cover A549 on Mantis and one perturbation family; iPSC and HEK293T are 3D snapshots. Cell type and microscope therefore change together in the main transfer experiment. HEK293T and A549 were both imaged on Mantis, but their fluorescence markers differ. Fully separating cell type, marker, and microscope effects would require imaging the same cell type and markers on both microscopes. The source plates also use different native time intervals and are subsampled to a common 2-hour grid. The one dense 10-min movie is unpaired, so its reference-free diagnostic measures prediction self-consistency rather than temporal accuracy. Densely sampled paired movies are needed to compare recurrent, video-diffusion, or motion-aware objectives. Future releases will also add broader perturbations and paired multi-organelle volumes across cell types.

\paragraph{Baselines.}
The baseline set contains regression, adversarial, conditional-mean flow-matching, and generative flow-matching controls. Future comparisons should include full-noise diffusion, latent-space samplers, autoregressive predictors, and multi-organelle training. Multi-organelle models may introduce signal from one target channel into another and require a dedicated evaluation of this failure mode.

\paragraph{Scope of DynaCell evaluation scores.}
DynaCell evaluation scores support conclusions for the measured organelles, cell lines, microscopes, perturbations, and phenotype representations. Clinical use, rare phenotypes, and long-range temporal behavior require dedicated validation data.

\FloatBarrier

\addtocontents{toc}{\protect\setcounter{tocdepth}{1}}
\section{Datasheet for the A549 and iPSC datasets}
\label{sec:datasheet}

This datasheet follows Gebru et al.\ (\textit{Datasheets for Datasets}, Commun.\ ACM 2021).

\subsection{Motivation}

\paragraph{Why were the datasets assembled?}
Most existing virtual-staining benchmarks are 2D; none combines volumetric time-lapse data with the same pixel, segmentation, and phenotype scoring tiers as the DynaCell evaluation suite.
The suite uses paired label-free and fluorescence 3D microscopy from the A549 and iPSC datasets to compare regression and generative virtual-staining models. The A549 dataset includes matched perturbation conditions and time-lapse imaging. Its released paired cadence supports time-resolved robustness analysis but not dense frame-to-frame temporal-consistency benchmarking.

\paragraph{Who funded this work?}
Data processing and release were funded by Biohub San Francisco (Biohub SF).
Hosting is supported by the AWS Open Data Sponsorship Program.

\paragraph{What tasks do the datasets support?}
The datasets enable regression and generative \emph{virtual staining} from label-free input, evaluated with three types of metrics: pixel fidelity, organelle-segmentation utility, and single-cell phenotype similarity.

\subsection{Composition}

\paragraph{What does each instance represent?}
A field of view (FOV): a 3D+time volume providing a label-free brightfield channel paired with a fluorescence channel for one of four organelle targets (nucleus, plasma membrane, endoplasmic reticulum, mitochondria).

\paragraph{How many instances are there?}
The A549 v1 release comprises 262 FOVs across 24 OME-Zarr archives (\(\sim\)407\,GB). A balanced test set contains 144 FOVs: 12 FOVs per organelle and condition across four organelles and three conditions (Mock, ZIKV, and DENV). The remaining curated FOVs form the training set.
The supplementary material contains an unpaired 125-frame A549 movie from a separate untreated FOV at its native 10-min cadence. It is not part of the paired benchmark count.
The iPSC training set contains 500 FOVs each for nucleus, plasma membrane, and mitochondria and 423 FOVs for ER (1{,}923 FOVs total).
The iPSC held-out evaluation subset contains 100 FOVs per organelle (400 FOVs total).
The DynaCell evaluation suite also uses eight HEK293T Mantis volumes from S-BIAD1287 (four membrane and four mitochondria) for an evaluation ablation. They are not part of the A549 or iPSC training and test splits.

\paragraph{What data does each instance consist of?}
Each FOV carries (a)~quantitative phase reconstructed from a brightfield z-stack via \texttt{waveorder}~\citep{chandler2024waveorder}; (b)~a fluorescence volume registered to the phase channel; (c)~per-FOV metadata covering cell line, condition, plate, well, acquisition timepoint, and quality-control flags.
Volumes are stored as RFC-9 zipped OME-Zarr archives (\texttt{.ozx}) following OME-NGFF v0.5.
Mantis voxel spacing is \(0.174\times 0.1494\times 0.1494\,\mu\)m (ZYX); A549 test FOVs are \(48\times640\times960\) and contain approximately 17,000 segmented cells across the four organelles. iPSC training FOVs are \(44\times 624\times 924\), iPSC test FOVs are \(40\times 512\times 512\), and HEK293T ablation FOVs are \(64\times960\times1184\) (all ZYX).

\paragraph{Are there labels or targets?}
Raw fluorescence channels serve as the prediction targets for all four organelles.
No manual per-cell annotations are provided. Cellpose v4 cpdino-ViT-L provides 2D nucleus and membrane instances, and Allen Institute workflows provide semantic ER and mitochondria masks.

\paragraph{Is there missing information?}
Each FOV carries one organelle target; balanced coverage across organelles and conditions is achieved across each dataset, not per FOV.
Frames affected by acquisition or registration failures, and frames flagged by manual biological quality control (missing cells or absent target fluorescence), are dropped during curation. Per-frame metrics are averaged over retained FOV--timepoint pairs, so FOVs contribute in proportion to their retained frames; tables state any metric-specific aggregation.

\paragraph{Are there recommended data splits?}
Yes.
The A549 test set (144 FOVs) is the held-out benchmark for the primary A549 results; well positions are disjoint from the training set.
The iPSC held-out subset (100 FOVs per organelle) provides an in-distribution evaluation split.
Researchers may train on the iPSC training set alone, on the A549 training set, or jointly to compare cross-cell-type generalization.
For each model, checkpoint selection uses an 80/20 validation split from the training FOVs. Test stores are separate and never used for hyperparameter or checkpoint decisions; the validation split is reproducible per configuration but is not one shared partition across all model families.

\subsection{Collection process}

\paragraph{How was the A549 component acquired?}
A549 cells (ATCC CCL-185) carrying fluorescent markers for H2B (nucleus), a CAAX membrane-targeting motif (plasma membrane), SEC61B (ER), and TOMM20 (mitochondria)~\citep{cidrosas2026vswitch,czbiohub2025organellebox} were imaged on the Mantis correlative label-free and light-sheet fluorescence microscope~\citep{ivanov2024mantis} at Biohub SF.
Mantis acquires co-registered label-free and fluorescence channels simultaneously.
Plates were imaged at native cadences of 10, 30, or 120 minutes within an evaluation window of approximately 5--23 hours post-infection (hpi) and subsampled to a shared 2-hour grid aligned to integer hpi (\(\sim\)10 frames per FOV) before splitting.
Zika and Dengue infections were performed under institutional BSL-2 protocols; the digital image release contains no physical or infectious material.

\paragraph{How was the iPSC component acquired?}
The iPSC component is derived from the Allen Institute WTC-11 hiPSC dataset~\citep{viana2023integrated}.
WTC-11 cells carry endogenous fluorescent tags for ER (SEC61B-mScarlet) and mitochondria (TOMM20-mScarlet); nucleus (Hoechst) and plasma membrane (CellMask) are chemically stained.
Imaging was performed on a 3D spinning-disk confocal microscope.
We reprocessed the raw transmitted-light data with \texttt{waveorder} to obtain quantitative phase paired with the original fluorescence channel.

\paragraph{Over what time span were data collected?}
The A549 dataset was collected from March 2025 to March 2026.

\subsection{Preprocessing, cleaning, and labeling}
\label{sec:data-preprocessing-details}

\paragraph{What preprocessing was applied?}
Brightfield z-stacks are reconstructed to quantitative phase using \texttt{waveorder}~\citep{chandler2024waveorder} (BSD-3-Clause; \url{https://github.com/Biohub/waveorder}).
Oblique light-sheet fluorescence volumes from Mantis are deskewed.
Phase and fluorescence channels are registered using a separate calibration field of view containing fluorescent beads~\citep{ivanov2024mantis}. Detected beads form nodes in a graph for each modality. Matching the graphs by spatial similarity yields a per-timepoint affine transformation.
The aligned volumes are cropped in the lateral (XY), axial (Z), and temporal (T) dimensions. In XY, only regions with valid overlap between modalities are retained. The axial crop extends one-third of its depth below the phase focus plane and two-thirds above it. Frames affected by acquisition or registration failures are excluded.
A final manual inspection of each timepoint flags biological and imaging failures that automated checks miss, including missing cells, absent target fluorescence, residual misregistration, out-of-focus images, and imaging artifacts (Fig.~\ref{fig:preprocessing}A). Temporal subsampling and dataset splits are described in Sec.~\ref{sec:splits}.

\paragraph{What is released?}
We release the registered label-free and fluorescence datasets along with model checkpoints used for benchmark evaluation via AWS Open Data.
Each archive groups multiple FOVs for one organelle--condition--split combination. A549 archives carry Phase3D, Brightfield, and a raw fluorescence target for H2B, CAAX, SEC61B, or TOMM20; iPSC archives carry four to five channels (Phase3D, Brightfield, Nuclei, Membrane, and Structure where the target is SEC61B or TOMM20).
Preprocessing and evaluation code are released at \url{https://github.com/Biohub/VisCy/tree/main/applications/dynacell} (BSD-3-Clause).

\paragraph{Was software development assisted by large language models?}
Yes.
LLMs assisted routine code completion and refactoring. The authors specified the scientific methods, reviewed assisted changes, and tested the affected pipelines.
LLMs were not used to design imaging experiments, choose biological conditions, or interpret the results.

\subsection{Uses}

\paragraph{What tasks have these datasets been used for?}
The DynaCell evaluation suite uses these datasets to benchmark four regression architectures (FNet3D, UNeXt2, VSCyto3D, and UNetViT3D), the adversarial model Pix2Pix3D, and CELL-Diff3D under both single-pass and sampled inference. Models were trained on iPSC, A549, or their joint set and evaluated with three metric tiers on in-distribution and transfer test sets (Sec.~\ref{sec:results}).

\paragraph{What other tasks could they support?}
The datasets could support self-supervised and contrastive representation learning on 3D live-cell dynamics, perturbation-biology research using virtually stained predictions, multitask learning across organelle targets, and domain adaptation between microscope platforms.

\paragraph{What tasks should they not be used for?}
Virtual stains evaluated with the DynaCell evaluation suite \emph{must not} be used for clinical or diagnostic decision-making.
Virtual stains are model predictions, not direct measurements; their accuracy on cell types, perturbation agents, or imaging platforms not represented in these datasets has not been validated.
Perturbation-response scores computed from virtually stained images are research-only model-derived signals, not clinical assays.

\subsection{Distribution}

\paragraph{How are the datasets distributed?}
The datasets are hosted through the AWS Open Data Sponsorship Program at \texttt{s3://dynacell/v1/} (HTTPS: \url{https://dynacell.s3.us-west-2.amazonaws.com/v1/}), accessible without registration.
The Registry of Open Data on AWS landing page at \url{https://registry.opendata.aws/dynacell/} provides discovery, access instructions, and citation metadata.
A 745\,MiB demo subsample with one A549 field of view, for inspecting the data without downloading the full release, is available at \texttt{s3://dynacell/v1/demo/}.

\paragraph{Under what license?}
The A549 dataset is released under CC-BY-4.0 (\url{https://creativecommons.org/licenses/by/4.0/}).
The iPSC dataset is redistributed under the upstream Allen Institute Terms of Use (\url{https://www.allencell.org/terms-of-use.html}, noncommercial research), which require citation of the source datasets~\citep{aics2018hipsc,viana2023integrated}.
Code is released under BSD-3-Clause.

\paragraph{Are there restrictions?}
The CC-BY-4.0 attribution requirement applies to the A549 dataset.
The iPSC dataset's restrictions are inherited from the Allen Institute Terms of Use: use is limited to noncommercial research and requires citation; commercial redistribution requires written permission from the Allen Institute.

\paragraph{Will a DOI or persistent identifier be provided?}
Not yet. The datasets have a stable URL at \href{https://dynacell.s3.us-west-2.amazonaws.com/v1/README.md}{https://dynacell.s3.us-west-2.amazonaws.com/v1/README.md}; an AWS Open Data registry descriptor is pending review.

\subsection{Maintenance}

\paragraph{Who maintains the datasets?}
The Computational Imaging Group at Biohub SF (Mehta Lab) hosts and maintains the datasets.
The named technical contact is Shalin B.\ Mehta (\texttt{shalin.mehta@czbiohub.org}).

\paragraph{How can errors be reported?}
Via GitHub Issues at \url{https://github.com/Biohub/VisCy} or by email to the contact above.

\paragraph{What is the versioning and update policy?}
Versions follow semantic versioning (\texttt{vMAJOR.MINOR.PATCH}).
The A549 test split is frozen for the v1 series to preserve benchmark integrity; new cell lines or microscope platforms will be added as new minor versions without modifying existing FOVs.
Biohub commits to long-term maintenance of the AWS Open Data hosting and the GitHub code repository, with no plans for removal.

\subsection{Cell-line provenance and ethics}

\paragraph{A549 (human lung adenocarcinoma).}
A549 is an immortalized cell line derived from human lung carcinoma tissue (Giard et al., 1973), available from ATCC (CCL-185) and used widely in cell biology and virology research.
No human-subjects research is involved; no consent is required for established cell lines under CZ Biohub institutional guidelines.

\paragraph{WTC-11 (hiPSC).}
WTC-11 is an induced pluripotent stem cell line derived from a healthy adult male donor at the Conklin Laboratory (Gladstone Institutes / UCSF) and available from the Coriell Institute (GM25256).
The Allen Institute used genome editing to insert endogenous fluorescent tags.
The original donor consented to research use of the derived line; the released image data contain no genomic sequence or other donor-identifying information.

\paragraph{Viral infection (A549).}
Zika and Dengue virus infections were performed under institutional BSL-2 protocols at Biohub SF.
The release consists only of digital microscopy volumes; no physical or infectious material is present or derivable from the datasets.
 \addtocontents{toc}{\protect\setcounter{tocdepth}{2}}

\newpage

\section*{NeurIPS Paper Checklist}

\begin{enumerate}

\item {\bf Claims}
    \item[] Question: Do the main claims made in the abstract and introduction accurately reflect the paper's contributions and scope?
    \item[] Answer: \answerYes{} %
    \item[] Justification: The abstract and Sec.~\ref{sec:introduction} state the four contributions claimed by the paper: the paired 3D time-lapse A549 dataset, the curated iPSC training corpus, the regression and generative baselines trained under a shared protocol, and the three-tier metric panel. The empirical findings (Sec.~\ref{sec:results}) and documented limitations (Sec.~\ref{sec:limitations}) match those claims.

\item {\bf Limitations}
    \item[] Question: Does the paper discuss the limitations of the work performed by the authors?
    \item[] Answer: \answerYes{} %
    \item[] Justification: Sec.~\ref{sec:limitations} briefly summarizes the work's scope and future directions, while App.~\ref{sec:limitations-detail} details metric assumptions, dataset diversity, baseline coverage, the bounded scope of DynaCell evaluation scores, and temporal-consistency limitations.

\item {\bf Theory assumptions and proofs}
    \item[] Question: For each theoretical result, does the paper provide the full set of assumptions and a complete (and correct) proof?
    \item[] Answer: \answerNA{} %
    \item[] Justification: The paper presents an evaluation suite with empirical results; it contains no theorems, lemmas, or formal proofs.

    \item {\bf Experimental result reproducibility}
    \item[] Question: Does the paper fully disclose all the information needed to reproduce the main experimental results of the paper to the extent that it affects the main claims and/or conclusions of the paper (regardless of whether the code and data are provided or not)?
    \item[] Answer: \answerYes{} %
    \item[] Justification: The data preprocessing and split protocol are documented in Sec.~\ref{sec:datasets}. Baselines, training-pool compositions, and the three-tier evaluation protocol are described in Sec.~\ref{sec:baseline-models} and Sec.~\ref{sec:eval-protocol}. Per-baseline hyperparameters appear in Tab.~\ref{tab:benchmark_training} and the augmentation pipeline in Tab.~\ref{tab:augmentation}. We additionally release preprocessed OME-Zarr volumes, trained model checkpoints, and evaluation code (App.~\ref{sec:license-hosting}).
    \item[] Guidelines:
    \begin{itemize}
        \item The answer \answerNA{} means that the paper does not include experiments.
        \item If the paper includes experiments, a \answerNo{} answer to this question will not be perceived well by the reviewers: Making the paper reproducible is important, regardless of whether the code and data are provided or not.
        \item If the contribution is a dataset and\slash or model, the authors should describe the steps taken to make their results reproducible or verifiable. 
        \item Depending on the contribution, reproducibility can be accomplished in various ways. For example, if the contribution is a novel architecture, describing the architecture fully might suffice, or if the contribution is a specific model and empirical evaluation, it may be necessary to either make it possible for others to replicate the model with the same dataset, or provide access to the model. In general, releasing code and data is often one good way to accomplish this, but reproducibility can also be provided via detailed instructions for how to replicate the results, access to a hosted model (e.g., in the case of a large language model), releasing of a model checkpoint, or other means that are appropriate to the research performed.
        \item While NeurIPS does not require releasing code, the conference does require all submissions to provide some reasonable avenue for reproducibility, which may depend on the nature of the contribution. For example
        \begin{enumerate}
            \item If the contribution is primarily a new algorithm, the paper should make it clear how to reproduce that algorithm.
            \item If the contribution is primarily a new model architecture, the paper should describe the architecture clearly and fully.
            \item If the contribution is a new model (e.g., a large language model), then there should either be a way to access this model for reproducing the results or a way to reproduce the model (e.g., with an open-source dataset or instructions for how to construct the dataset).
            \item We recognize that reproducibility may be tricky in some cases, in which case authors are welcome to describe the particular way they provide for reproducibility. In the case of closed-source models, it may be that access to the model is limited in some way (e.g., to registered users), but it should be possible for other researchers to have some path to reproducing or verifying the results.
        \end{enumerate}
    \end{itemize}

\item {\bf Open access to data and code}
    \item[] Question: Does the paper provide open access to the data and code, with sufficient instructions to faithfully reproduce the main experimental results, as described in supplemental material?
    \item[] Answer: \answerYes{} %
    \item[] Justification: App.~\ref{sec:license-hosting} summarizes data and code released under permissive terms where feasible (CC-BY-4.0 for the A549 portion, Allen Institute Terms of Use for the iPSC derivative, and BSD-3-Clause for code) and AWS Open Data hosting for long-term maintenance. App.~\ref{sec:reviewer-sample} describes a public data sample under the 4\,GB E\&D track guideline. Benchmark code is available at \url{https://github.com/Biohub/VisCy/tree/main/applications/dynacell} and will be maintained. The evaluation package will be released through PyPI upon publication.

\item {\bf Experimental setting/details}
    \item[] Question: Does the paper specify all the training and test details (e.g., data splits, hyperparameters, how they were chosen, type of optimizer) necessary to understand the results?
    \item[] Answer: \answerYes{} %
    \item[] Justification: Splits and benchmarking axes are specified in Sec.~\ref{sec:datasets}; baseline architectures, loss functions, and the iPSC, A549, and joint training sets in Sec.~\ref{sec:baseline-models}; full hyperparameters in Tab.~\ref{tab:benchmark_training}; and augmentations in Tab.~\ref{tab:augmentation}. Each model selects checkpoints on a reproducible 80/20 validation split from its training FOVs, while the separate test stores remain untouched; we do not claim one shared validation partition across model families. Data manifests and all training configurations are released with the benchmark.

\item {\bf Experiment statistical significance}
    \item[] Question: Does the paper report error bars suitably and correctly defined or other appropriate information about the statistical significance of the experiments?
    \item[] Answer: \answerYes{} %
    \item[] Justification: Per-FOV variability is reported as $\pm$ one standard deviation across fields of view on the perturbation-robustness trajectories in Fig.~\ref{fig:a549-robustness}B. Benchmark tables report FOV or resampling variability where available. We did not repeat every training run, so these intervals measure variation across test fields or metric resamples, not variation across model initializations.

\item {\bf Experiments compute resources}
    \item[] Question: For each experiment, does the paper provide sufficient information on the computer resources (type of compute workers, memory, time of execution) needed to reproduce the experiments?
    \item[] Answer: \answerNo{} %
    \item[] Justification: Tab.~\ref{tab:benchmark_training} reports training budgets and hyperparameters, and App.~\ref{sec:compute-cost} reports accelerator types and aggregate accelerator-hours, including failed, crashed, restarted, and running attempts. The paper does not report memory use and execution time for every experiment.
    
\item {\bf Code of ethics}
    \item[] Question: Does the research conducted in the paper conform, in every respect, with the NeurIPS Code of Ethics \url{https://neurips.cc/public/EthicsGuidelines}?
    \item[] Answer: \answerYes{} %
    \item[] Justification: The work uses established cell lines under institutional guidelines; viral infection experiments were conducted under BSL-2 protocols; no personal data or human subjects are involved. The A549 data are released under CC-BY-4.0, and the iPSC derivative follows the Allen Institute Terms of Use, with full provenance documentation.

\item {\bf Broader impacts}
    \item[] Question: Does the paper discuss both potential positive societal impacts and negative societal impacts of the work performed?
    \item[] Answer: \answerYes{} %
    \item[] Justification: App.~\ref{sec:broader-impact} discusses positive impacts (reducing phototoxicity, enabling multiplexed live-cell imaging) and restrictions on use (virtual stains must not be used for clinical or diagnostic decisions; perturbation-response scores from virtual stains are research signals only).
    
\item {\bf Safeguards}
    \item[] Question: Does the paper describe safeguards that have been put in place for responsible release of data or models that have a high risk for misuse (e.g., pre-trained language models, image generators, or scraped datasets)?
    \item[] Answer: \answerNA{} %
    \item[] Justification: The dataset comprises digital microscopy volumes of established cell lines with no identifiable personal information. The release contains no physical or infectious material. The misuse risk is low.

\item {\bf Licenses for existing assets}
    \item[] Question: Are the creators or original owners of assets (e.g., code, data, models), used in the paper, properly credited and are the license and terms of use explicitly mentioned and properly respected?
    \item[] Answer: \answerYes{} %
    \item[] Justification: The Allen Institute hiPSC dataset~\cite{viana2023integrated} (Allen Institute Terms of Use), waveorder phase-reconstruction library (BSD-3-Clause), and VisCy benchmark code (BSD-3-Clause) are cited with their terms stated in App.~\ref{sec:license-hosting} and the supplementary datasheet. Allen Institute organelle-segmentation models used for evaluation remain under their original terms.

\item {\bf New assets}
    \item[] Question: Are new assets introduced in the paper well documented and is the documentation provided alongside the assets?
    \item[] Answer: \answerYes{} %
    \item[] Justification: The A549 and iPSC datasets are documented with a datasheet following Gebru et al. (App.~\ref{sec:datasheet}) and Croissant metadata with responsible-AI fields. Code for the DynaCell evaluation suite is documented in the VisCy repository at \url{https://github.com/Biohub/VisCy/tree/main/applications/dynacell}.

\item {\bf Crowdsourcing and research with human subjects}
    \item[] Question: For crowdsourcing experiments and research with human subjects, does the paper include the full text of instructions given to participants and screenshots, if applicable, as well as details about compensation (if any)?
    \item[] Answer: \answerNA{} %
    \item[] Justification: No crowdsourcing or research with human subjects was performed. All data derive from established immortalized cell lines imaged in vitro.

\item {\bf Institutional review board (IRB) approvals or equivalent for research with human subjects}
    \item[] Question: Does the paper describe potential risks incurred by study participants, whether such risks were disclosed to the subjects, and whether Institutional Review Board (IRB) approvals (or an equivalent approval/review based on the requirements of your country or institution) were obtained?
    \item[] Answer: \answerNA{} %
    \item[] Justification: The research does not involve human subjects. Cell lines used (A549, WTC-11) are established lines available from public repositories; no human participants were enrolled. Viral infection experiments were conducted under institutional BSL-2 biosafety protocols.

\item {\bf Declaration of LLM usage}
    \item[] Question: Does the paper describe the usage of LLMs if it is an important, original, or non-standard component of the core methods in this research? Note that if the LLM is used only for writing, editing, or formatting purposes and does \emph{not} impact the core methodology, scientific rigor, or originality of the research, declaration is not required.
    \item[] Answer: \answerNA{} %
    \item[] Justification: LLMs assisted routine code completion and refactoring. The authors specified the scientific methods, reviewed assisted changes, and tested the affected pipelines. LLMs are not a component of the virtual-staining models or evaluation protocol; their use is also documented in the supplementary datasheet and Croissant metadata.

\end{enumerate}
 
\end{document}